%% file: final-report-features-on-fisheye.tex
\documentclass[a4paper,10pt]{article}
\usepackage{authblk} 
\usepackage[utf8x]{inputenc} 
\usepackage{textcomp} 

\usepackage[french,english]{babel} 
\usepackage[english]{varioref}
\usepackage{babelbib} 
\usepackage[T1]{fontenc} 
\usepackage{lmodern} 

\usepackage{cite} 

\usepackage{color} 

\usepackage[colorlinks=true,linkcolor=black,citecolor=black,filecolor=black,menucolor=black,urlcolor=black]{hyperref} 

\usepackage{tabto} 

\usepackage{array} 
\usepackage{multirow} 
\usepackage{hhline} 
\usepackage{tabularray} 

\usepackage{calc} 

\usepackage{vmargin}
\usepackage{graphicx} 
\usepackage{subfig} 

\usepackage{amsmath} 
\usepackage{amsfonts} 
\usepackage{amssymb} 

\usepackage{adjustbox} 

\title{Feature points evaluation on omnidirectional vision with a photorealistic fisheye sequence\\\textit{A report on experiments done in 2014}}
\author[1]{Julien \textsc{Moreau}}
\author[2]{Sébastien \textsc{Ambellouis}}
\author[3]{Yassine \textsc{Ruichek}}
\affil[1]{Université de technologie de Compiègne, CNRS, Heudiasyc, Compiègne, France}
\affil[2]{Université Gustave Eiffel, COSYS/LEOST, Villeneuve d'Ascq, France}
\affil[3]{Université de technologie de Belfort-Montbéliard, CIAD, Belfort, France}

\begin{document}
\maketitle




\section{Foreword}

\paragraph{What is this report:\\}
This is a scientific report, contributing with a detailed bibliography, a dataset which we will call now PFSeq for ``Photorealistic Fisheye Sequence'' and make available at \url{https://doi.org/10.57745/DYIVVU}, and comprehensive experiments.
This work should be considered as a draft, and has been done during my PhD thesis ``Construction of 3D models from fisheye video data—Application to the localisation in urban area'' in 2014~\cite{Moreau16}.
These results have never been published.
The aim was to find the best features detector and descriptor for fisheye images, in the context of self-calibration, with cameras mounted on the top of a car and aiming at the zenith (to proceed then fisheye visual odometry and stereovision in urban scenes).
We face a chicken and egg problem, because we can not take advantage of an accurate projection model for an optimal features detection and description, and we rightly need good features to perform the calibration (\textit{i.e.} to compute the accurate projection model of the camera).

\paragraph{What is not this report:\\}
It does not contribute with new features algorithm.
It does not compare standard features algorithms to algorithms designed for omnidirectional images (unfortunately).
It has not been peer-reviewed.
Discussions have been translated and enhanced but the experiments have not been run again and the report has not been updated accordingly to the evolution of the state-of-the-art (read this as a 2014 report).

\section{Introduction}

\begin{figure}[htbp]
  \centering
  \includegraphics[keepaspectratio=true, width=\columnwidth]{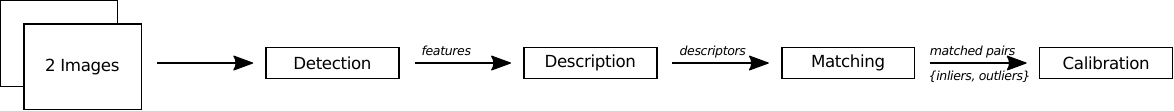}
  \caption{General pipeline of camera self-calibration based on matched features. \label{fig:scalib}}
\end{figure}

Self-calibration pre-processing is summarized in Figure~\ref{fig:scalib}.
It is related to feature-based visual odometry, as we need to find the matching features in two pictures of the same scene.
These matches are the projection of the same element of the world.
Taken with the assumption of a rigid world, they can be used to estimate the relationship between the two views in terms of epipolar geometry (usually with the methods called the 7-points or the 8-points algorithms).
This allows to extract the relative pose between both viewpoints, leading to visual odometry or to extrinsic calibration.
The missing information is the true scale, and it can be recovered when we know the baseline between both viewpoints or given distances in the scene, or if we rely also on additional sensors.
In this work, we rely on a 9-points algorithms, which combines in the same operation the estimation of a fisheye projection parameter, that is, it adds intrinsic calibration.
Calibration process itself has to deal with the presence of erroneous matches issued from this pre-processing.
It usually relies on a \textsc{ransac} (RANdom SAmple Consensus) strategy on top of the calibration algorithm, thanks to its ability to separate matches into inliers (correct matches according to estimated model) and outliers (erroneous matches) sets.
\\

In order to reach an accurate and stable camera self-calibration, it is essential to get as many features as possible, detected and described with accuracy to be matched properly (inliers).
Regarding omnidirectional images and their typical extreme distortions, methods robust to scale variations and rotations are usually the way to go.
Authors use such algorithms, even if they were initially designed with planar images in mind (\textit{i.e.} captured with classical perspective cameras).
This methodology is not so accurate because classic algorithms do not take into account the full specific geometry of omnidirectional visual sensors.
This is why other authors propose to rectify the omnidirectional images into planar images, to be able to process them as such with standard features detection algorithms.
The drawbacks of this procedure are:
\begin{itemize}
 \item the need to know the sensor calibration to be able to undistort the images (at least with an approximation),
 \item the interpolation, that shall follow the lens projection and that is prone to generate errors in the undistorted image.
\end{itemize}
To overcome these limitations, features algorithms specifically made for omnidirectional images are needed.
\\

A state of the art about these omnidirectional features and existing features benchmarks is given in sections~\ref{sec:bib-omni-features} and \ref{sec:bib-benchmarks}, respectively.
\\
In section~\ref{sec:our-benchmark}, we propose a simulated dataset and benchmark, PFSeq, to focus on the challenges of features detection and matching on fisheye images, within the scenario of a moving camera observing tall buildings.
PFSeq is done so that it can also be used to evaluate fisheye self-calibration, visual odometry, sky segmentation and stereo matching and reconstruction through true dense distance maps.
\\
Sections~\ref{sec:eval-simu} and \ref{sec:eval-reel} show tests and evaluations, respectively on PFSeq with its ground truth, and on on real images to validate the results when transferring to real domain.
Many standard features detection and description algorithms are compared, to show their abilities on omnidirectional fisheye images.
\\
Finally, a discussion on our findings is given in section~\ref{sec:conclu}.


\section{Feature points designed for omnidirectional imagery}
\label{sec:bib-omni-features}


Many users simply use features designed for perspective images with a sub-optimal effectiveness.
Few authors investigate features and descriptors adapted to omnidirectional images.
\\
In 2004, SIFT detector and descriptor is proposed by~\cite{Lowe04}.
It is a breakthrough for the features algorithms invariant to scale and orientation changes.
SIFT is still recognized as the reference among the feature points algorithm.
Regarding omnidirectional images, nearly all of following researches try to adapt SIFT to these geometries.
Thus, we review first non-SIFT methods, then SIFT variants.
\\

\subsection{Non-SIFT features algorithms for omnidirectional vision}

\cite{Ieng03} is probably the first work on this problem.
Their approach is to change the shape and the size of the correlation windows accordingly to omnidirectional catadioptric cameras geometry.
Authors show results with dynamic windows centered on Harris corner detector, but proposed principles are generic and can be applied to any feature point.
Windows are built following the surface of the catadioptric sensor instead of the image plane.
This process requires to know intrinsic parameters of the sensor, as such, to calibrate it beforehand.
To get squared windows in the image plane, windows are diamond-shaped onto the sensor's surface.
To get descriptors robust to rotations, these surfaces are oriented along the radial direction.
They are defined considering an angular aperture along azimuth and elevation from the point's back-projection ray.
The fixed angular aperture lead to increasing window size accordingly to the distance from the point to the optical center.
Then the windows are resized to a common size, using bilinear interpolation.
To improve the quality of the method, the authors propose to use multiple aperture angles, as it is related to observed objects' distance.
\\

Since the publication of SIFT~\cite{Lowe04}, the only different method is the matching method for regions proposed by~\cite{Mauthner06}.
For each detected region, a virtual perspective camera is estimated to get a perspective image of the region with no distortion.
These recomputed regions are then compared for the matching process.
They experiment this with catadioptric sensors.
They show the effectiveness of their method by applying the region matching to stereovision and comparing the estimated 3D position to a LiDAR ground truth.
\\

\subsection{SIFT variants tuned for omnidirectional vision}
\label{ssec:bib-omni-features-sift}

Hansen \textit{et al.} are the first to propose a SIFT detector implementation for fisheye images~\cite{Hansen07b}.
SIFT relies on gaussian filters for its multi-scale representation.
With omnidirectional images, using gaussian filter on image plane is not ideal.
The authors propose to use instead a spherical diffusion function, that can be linked to a gaussian on a sphere~\cite{Bulow04}.
This method only requires to know the optical centre and the radius of the fisheye projection, not all the intrinsic parameters.
This works in the frequency domain.
They also provide a method to fit SIFT descriptor to omnidirectional images.
With classical images, the descriptor is built from a circular region around the point in image plane.
In the case of omnidirectional images, the authors propose to consider also a circular region around the point, but to project it on the sphere surface.
This region is defined thanks to a solid angle, whose origin is the sphere centre and axis given by considered point.
The angle itself is fixed in accordance with the scale level of the image.
This area can then be projected onto the image plane to extract the meaningful values for the descriptor computation.
However, projected area is not circular.
To simplify the process, authors propose to fit a circle onto this, this approximation allows to compute the descriptor as in the case of classical SIFT.
In their evaluation, they show an improvement of the matching on fisheye images with their method.
\\
In their second paper~\cite{Hansen07a}, Hansen \textit{et al.} release a weakness in their algorithm: the spherical Fourier transform is heavy to compute and requires large bandwidth to avoid aliasing effects.
Hence, they introduce a new anti-aliasing filter that allows to save most of computation costs with a minimal loss in the final matching quality.
This method is later called sSIFT, as spherical SIFT, in~\cite{Hansen08}.
\\

Hansen \textit{et al.} also propose other works for semi-calibrated omnidirectional sensors (knowledge of the optical centre and the radius)~\cite{Hansen08}.
They present a new variant of SIFT for omnidirectional images called
pSIFT, as parabolical SIFT.
It consists in approximating the spherical diffusion by considering the stereographic projection model.
This method is not affected by the bandwidth problem and high computation cost to reduce aliasing.
They compare their two sSIFT and pSIFT algorithms as well as the original SIFT on fisheye images (projection type is undetermined) and on equiangular projection catadioptric images.
For the fisheye sequence, the best algorithm is sSIFT, followed by pSIFT.
For the high resolution catadioptric sequence, sSIFT is also the more accurate, but it generally allows for less matched points than pSIFT, itself providing less points couples than original SIFT.
For the lower resolution catadioptric sequence, the best algorithm is pSIFT by a large margin over SIFT, who in this case is better than sSIFT.
Hansen \textit{et al.} also present their result in an article~\cite{Hansen10}, including additional calibration and visual odometry applications.
\\

In the work of~\cite{Hadj08}, the authors propose to use the Harris detector, together with a SIFT descriptor adapted to the geometry of omnidirectional images.
Their contribution is a way to project a 2D omnidirectional image onto a sphere and then to process it as a sphere.
Hence they can work in a spherical frequency domain.
The goal is to improve the accuracy of all the processing done on omnidirectional images.
The detection and description application is a way to show the effectiveness of their proposal, while their final goal is a visual odometry application.
However, the projection of the image on a sphere requires to know all the calibration parameters.
For the detector, they choose Harris algorithm as it is faster and has a higher repeatability than SIFT without multi-scale processing.
For the descriptor, they use the SIFT variant of~\cite{Hansen07b}.
In their evaluation, on real images, they show a clear improvement when using the adapted features algorithms for essential matrix estimation.
With classical features algorithms, too many matching errors occur, leading to wrong essential matrix estimation.
\\

Authors of~\cite{Cruz11} propose a SIFT in spherical coordinates and two descriptors: LSD (Local Spherical Descriptor) and LPD (Local Planar Descriptor).
Similarly to~\cite{Hansen07a}, they process the image projected onto a sphere, and must propose an anti-aliasing method.
They apply riemannian geometry.
Points are matched in spherical coordinates.
LSD descriptor serves to find correspondences between omnidirectional images, while LPD serves to find correspondences between omnidirectional and planar images.
LSD uses the spherical representation of the image.
Region size depends on the scale where the feature is detected.
LSD descriptor is a 3-dimensional histogram (2 for space and 1 for orientation).
LPD is made to be comparable with classic SIFT descriptor.
It consists of a SIFT descriptor computed for an approximated projection of the feature-centered region on a plane.
To make this projection, the authors apply stereographic projection on the tangent plane to the sphere, that passes through the antipodal point to considered feature.
Values in this flattened region are interpolated to get a fixed-size window.
They evaluate this method on spherical images issued from a Ladybug camera (an assembly of 6 wide angle views to cover an omnidirectional field), and on parabolic catadioptric images.
They show that for omnidirectional images, their spherical SIFT with LSD is the best choice.
When using jointly spherical and planar images, their spherical SIFT with LPD (for the omnidirectional image) or with classic SIFT descriptor (for the planar image), gives the best results.
To reduce computation cost, they finally suggest to try a similar adaptation of SURF detector~\cite{Bay06} with GLOH descriptor~\cite{Mikolajczyk05a}.
\\

Arican \textit{et al.} propose OmniSIFT method~\cite{Arican10b, Arican12}, they use with catadioptric sensors.
They apply riemannian geometry to preserve the geometry and visual information of catadioptric omnidirectional cameras.
They propose an algorithm based on the sphere, and a polar descriptor.
Smoothed images are computed on the sphere thanks to stereographic projection and Laplace-Beltrami operators that fit with parabolic and hyperbolic mirrors. 
An other contribution is the use of discrete operators to control the scale factor, and additional optimisations to save computations.
This method provides a uniform smoothing for the sphere, while being computed directly on the image plane.
This is an advantage over~\cite{Hansen07b}, where the image must be projected then smoothed on the sphere surface.
Proposed descriptor is declined in two versions: an oriented and a non-oriented polar descriptor.
They are computed considering a circular area around the point on sphere surface, whose size is related to the scale factor.
This region is projected onto the image plane to extract the points used for their computation.
They adapt the gradients computation to consider the sampling distance between two neighbor pixels (colatitude angle varies with the position in the image).
Their descriptor is inspired by GLOH descriptor~\cite{Mikolajczyk05a} and thus discretizes polar coordinates around the feature point.
OmniSIFT algorithm is compared with and outperforms previous propositions of~\cite{Hansen07b} and~\cite{Cruz11}.
Both proposed descriptors are better than these previous works, with an advantage for the oriented descriptor, except in the case of matching after pure translational motion between the viewpoints.
While being the best-performing algorithm for omnidirectional images, it requires less computations where it is similar to classical SIFT.
The drawback of this proposition is its conception restricted to catadioptric sensors.
Operators adapted to fisheye image might be required to extend this work.
\\

\cite{Puig11, Puig13} extend the Laplace-Beltrami operator to any kind of catadioptric central sensor, based on the spherical camera model.
The spherical model can fit any omnidirectional projection as long as we choose the good projection function.
Riemannian geometry allows to encode the geometry of the mirrors and to work on the surface of the catadioptric mirror, and to project the catadioptric image on a unit sphere.
This allows also to make computations on the image plane directly while considering the mirror geometry.
To smooth omnidirectional images, they fit coefficients to the sensor then can convolute the image with some simple kernels.
This method surpasses the classical SIFT.
Unfortunately, while being an extension of~\cite{Arican10b, Arican12}, the authors do not provides this comparison.
We can expect similar results.
C++ code of their operator was available at \url{http://webdiis.unizar.es/~lpuig/}.
This method might work with fisheye projections, provided the good model coefficients, this is something to test.
\\

\cite{Lourenco11, Lourenco12} propose RD-SIFT detector algorithm, and a lightweight variant called sRD-SIFT.
Their method has the advantage of directly processing the planar image (no intermediate projection on a sphere), and the camera does not need to be accurately calibrated.
Distortions are modelled with the first order division model (only one parameter), which is compatible with fisheye but not with catadioptric sensors.
Optical centre is approximated with the image centre.
To compute images at different scales, because of the problem of gaussian blur for omnidirectional geometry, they rectify the image distortions, apply the gaussian filter, then apply back the distortions.
Note that they do not compute SIFT on the rectified images, they only use the rectification to compute smoothed images.
For sRD-SIFT lightweight variant, they change the gaussian filter to include the parameter of the division model to get an adaptive kernel according to the distance to the optical centre, allowing to avoid the de-distortion re-distortion steps and interpolations.
Hence, sRD-SIFT computation time is comparable to classical SIFT.
Proposed algorithms are compared against the classical SIFT applied on rectified images on a plane, they call rectSIFT, and against pSIFT~\cite{Hansen08}.
While they approximate the calibration parameters for their own algorithms, they do an accurate calibration for pSIFT.
It is interesting to note that RD-SIFT is still good with very inaccurate calibration.
RD-SIFT detector is good for highly distorted images, \textit{i.e.} fisheye, but is not the best for wide angle images that have less than 45\% radial distortions. In this case, rectSIFT is better.
pSIFT detection has a slight advantage over sRD-SIFT thanks to its invariance to rotation.
For the descriptors, authors propose to compute SIFT descriptors first, then to apply a distortion correction on these, avoiding the need to interpolate image data.
Their descriptors show to be better than classical SIFT descriptors as is, but the best of all is the descriptor of rectSIFT.
Still, sRD-SIFT descriptor seems to be slightly better than pSIFT.
An other evaluation, based on the results of a visual odometry algorithm with \textsc{ransac}, shows that sRD-SIFT provides the most inliers, closely followed by pSIFT, and far more than rectSIFT.
It gives also the best accuracy on the reprojection test.
Authors show that sRD-SIFT is the best for visual odometry applications on endoscopy and mobile robot.
Matlab source code was available at \url{http://arthronav.isr.uc.pt/~mlourenco/srdsift/}.

\section{Evaluation procedure of detectors and detector / descriptor / matching together}
\label{sec:bib-benchmarks}

Many works aim at comparing features detectors and descriptors.
One of the most influential is~\cite{Mikolajczyk05a}, and they provide a set of real images for the tests.
In this section, we present previous features evaluation works (done for classical perspective images).\\

For ease of readability, let us define vocabulary for the following of the report:
\begin{itemize}
 \item Matches: matched points according to a matching criteria using benchmark reference information, consider detections only, can be seen as reference matches or potential matches to reach with automatic matching based on descriptors.
 \item Attempted matches: matched points using an automatic matching algorithm, it is split into two subsets that are the true (or correct, good) and the false (or incorrect, wrong) matches, whether they validate or not benchmark reference information.
\end{itemize}

\subsection{Benchmarks on perspective images}
\label{ssec:bib-eval-features}

Mikolajczyk \textit{et al.} evaluate detectors and descriptors in~\cite{Mikolajczyk05b} and~\cite{Mikolajczyk05a}.
Their benchmark uses real images of planar scenes, available at \url{http://www.robots.ox.ac.uk/~vgg/research/affine/}.
They focus on region features (blob-type points) detectors.
To check evaluated matches, as they use planar scenes, they rely on the homography between the two images to project the points and their local region from the evaluated picture to the reference picture.
The matches are correct if their feature areas overlap surpasses a fixed threshold (low superposition error).
Regions are elliptic and defined by the algorithms, meaning that they can vary in size and require a normalization step to be comparable.
To match the points for the descriptors evaluation, they simply check if the nearest neighbour feature in the descriptor space is below a given threshold.
\\
To evaluate the detectors, \cite{Mikolajczyk05b} use following measures:
\begin{itemize}
 \item repeatability score, as the ratio of matching pairs from all the detections over the smaller number of detected points in both images,
 \item accuracy in terms of localization of estimated region, illustrated by the evolution of the repeatability in function of the threshold on superposition error.
\end{itemize}
To evaluate the descriptors using the simple matching procedure described above, these authors use the matching score in~\cite{Mikolajczyk05b}, and introduce recall and 1-precision in~\cite{Mikolajczyk05a}:
\begin{itemize}
 \item matching score, as the ratio of the correct matches over the smaller number of detected points in both images,
 \item recall, defined as the number of correct matches over the number of matching pairs as for the repeatability numerator,
 \item 1-precision (read ``one minus precision''), defined as the number of false matches over the number of attempted matches (\textit{i.e.} correct matches + false matches).
\end{itemize}

\begin{figure}[htbp]
  \centering
  \includegraphics[keepaspectratio=true, width=80mm]{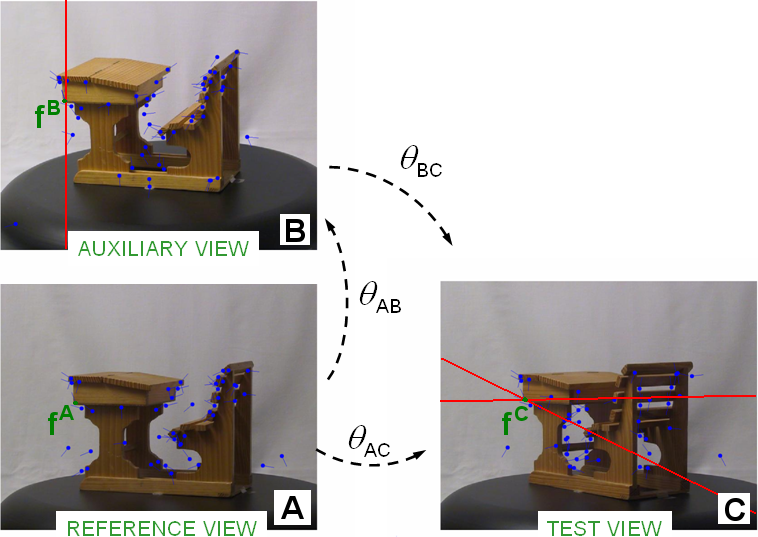} 
  \caption{Illustration of the method of~\cite{Moreels05} to find the true matching points. \label{fig:moreels}}
\end{figure}

Moreels \textit{et al.}~\cite{Moreels05} propose an other method to allow the generation of ground truth for matching points, using two epipolar constraints.
The advantage over previous works using homography, is to be able to consider non-planar scene.
In order to find the true positions of detected points in the test image, they apply trifocal stereovision where three calibrated cameras are set up following the shape of a L.
This gives a reference view (in the middle of the L), an auxiliary view above, and a test view on the right, as depicted in Figure~\ref{fig:moreels}.
They place 3D objects on a programmable turntable.
This way, they can reliably and automatically check the matches following geometry rules (they claim to reach approximately 2\% of erroneous references).
They evaluate detectors and descriptors together, and for that, they match the points following a fixed method: the ratio test procedure proposed by~\cite{Lowe04}, as a still simple and more realistic approach in real applications than a nearest neighbour search.
They propose following metrics, different from the previous ones:
\begin{itemize}
 \item detection rate, as the ratio of the correct matches over the number of attempted matches,
 \item false alarm rate, as the ratio of incorrect matches (false alarms) over the number of attempted matches and normalized by the size of the full database features.
\end{itemize}
Their benchmark fit better 3D object recognition applications than~\cite{Mikolajczyk05b}.
\\

Gauglitz \textit{et al.}~\cite{Gauglitz11} propose an other dataset to evaluate feature detection and description,
based on videos, and targeting at applications such as tracking and pose estimation.
Similarly to~\cite{Mikolajczyk05b}, their data is based on planar scenes in order to allow the use of homographies to generate ground truth.
They compare detectors and descriptors separately, and also combinations of detectors and descriptors.
\\
To evaluate the detectors, \cite{Gauglitz11} propose a new definition of repeatability:
\begin{itemize}
 \item repeatability, computed as the number of matches over the number of points in the reference frame ($t-1$ picture) instead of the minimum number of points between the two used images.
\end{itemize}
With the original definition of the repeatability, for example, it is possible to reach 1 in the case the detector detects only 1 point in one of the images.
Proposed definition is turned asymmetric and makes this drawback more acceptable with aimed applications, by considering $t-1$ frame as a reference frame for the evaluated $t$ frame.
\\
To evaluate the descriptors, \cite{Gauglitz11} use a matching process taking the closest description vector and the metric:
\begin{itemize}
 \item precision of the first nearest neighbor (precision of 1NN, or just precision), as the ratio of the correct matches over the number of attempted matches.
 It is similar to the detection rate of~\cite{Moreels05}.
\end{itemize}
They find this metric to be the most relevant for online tracking applications.
\\
In addition, to provide also an evaluation of the descriptors alone (\textit{i.e.} not depending on the detections), they simulate an ideal detector based on the reprojection of detected points from the first image to the second using the true homography transformation.
To enrich their benchmark on tracking, they evaluate a reprojection error of tracked points using an estimated homography from these points instead of the ground truth homography.
Finally, they provide a comprehensive analysis of the features detectors and descriptors in the context of points tracking.
Their work shows that there are many parameters to consider and that
the algorithms shall be carefully chosen in accordance to the usage context.
\\

Li proposes an evaluation of detectors and descriptors in a chapter of his thesis~\cite{Li13}, focused on egomotion estimation from stereo setup.
The common point between his application and ours (fisheye camera self-calibration, odometry and stereovision), is that the final estimations depend on the accuracy and also the spreading of used feature points.
\\
As such, he proposes following metrics to evaluate the detectors:
\begin{itemize}
 \item repeatability, using as denominator the number of detected features in the reference frame following Gauglitz~\cite{Gauglitz11} variant,
 \item uniformity, in order to evaluate the distribution of detected points in the image. The principle is to divide the image into a grid of fixed cells and to count the proportion of filled cells according to a given threshold, \textit{i.e.} cells that contain enough points.
\end{itemize}
To be able to measure the repeatability, he compares two images issued from a homography transform, allowing the computation of their true projection from an image to another.
To measure the uniformity, he splits the image in cells of $80\times{}80$ pixels and sets the filling threshold to 2 points per cell.
Proposed uniformity has two limitations:
the score is not normalised with the number of feature points to leverage its direct influence, and the arbitrary grid design might favor certain scenes accordingly to the image content.
For example within our application, a plain wall or the sky may not give any feature.
\\
To evaluate descriptors and matching, Li presents following metrics:
\begin{itemize}
 \item survival rate, defined as the number of attempted matched or tracked points in the research image over the number of detected points in the reference image,
 \item accuracy factor, defined as the number of inliers among attempted matches after a \textsc{ransac} scheme over the number of detected points in the reference image. It is used only with scenes without moving objects, to focus on mismatches caused by the algorithms and not by moving elements.
\end{itemize}
Note that the survival rate, while being similar to the matching score, does not rely on any ground truth to filter correct matches, and it does not take into account the points of the scene that go out of the field of view of the research frame: they are wrongly counted as lost points.
The accuracy factor, even if it relies on a \textsc{ransac} that can not be perfect, informs on the ability of the algorithms to provide more or less correct matching points.

\subsection{Benchmarks on omnidirectional images}

At the time this work has been done, there was no publicly available benchmarks to evaluate features on omnidirectional images.
The only author we found to propose a solution is Arican \textit{et al.}~\cite{Arican10b, Arican12}.
They simulate omnidirectional views of a scene containing a finite plane on a uniform blue background, which finally results in a small useful area from the full image.
They project real images onto the finite plane to create variants with the same viewpoints.
Knowing the rotation and translation between each viewpoint, together with the reference projection parameters, they are able to map true corresponding points.
Based on this reference, they compute repeatability, matching score, recall and 1-precision, with experiments varying translation and rotation separately.
In~\cite{Arican12} they also provide some result of hybrid matching evaluation between a pinhole view and an omnidirectional view from a similar synthetic scene.

\section{PFSeq dataset and benchmark proposal}
\label{sec:our-benchmark}

There is a lack of features benchmark on challenging omnidirectional images.
In this section, we detail our proposition to evaluate features with fisheye images, including the choice of some of the usual metrics together with a discussion about their limits.

\subsection{Our problem}

Evaluation data provided by the community fit to the pinhole projection model.
New data must be created for spherical projections and shared to the community.
Up to our knowledge, there is currently no solution to compute ground truth for real-world fisheye images with the guarantee of a perfect accuracy on the whole field of view.
\\
So, in this work, we propose first to generate virtual omnidirectional images with ground truth.
To be as close as possible to real images, we simulate at best various kind of optical and chromatic distortions that occur in reality.
Benchmark metrics and procedures are also discussed.
This allows to extensively benchmark detectors / descriptors / matching algorithms together, done in section~\ref{sec:eval-simu}.
In a second time, we propose an extension to validate the results with real-world image through stability evaluations on the targeted calibration application on a set of real omnidirectional images, done in section~\ref{sec:eval-reel}.

\subsection{Proposed sequence: PFSeq}
\label{ssec:pfseq}

To benchmark the performance of features on fisheye images, within the scope of a wide range of final applications (self-calibration, odometry and stereovision), a true ground truth is needed.
Proposed sequence of fisheye images has been done in computer-generated imagery, with Blender software v2.69 and Cycles engine (\url{http://www.blender.org/}), and is called ``Photorealistic Fisheye Sequence'' (PFSeq).
It contains ground truth distance maps in order to test whether pairs of points match, as well as the pose of reference camera for all the frames, and more.
Thanks to this information, PFSeq can be extended to benchmark various applications, making it suitable for:
\begin{itemize}
 \item Fisheye camera self-calibration thanks to the simulated setup parameters (intrinsics when using PFSeq's projection models, and extrinsics),
 \item Fisheye visual odometry thanks to the reference poses,
 \item Fisheye stereovision thanks to the distance maps,
 \item Fisheye feature points thanks to the distance maps,
 \item Fisheye sky segmentation thanks to the reference masks.
\end{itemize}

\begin{figure}[htbp]
  \centering
  \includegraphics[width=8cm]{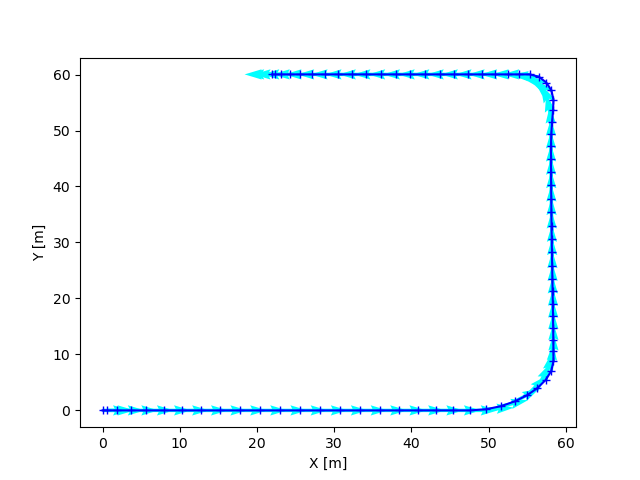}
  \caption{PFSeq trajectory and poses.}
\label{fig:trajectory}
\end{figure}

\paragraph{Scenario\\}
PFSeq plays the scenario of a car driving a U-shaped way in an urban canyon, shown in figure~\ref{fig:trajectory}.
It embeds a pair of fisheye cameras on its roof, along its longitudinal axis and looking at sky, recording the sequence when the car is navigating.
Generated frames are available in two variants to fit equidistant ($r = f \theta$, $r$ being the radius onto the image plane, $f$ the focal length, and $\theta$ the incidence angle) and equisolid angle ($r = 2 f \sin(\frac{\theta}{2})$) projection models, that are the most common one-parameter linear function models for fisheye projection.
We focused on the realistic rendering and do not
simulate more accurately real lenses with more complete models, such as non-central models.
\\

To reach photorealistic rendering, care has been done on following aspects:
\begin{itemize}
 \item Buildings' textures are issued from real photos.
 \item A blueish area makes the transition of the image circle.
 \item Purple fringing issued from chromatic aberrations.
 \item Lens flare/ghosting and sunstar when the sun is visible.
 \item Optical dispersion blur.
 \item Motion blur.
\end{itemize}

\paragraph{Sensors setup and data\\}

\begin{figure}[htbp]
  \centering
  \includegraphics[width=6cm]{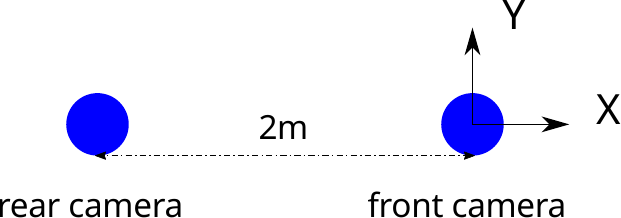}
  \caption{Stereo setup of PFSeq, viewed from top ($Z$ axis).}
\label{fig:setup}
\end{figure}

The sensors are made of a RGB fisheye stereovision setup with a baseline of 2m along $X$ axis, depicted in figure~\ref{fig:setup}.
Front camera is the reference camera of the stereo setup.
To avoid evident and unrealistic calibration parameters, cameras effective intrinsic parameters are set to fit a field of view of 181.8\textdegree{}, and a rotational shift is added on the rear camera as follows: -8\textdegree{} along $X$ axis, 6\textdegree{} along $Y$ axis, and -7\textdegree{} along $Z$ axis.
Rotation follows gimbals Euler XZY convention.
\\

\begin{figure}[htbp]
  \centering
  \subfloat[Realistic image.]{
  \includegraphics[width=45mm]{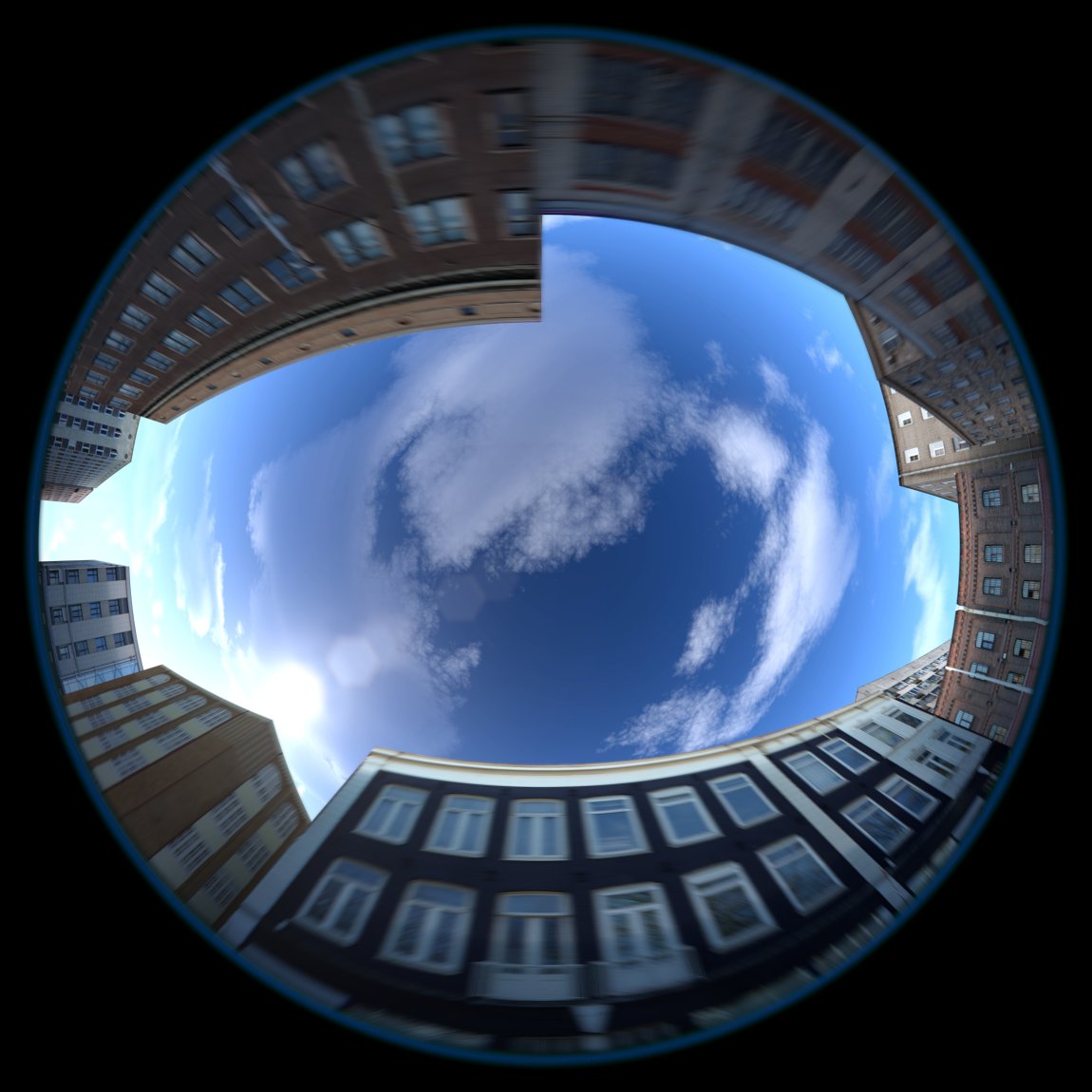}
  }
  \subfloat[Without blur and aberrations.]{
  \includegraphics[width=45mm]{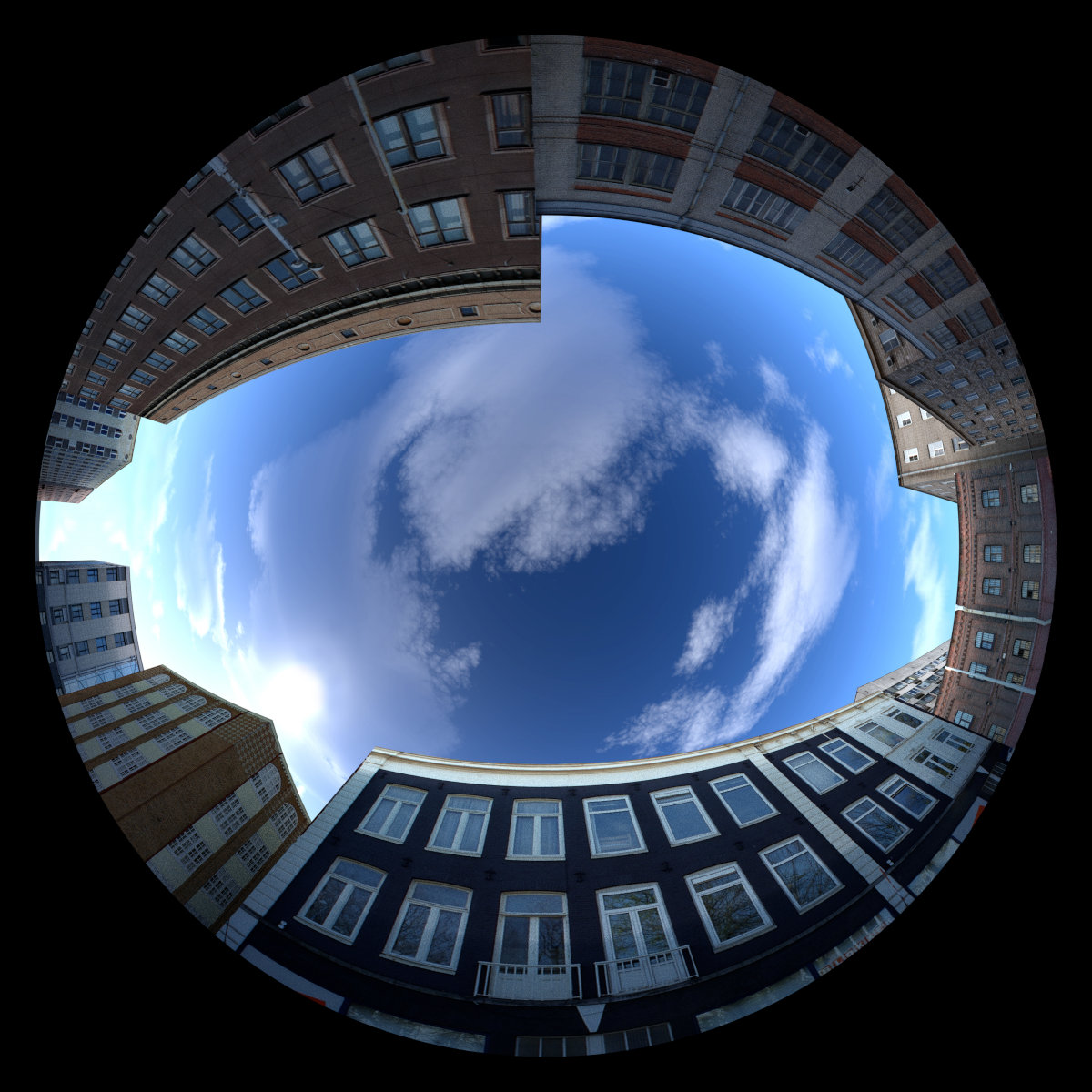}
  }
  \\
  \subfloat[Distance map.]{
  \includegraphics[width=45mm]{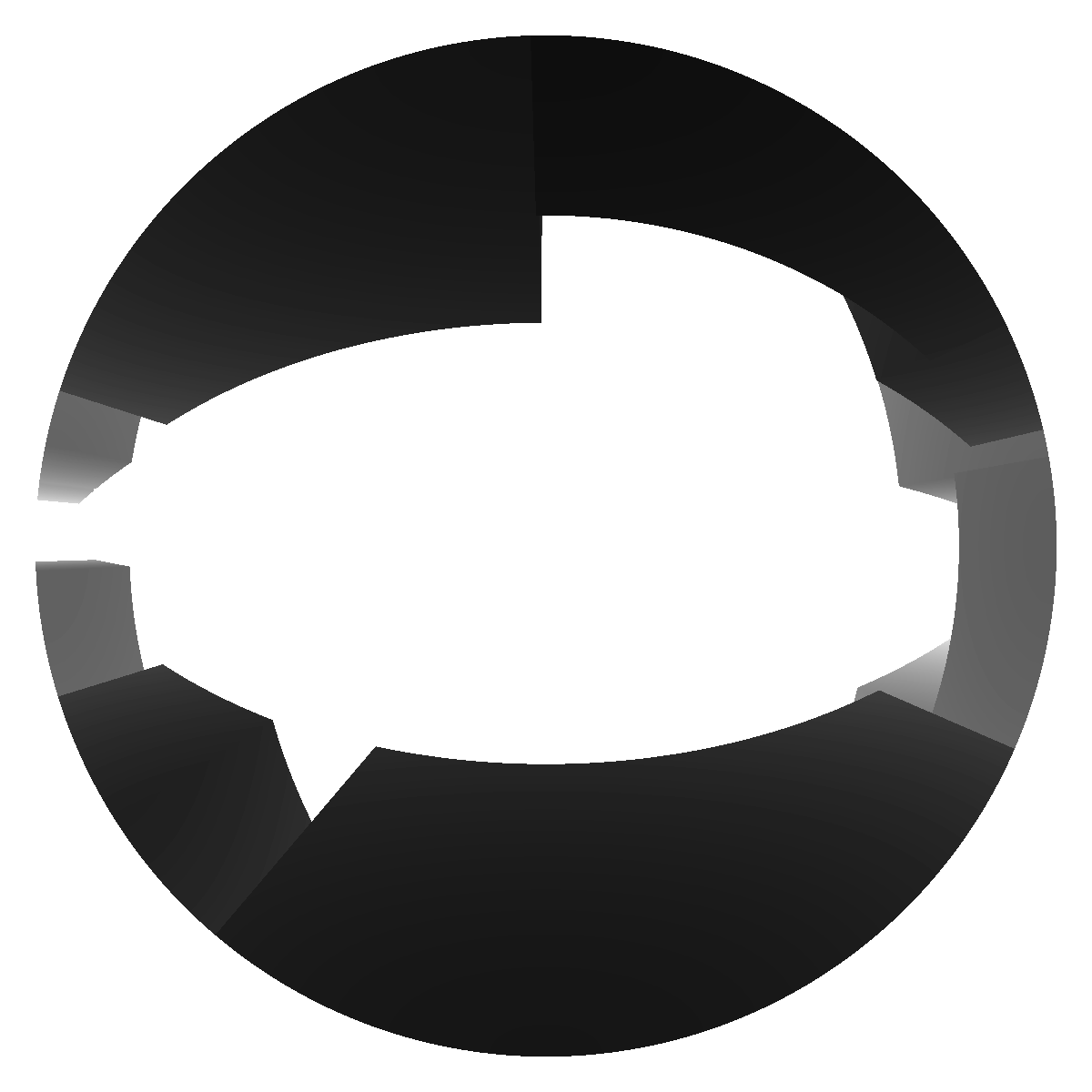}
  }
  \subfloat[Sky segmentation mask.]{
  \includegraphics[width=45mm]{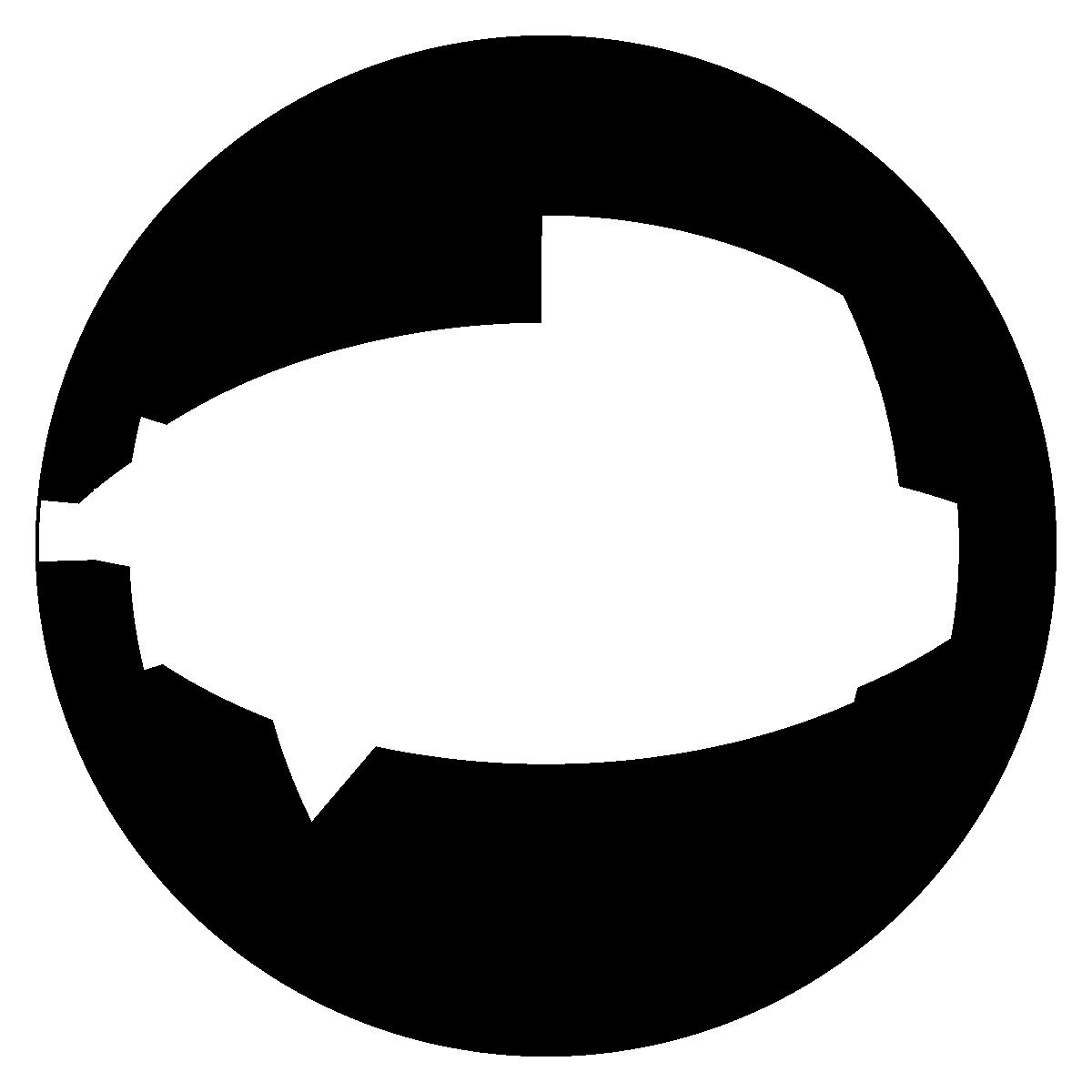}
  }
  \caption{
An example of available data in PFSeq.
}
\label{fig:examples}
\end{figure}

Cameras resolution is 1200 \texttimes{} 1200px.
Image circle diameter is 1200px for the equidistant images, and 1124px for the equisolid angle images.
Images are also available without the realistic optical and motion distortions.
True distances in the range maps are given up to 500m.
Sky masks are also provided.
Pose of the front camera is given for each shot as follows: $X$ and $Y$ position, rotation around $Z$.
Other parameters are null (simulated trajectory is done to be simple).
An illustration of available data is given in figure~\ref{fig:examples}.

\paragraph{Subset of typical situations\\}

\begin{figure}[htbp]
  \centering
  \subfloat[Pair 0 ``translation''.]{
  \includegraphics[width=45mm]{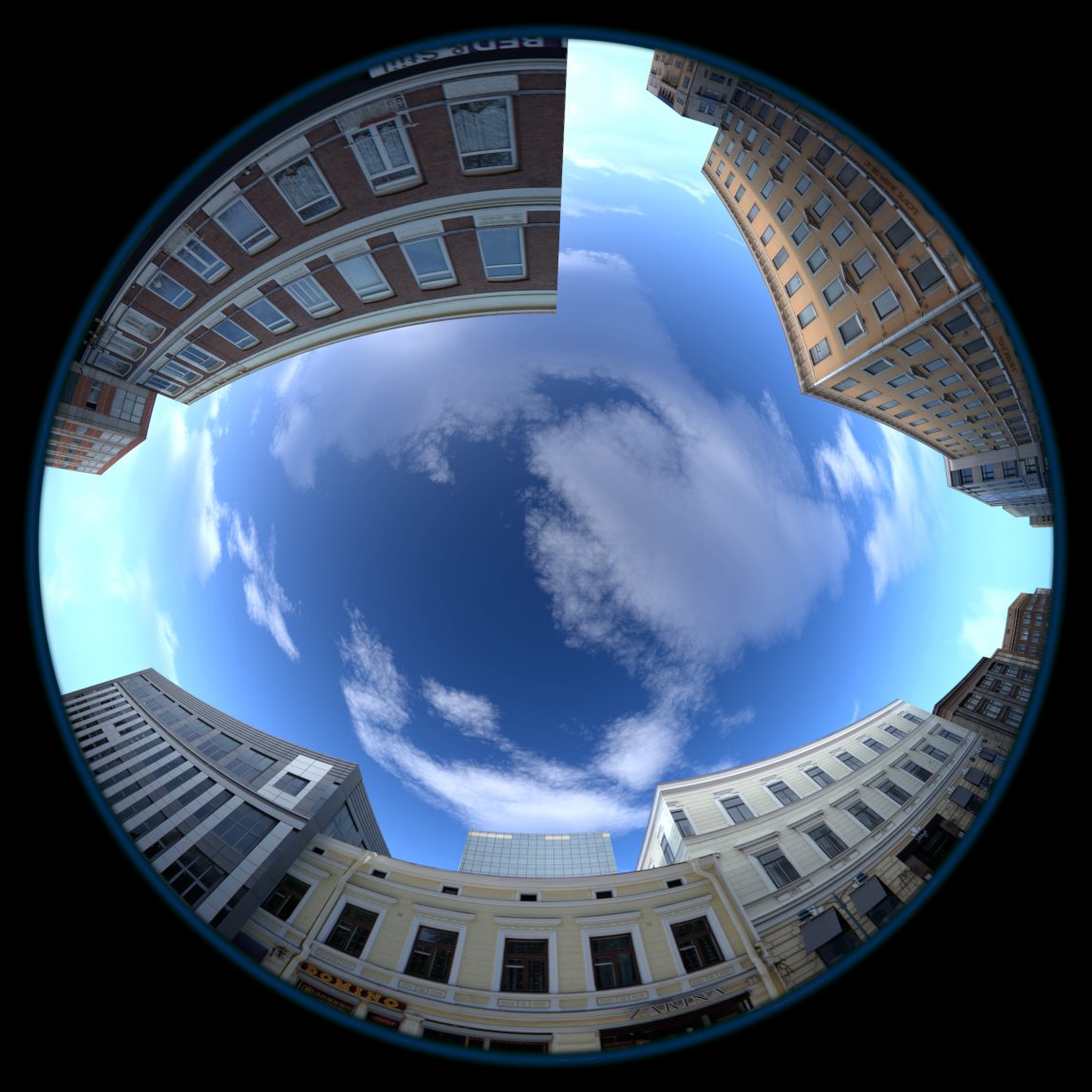}
  }
  \subfloat[Pair 40 ``translation motion''.]{
  \includegraphics[width=45mm]{./figures/Image0040.jpg}
  }
  \subfloat[Pair 24 ``translation and rotation motion''.]{
  \includegraphics[width=45mm]{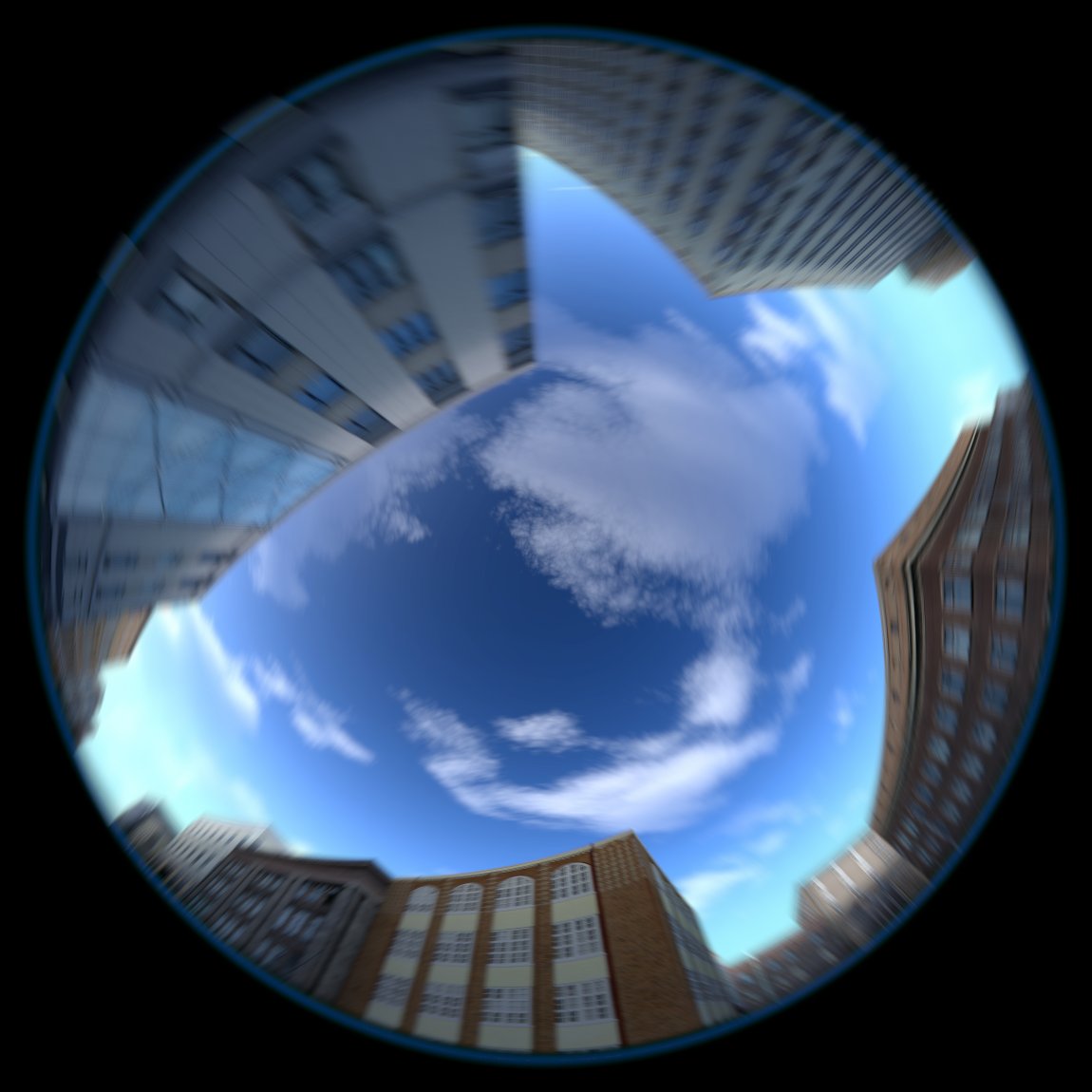}
  }
  \caption{
Selected typical frames of the various conditions related to the motion of the vehicle. Only front image of equisolid angle projection is shown.
}
\label{fig:3images}
\end{figure}

Along the sequence, we can raise 3 typical situations (figure~\ref{fig:3images}) to evaluate the features:
\begin{enumerate}
  \item ``translation'', no motion blur, the first frame of the sequence (pair no.~0),
  \item ``translation motion'', motion blur, like most of the images of PFSeq (pair no.~40),
  \item ``translation and rotation motion'', motion blur in a street corner, the most challenging case (pair no.~24).
\end{enumerate}

\subsection{Evaluation metrics}
\label{ssec:evmetrics}

Important information to highlight are the ability of the algorithms to provide many correct matches, accurate features, and features spread from the full scene.
Following metrics are used to estimate the quality of the detectors and of the complete detectors / descriptors / matching association.
They are inspired from existing works, described in section~\ref{ssec:bib-eval-features}, and rely on the definition of a matching criteria from the available reference data.

\paragraph{Proposed matching criteria\\} 
In addition to the image frames, we extracted the distance for each point into the views, in the form of distance maps.
Thanks to the knowledge of the true projection parameters and the distance map, it is possible to compute the 3D reprojection of each point in the world.
Then, the true relative pose between the two virtual cameras allows then to find the true matching points in their images using their 3D reprojection.
This will allow to find the number of matches and the number of correct matches used in selected evaluation metrics.
\\

Two ideas emerge to match the points in 3D space:
using an angular threshold, or using a metric threshold.
\begin{itemize}
 \item The former consider a given solid angle around the 3D projection rays of the points as threshold, and has the property to increase the tolerance to match the points proportionately to their distance to the cameras.
This might be seen as a quality as we require matching points spread at varied distances from the sensor.
This fits well 2D-only processing applications.
 \item The latter solution is based on a fixed metric distance threshold.
It can be seen as looking if the candidate match's 3D reprojection is inside a research sphere centred on the reference point's 3D reprojection.
Consequently, we can expect to have less validated points on far elements of the scene and in the sky than on close elements, but they all fit an expected absolute tolerance.
This fits well 3D-related applications.
\end{itemize}

For both alternatives, if multiple candidates are into the research ray or sphere, the closest matches are kept.
Research solid angle or radius criteria are simple but they can miss good matches if too small, or they can lead to erroneous matches if too large.
For this reason, we must compute results for a set of varied research angle or radius.
A highly accurate detector shall provide high scores for small solid angles or short radii: this allows to evaluate detection accuracy.
\\

In targeted applications (calibration, stereovision, visual odometry), the most important criteria is to extract matched points with a maximum 3D absolute location accuracy in order to bound the risks of outliers presence and to foster an accurate estimation with the optimisation process.\\
For this reason, we adopt the solution based on the research sphere for PFSeq benchmark, and vary the research radius to all the following distance tolerances:
5mm, 1cm, 2cm, 3cm, 4cm, 5cm, 6cm, 7cm, 8cm, 9cm, 10cm, and 15cm.

\paragraph{To evaluate the detector: Number of matches, Repeatability, accuracy\\}

Number of matches (matching pairs of detected points) and repeatability criteria only deal with the detector.
They need two shifted views of the same scene, and they are expressed accordingly to the chosen matching criteria and thresholds (test based on a distance threshold in the 3D projection described in previous paragraph).

\begin{itemize}
 \item The number of matches for a given distance threshold, ${\sharp \text{Matches}}_{threshold}$, is defined as the number of detection pairs that are closer than the threshold in the originating 3D scene to be assumed as matches.
In other words, it shows the potential number of matching points using this detector.

 \item The repeatability, for a given threshold, follows~\cite{Mikolajczyk05b} formulation and is defined by:
\begin{equation}
\text{Repeatability}_{threshold} = \frac{{\sharp \text{Matches}}_{threshold}}{\sharp \text{Detections}}
\end{equation}
with ${\sharp \text{Matches}}_{threshold}$ being the number of matches among the detections as defined above, and $\sharp$Detections the minimum number of detections between both images.
\\
Repeatability provides a measure of the maximal ratio of correct matches we could reach with evaluated detector.\\
However, this metric is not always meaningful, as it is possible to reach 100\% in ``degenerated'' cases: \textit{e.g.} when we consider all the overlaid points of the two images, or when there is only one detection in one of the images.\\
Knowing these particular cases, in realistic scenarios when we do not impose an extremely low or high number of detections, measured values shall reflect well the quality of a detector.

 \item The accuracy is not a direct metric, the use of fine and coarse distance thresholds through previous metrics illustrate how accurate can be the detectors.
\end{itemize}

It is essential to look at the repeatability together with the number of matches to fairly compare the algorithms depending on the order of magnitude of the number of matches they reach.
As long as features algorithms provide enough features for targeted application, repeatability is the most important.
However, in some cases, matches quantity is too low and we require more feature points at the cost of a lower still acceptable repeatability.
This can occur typically when relying on \textsc{ransac} loops to extract enough inliers for robust estimations.
The same, considering close repeatabilities, having lot more matches means the number of features detections is much larger in the same proportions, which is not desirable either as it complicates the matching process and adds computation time.

\paragraph{To evaluate the full detector / descriptor / matching pipeline: Number of correct (attempted) matches and Matching score\\}

To evaluate the descriptor and the matching steps, we do not have other viable choice than evaluating the full detection / description / matching pipeline.
The reasons are:
\begin{enumerate}
 \item A common trick to evaluate the detectors and descriptors is to match the description vectors with an unified process (often nearest neighbors, or ratio test of ~\cite{Lowe04}) and check the correctness of the matches using available dataset constraints, such as in~\cite{Mikolajczyk05a}, \cite{Moreels05} and \cite{Gauglitz11}.
In~\cite{Gauglitz11}, an extension of this idea is made possible to evaluate the descriptor only, from a sparse set of labelled true matching points.
While being simple, these evaluations are prone to be biased by the matching process.
\\
To avoid the matching algorithm contribution, we could imagine
a metric based on the distances between description vectors of matching and non-matching features.
But, to apply such principle, compared descriptors must have homogeneous and comparable distance measures, which is not the case: for example, SIFT descriptors comparison is made with the euclidean norm while FREAK descriptors comparison rely on Hamming distance.
For this reason, a descriptors-only evaluation can not be representative, we must evaluate the descriptor and matching algorithms together.

 \item In addition, we argue that a descriptor and matching evaluation together is of little interest for real applications because we need to take into account the variable detection results given by the available detectors, and because the descriptors are designed to be optimal with certain detectors only.
\end{enumerate}


The number of correct (attempted) matches and the matching score reflect the overall quality of the full detector / descriptor / matching pipeline.
Similarly to the number of matches and the repeatability, they rely on the chosen matching criteria to define whether the match is correct or not (test based on a 3D distance threshold).

\begin{itemize}
 \item The number of correct (attempted) matches for a distance threshold, $\sharp \text{Correct matches}_{threshold}$, is the quantity of attempted matches that are closer to each other than the 3D distance threshold.

 \item The matching score, for a given threshold, is defined by:
\begin{equation}
\text{Matching score}_{threshold} = \frac{\sharp \text{Correct matches}_{threshold}}{\sharp \text{Detections}}
\end{equation}
where $\sharp \text{Correct matches}_{threshold}$ is the number of correct (attempted) matches as defined above, and $\sharp$Detections the minimum number of detections between both images.\\
For PFSeq benchmark, we chose to use the matching score over the precision and other related metrics as a more global metric: in real use, we wish a maximum number of the detections to be matched and correctly matched.
Whatever the repeatability of the detector, if many detections can not be matched with used algorithms then the full mix detection / description / matching is not relevant.\\
In addition, it is easy to compare the matching score to the repeatability, as it can be seen as a subset of the ratio of the potential good matches reflected by the repeatability (with the property to be at max equal to the repeatability).\\
To compare matching score with repeatability, we could also imagine a ratio score between them, giving the equation $\frac{\sharp \text{Correct matches}_{threshold}}{{\sharp \text{Matches}}_{threshold}} \in [0;1]$.
However, we do not recommend using this new metric as we could have good score even in the case of very low repeatability and matching score, loosing the essential information to check: having a high number of good attempted matches among the detections.
\end{itemize}


At a first glance, a detector with a high repeatability could have more potential to allow then for a high matching score.
However, a slightly less good detector in terms of repeatability might be less prone to matching errors, in the case the description vectors would present less similarities and ambiguities hard to solve.
This means the matching score is a more global and more relevant metric than the repeatability to validate the algorithms choice for real application.

\paragraph{Draft of a spreading score metric\\}
\label{par:spreadingscore}

In this study, we miss a criteria that reflects the ability to get matches from the full scene, \textit{i.e.} from all textures, distances, and illumination conditions.
For general use, the matches and correct (attempted) matches shall be spread over the full image surface, instead of being concentrated in one or few zones.
For the future, we can imagine a way to compute a spreading score following one of these ideas:
\begin{itemize}
 \item Occupied cells ratio: as an adaptation of the uniformity criteria of~\cite{Li13} for circular images, we can define cells of the same area in polar coordinates to cover the image disc, in order to count the cells containing enough points.
 This is simple but relies on an arbitrary definition of the cells grid.

 \item Standard deviation of points coordinates: easy to compute but may be hard to read as we have two components in euclidean or in polar coordinates (that shall be normalized in some way to be comparable across images of various resolution).
 But standard deviation has the issue to favor points distributed along the limits of the coordinates ranges instead of an uniform coverage.

 \item Average of the distances between the points: it provides a single score, but the addition of a single point far from all the others can drastically change the score.
 Also, shall we compare every points with all the other points or only to their previous and following neighbours along each direction?
 Like the standard deviation, it might favor points distributed along the limits of coordinates ranges.

 \item A more complex metric: we want points 1) covering the full image disc, and 2) equally spaced.
 So, we shall consider 1) minimum and maximum coordinates (\textit{e.g.} the ratio between the ranges of feature points coordinates and image coordinates), and 2) a target distance between neighbour points depending on the number of points (\textit{e.g.} the points coordinates ranges divided by the number of points).
 If there is only one point, the score shall be minimal, then it shall increase.
 However, more points does not always means a better distribution.
 Following these constraints, 
 we can look for a formulation that behave a monotonic way and with no biases within special configurations.
\end{itemize}

Spreading of the features location is not explicitly handled in following results.
We will assume that the more matches and correct (attempted) matches we get, the better is the feature spreading.

\section{Comparison of standard planar features on PFSeq simulated fisheye images}
\label{sec:eval-simu}

In following evaluations, we use only the equisolid angle variant of PFSeq.
Plots are best viewed in color.

\subsection{Detectors presentation}
\label{ssec:detecteurs}

Many points and features detection algorithms have been proposed.
Following their nature and design objectives, they perform with more or less robustness to some conditions.
The most common identified transformations in the literature are~\cite{Mikolajczyk05a}:
\begin{itemize}
  \item blur (or sharpness loss that could also be caused by flare, low quality optics, etc),
  \item viewpoint changes,
  \item scale changes (or to zoom due to focal changes),
  \item rotations (orientations),
  \item illumination changes (bright with dark, lighting color changes, shadows, etc),
  \item or also the digital image encoding with lossy compression, such as JPEG, that may generate artifacts.
\end{itemize}
For omnidirectional vision and related applications, we require at least algorithms robust to:
\begin{itemize}
  \item scale changes, because elements size vary together with the viewpoint, and also because fisheye optical deformations make objects smaller or bigger when observed at its centre or at its edges;
  \item rotation/orientation changes, because a slight viewpoint change on circular images show elements with a notably different orientation.
\end{itemize}
Robustness to viewpoint changes is usually assured as long as scale and orientation changes robustness are assessed together, that is why we skip it in this list.

We summarize in table~\ref{tab:recap-detecteurs} the status of many points detection algorithms available in OpenCV, regarding the relevant robustnesses for omnidirectional vision.
The advantages and drawbacks of these algorithms are reported below:
\begin{enumerate}
  \item Harris method~\cite{Harris88}, very fast and seminal corner detector, many improvements on robustness have been proposed since.
  \item Shi-Tomasi method~\cite{Shi94}, (or Kanade-Tomasi) very fast corner detector.
  \item Shi-Tomasi method over Harris detector~\cite{Shi94}, improvement of Harris with Shi-Tomasi method, more stable for tracking.
  \item SIFT~\cite{Lowe04} (\textit{Scale Invariant Feature Transform}), relatively sensitive to noise, usually good and robust except for vegetation.
  \item FAST~\cite{Rosten06} (\textit{Features from Accelerated Segment Test}), very fast (can be applied in real-time video analysis), low robustness to noise.
  \item SURF~\cite{Bay06} (\textit{Speeded Up Robust Features}), often described as a faster SIFT, good quality over computation ratio, also bad for vegetation.
  \item CenSurE~\cite{Agrawal08} (\textit{Center Surround Extremas}), fast, compared to the fastest SURF variant rotation robustness is average but still better. Authors show that CenSurE shall be better than SIFT, SURF and Harris on visual odometry applications thanks to its points location accuracy.
  \item MSER~\cite{Nister08} (\textit{Maximally Stable Extremal Regions}), proposed implementation is linear in computation time. Robust to perspective changes, the way it detects points in the middle of regions makes it slightly robust to scale and rotation changes.
  \item ORB~\cite{Rublee11} (\textit{Oriented FAST and Rotated BRIEF}), fast, evolution of FAST algorithm with a pyramid processing to add scale changes robustness to the detector, robust to noise.
  \item BRISK~\cite{Leutenegger11} (\textit{Binary Robust Invariant Scalable Keypoints}), rather fast, detector based on AGAST~\cite{Mair10} (itself based on FAST), adds robustness to scale changes, more sensitive to noise than blob detectors, can work on vegetation.
\end{enumerate}

\begin{table}[htbp]
\caption{List of investigated detectors.
Boldened detectors are those who might be robust enough to scale and orientation changes, and that we compare in our evaluations on fisheye images.\label{tab:recap-detecteurs}}
\centering
  \begin{tabular}{l|c|c|c}
    \multicolumn{1}{c|}{Algorithm} & Feature type & Robust to scales & Robust to rotations \\
    \hline \hline
    Harris & corner & no & no \\
    \hline
    Shi-Tomasi & corner & no & no \\
    \hline
    Shi-Tomasi on Harris & corner & no & no \\
    \hline
    \textbf{SIFT} & blob & yes & yes \\
    \hline
    FAST & corner & no & no \\
    \hline
    SURF & blob & yes & average \\
    \hline
    \textbf{CenSurE} & blob\up{1} & yes & average+ \\
    \hline
    MSER & region & weak & weak \\
    \hline
    \textbf{ORB} & corner & yes & yes \\
    \hline
    \textbf{BRISK} & corner & yes & yes
    \\
    \multicolumn{4}{l}{\up{1}: CenSurE detects blobs whose the goal is to be more accurate than corners.}
  \end{tabular}
\end{table}

From this review, we chose to test the detection algorithms SIFT, CenSurE, ORB and BRISK.
\\

Evaluated detection algorithms are the OpenCV v2.4.8 implementations (\url{http://opencv.org/}).
They have different parameters that affect obtained results.
In order to provide an insightful investigation for the best method to use with fisheye images, we compare them by varying their parameters.

\paragraph{SIFT detector parameters are below:}
\begin{itemize}
  \item {\ttfamily nfeatures}, number of features to keep (the betters), possible to keep all when set to zero or to a negative value.
  \item {\ttfamily nOctaveLayers}, number of layers per octave, to be increasingly blurred.
  The optimum number of octaves (pyramid levels) is automatically fixed accordingly to the image resolution following the formula: $\log_2(\min(width,height))$.
  {\ttfamily nOctaveLayers} induces the number of difference of gaussians per octave ({\ttfamily nOctaveLayers – 1}). When a higher value is chosen then more steps are added, 
  that can sometimes improve accuracy and scale robustness. Original authors~\cite{Lowe04} get their best repeatability with {\ttfamily nOctaveLayers} value set to 3, which can be seen as a recommended value.
  \item {\ttfamily contrastThreshold}, contrast threshold to filter weak features in semi uniform regions (with low contrast). When increased then less features are output.
  \item {\ttfamily edgeThreshold}, threshold for the filter in charge of eliminating edges points. When increasing then the filter is more tolerant and more features are kept. It is based on the Hessian matrix.
  \item {\ttfamily sigma}, standard deviation of the initial gaussian blur cumulatively increased for each layer of the octaves. If input images are already rather blurry (could be a sharpness lack due to the sensor, the lens...) we can tune and reduce this value.
\end{itemize}

\paragraph{CenSurE parameters are below (called Star in OpenCV implementation):}
\begin{itemize}
  \item {\ttfamily maxSize}, maximum size of the features in terms of patterns, can take a fixed set of values between 1 and 128.
  \item {\ttfamily responseThreshold}, shall have a limited influence since it is updated when more extreme responses are measured.
  \item {\ttfamily lineThresholdProjected}, threshold to eliminate points lying on lines since they will be inaccurately localized. After experiments, the algorithm outputs 0 points if set too low.
  \item {\ttfamily lineThresholdBinarized}, complementary to {\ttfamily lineThresholdProjected}, must choose a value $\leq$ {\ttfamily lineThresholdProjected}. For our experiments we set it to {\ttfamily lineThresholdProjected} $- 2$.
  \item {\ttfamily suppressNonmaxSize}, same size for all the dimensions of the 3D window ($x$,$y$,scale) used for the non-maxima suppression, in order to keep only local extremas as detected points. The authors set this value to 3. The larger the window is the more features are filtered out, and remaining features have stronger response. This size must be an odd value.
\end{itemize}

\paragraph{ORB detector parameters are below:}
\begin{itemize}
  \item {\ttfamily nfeatures}, max number of detected features to keep (wanted number of features, no options to keep them all).
  \item {\ttfamily scaleFactor}, scale applied for each following pyramid level, must be $> 1$, an usual value for most algorithms is 2 but in this case it is recommended to reduce this scale to improve greatly the matching (no layers strategy within each octave as in SIFT). However, setting this too close to 1 requires much more computations to cover big scales.
  \item {\ttfamily nlevels}, number of levels of the pyramid representation. 
  \item {\ttfamily edgeThreshold}, width of the image contour where no features are detected, must be equal to {\ttfamily patchSize}.
  \item {\ttfamily firstLevel}, pyramid level to put 1:1 scale image, must be 0 in OpenCV implementation.
  \item {\ttfamily scoreType}, default value rely on Harris scores to sort the features, the other option is to use FAST but this is said to give less stable points.
  \item {\ttfamily patchSize}, $\geq 3$ and odd, patch size for the oriented BRIEF descriptor. Do not change this default value (31) as this will randomly generate a BRIEF pattern that will not be optimal.
\end{itemize}

\paragraph{BRISK detector parameters are below:}
\begin{itemize}
  \item {\ttfamily thresh}, detection threshold of the FAST/AGAST.
  \item {\ttfamily octaves}, number of levels of the pyramid representation, if set to 0 then the detector uses only the full scale.
  \item {\ttfamily patternScale}, scale of the pattern to sample point's surrounding, must be $\geq 1$.
\end{itemize}

Table~\ref{tab:params-detecteurs} summarizes the parameters of evaluated detectors, as well as their types and tested values.

\begin{table}[htbp]
\caption{Parameters of evaluated detectors.
Boldened parameters are those who are evaluated in this paper on fisheye images.
We vary them around the default values, assumed optimal according to authors tests.\label{tab:params-detecteurs}}
\centering
\begin{adjustbox}{width=\textwidth}
  \begin{tabular}{l|c|c|c|c}
    \multicolumn{1}{c|}{Algorithm} & Parameter & Type & Default & Tested values \\
    \hline \hline
    \multirow{5}{*}{SIFT} & {\ttfamily nfeatures} & int & 0 & 0 \\
    \cline{2-5}
    & {\ttfamily nOctaveLayers} & int & 3 & 3 \\
    \cline{2-5}
    & {\ttfamily \textbf{contrastThreshold}} & double & 0.04 & 0.02 , 0.04 , 0.08 \\
    \cline{2-5}
    & {\ttfamily \textbf{edgeThreshold}} & double & 10 & 5 , 10 , 20 \\
    \cline{2-5}
    & {\ttfamily \textbf{sigma}} & double & 1.6 & 1 , 1.6 , 3 \\
    \hline
    \multirow{5}{*}{CenSurE} & {\ttfamily \textbf{maxSize}} & int & 16 & 8 , 16 , 32 \\
    \cline{2-5}
    & {\ttfamily responseThreshold} & int & 30 & 30 \\
    \cline{2-5}
    & {\ttfamily \textbf{lineThresholdProjected}} & int & 10 & 10 , 20 \\
    \cline{2-5}
    & {\ttfamily lineThresholdBinarized} & int & 8 & $lineThresholdProjected - 2$ \\
    \cline{2-5}
    & {\ttfamily \textbf{suppressNonmaxSize}} & int & 5 & 3 , 5 , 7 \\
    \hline
    \multirow{7}{*}{ORB} & {\ttfamily \textbf{nfeatures}} & int & 500 & 500 , 1000 , 3000 \\
    \cline{2-5}
    & {\ttfamily scaleFactor} & float & 1.2 & $2^{{1}/{nOctaveLayers_{|SIFT}}}$ \up{(1)} \\
    \cline{2-5}
    & {\ttfamily nlevels} & int & 8 & $\log_2(\min(width,height)) \times nOctaveLayers_{|SIFT}$ \up{(2)} \\
    \cline{2-5}
    & {\ttfamily edgeThreshold} & int & 31 & $patchSize$ \\
    \cline{2-5}
    & {\ttfamily firstLevel} & int & 0 & 0 \\
    \cline{2-5}
    & {\ttfamily scoreType} & int & Harris score & Harris score \\ 
    \cline{2-5}
    & {\ttfamily patchSize} & int & 31 & 31 \\
    \hline
    \multirow{3}{*}{BRISK} & {\ttfamily \textbf{thresh}} & int & 30 & 10 , 20 , 30 , 60 \up{(3)} \\
    \cline{2-5}
    & {\ttfamily octaves} & int & 3 & $\log_2(\min(width,height))$ \up{(4)} \\
    \cline{2-5}
    & {\ttfamily \textbf{patternScale}} & float & 1 & 1 , 2 \up{(5)}
    \\
    \multicolumn{5}{l}{\up{(1)}: Scale factor tuned in order to imitate the effect of the SIFT's layers and octaves together.}
    \\
    \multicolumn{5}{l}{\up{(2)}: Pyramid levels to imitate the number of SIFT's layers and octaves together.}
    \\
    \multicolumn{5}{l}{\up{(3)}: Value 20 tested because FAST part of ORB detector uses a fixed threshold of 20.}
    \\
    \multicolumn{5}{l}{\up{(4)}: Inspired by the automatic setting of SIFT in order to get the same number of octaves.}
    \\
    \multicolumn{5}{l}{\up{(5)}: More a parameter for the feature descriptor, but we observed a slight effect on the detection results.}
  \end{tabular}
\end{adjustbox}
\end{table}

%

\subsection{Descriptors presentation and matching algorithms}
\label{ssec:descripteurs}

Descriptors are a set of distinctive values, usually represented as a vector, associated to the detected keypoints.
They can be computed with the same algorithm that does the detection, or with the use of an other descriptor algorithm.
Their role is to be able to compare them in order to find the similar points.
Typically, it is done with a distance measure in the descriptor vector space.
The matching of the image points with no additional knowledge can not be done without the descriptors.


Reported descriptors available in OpenCV are below: 
\begin{enumerate}
  \item SIFT~\cite{Lowe04} (\textit{Scale Invariant Feature Transform}), robust to perspective and viewpoint changes, this descriptor allows sometimes to provide too many matches (high false matches ratio), a reference in the field.
  \item SURF~\cite{Bay06} (\textit{Speeded Up Robust Features}), often described as a faster SIFT, good quality/computation ratio, approximate management of the orientations, there is also a \textit{upright} SURF (U-SURF) that drops all rotation robustness.
  \item BRIEF~\cite{Calonder10} (\textit{Binary Robust Independent Elementary Features}), very fast, robust to perspective changes but not explicitly designed for scale and orientation robustness, fit better very fast detectors such as CenSurE or FAST.
  \item ORB~\cite{Rublee11} (\textit{Oriented FAST and Rotated BRIEF}), fast, evolution of BRIEF that includes the rotation information in the descriptor, also robust to noise.
  \item BRISK~\cite{Leutenegger11} (\textit{Binary Robust Invariant Scalable Keypoints}), fast enough, inspired by the DAISY descriptor~\cite{Tola10} for the sampling method.
  \item FREAK~\cite{Alahi12} (\textit{Fast Retina Keypoint}), this aims to imitate the human visual system, adapts the spatial distribution proposed in DAISY to the retina model, fast and robust, the authors used it over BRISK detector.
\end{enumerate}

\begin{table}[htbp]
\caption{List of investigated descriptors.
Boldened descriptors are those who are robust enough to scale and orientation changes, and that we compare in our evaluations on fisheye images.\label{tab:recap-descripteurs}}
\centering
  \begin{tabular}{l|c|c}
    \multicolumn{1}{c|}{Algorithm} & Robust to scales & Robust to rotations \\
    \hline \hline
    \textbf{SIFT} & yes & yes \\
    \hline
    SURF & yes & average \\
    \hline
    BRIEF & no & no \\
    \hline
    \textbf{ORB} & yes & yes \\
    \hline
    \textbf{BRISK} & yes & yes \\
    \hline
    \textbf{FREAK} & yes & yes
  \end{tabular}
\end{table}

As for the descriptors, omnidirectional vision needs descriptors that are robust to scale and rotational changes.
Table~\ref{tab:recap-descripteurs} summarizes available descriptors in OpenCV and their robustnesses for aimed application.
\\
From this review, we chose to test the description algorithms SIFT, ORB, BRISK and FREAK.
\\

Evaluated description algorithms are the OpenCV v2.4.8 implementations (\url{http://opencv.org/}).
GLOH descriptor~\cite{Mikolajczyk05a}, despite being cited in section~\ref{ssec:bib-omni-features-sift} as a basis for the omnidirectional features descriptors of~\cite{Cruz11, Arican12}, is not present in our selection as it is not included in OpenCV.
We did not implement this descriptor, however, it would be worth to consider and try, as well as some newer works.
\\
Investigated descriptors have different parameters to evaluate in order to quantify their contribution to the matching of described point's quality.
One of the principal parameters is the number of octaves (pyramide levels) for the robustness to scale changes, already described for the algorithms doing both detection and description in section~\ref{ssec:detecteurs}.

\paragraph{SIFT descriptor parameters:\\}
Please refer to SIFT detector parameters in section~\ref{ssec:detecteurs}, they are the same.

\paragraph{ORB descriptor parameters are below:\\}
These are the additional ORB parameters used for the descriptor only, the others parameters remains the same as for the detector:
\begin{itemize}
  \item {\ttfamily WTA\_K}, how many points produce each element of the oriented BRIEF descriptor, can be 2, 3 or 4, matching step's Hamming distance dimensions has to be tuned accordingly.
\end{itemize}

\paragraph{BRISK descriptor parameters are below:\\}
The BRISK parameter more specific to the descriptor is the following:
\begin{itemize}
  \item {\ttfamily patternScale}, scale of the pattern to sample point's surrounding, must be $\geq 1$.
\end{itemize}

\paragraph{FREAK descriptor parameters are below:}
\begin{itemize}
  \item {\ttfamily orientationNormalized}, to normalize the orientation information.
  \item {\ttfamily scaleNormalized}, to normalize the scale information.
  \item {\ttfamily patternScale}, scale of the description pattern.
  \item {\ttfamily nOctaves}, number of octaves processed by the former features detectors, in order to adequately describe them.
  \item {\ttfamily selectedPairs}, option: possible to keep only the 512 indexes of the best pairs described from a greylevel image with the function FREAK::selectPairs.
\end{itemize}

The descriptors and their parameters are summarized in Table~\ref{tab:params-descripteurs}.

\begin{table}[htbp]
\caption{Parameters of evaluated descriptors.
Boldened parameters are those who are evaluated in this paper on fisheye images.
We vary them around the default values, assumed optimal according to authors tests.\label{tab:params-descripteurs}}
\centering
  \begin{tabular}{l|c|c|c|c}
    \multicolumn{1}{c|}{Algorithm} & Parameter & Type & Default & Tested values \\
    \hline \hline
    SIFT & \multicolumn{4}{c}{Follows SIFT detector settings of Table~\ref{tab:params-detecteurs} or defaults when using an other detector} \\ 
    \hline
    ORB & {\ttfamily \textbf{WTA\_K}} & int & 2 & 2 , 3 , 4 \\
    \hline
    BRISK & {\ttfamily \textbf{patternScale}} & float & 1 & 1 , 2 \\
    \hline
    \multirow{5}{*}{FREAK} & {\ttfamily orientationNormalized} & bool & true & true \\
    \cline{2-5}
    & {\ttfamily scaleNormalized} & bool & true & true \\
    \cline{2-5}
    & {\ttfamily \textbf{patternScale}} & float & 22 & 11 , 22 , 44 \\
    \cline{2-5}
    & {\ttfamily nOctaves} & int & 4 & $\log_2(\min(width,height))$ \up{(1)} \\
    \cline{2-5}
    & {\ttfamily selectedPairs} & vector<int> & N/A & N/A
    \\
    \multicolumn{5}{l}{\up{(1)}: Number of octaves deduced from the automatic setting also used in SIFT.}
  \end{tabular}
\end{table}


\paragraph{The 3 compared keypoints matching methods are below:}
\begin{enumerate}
  \item \textbf{RatioTest}~\cite{Lowe04} (proposed together with SIFT), based on the comparison of the two closest descriptors to check if the best one is clearly different from the others,
  \item \textbf{BruteForce}, exhaustive research method, shall provide the optimal solution at the cost of the longer computation time,
  \item and \textbf{FLANN}~\cite{Muja09} (Fast Library for Approximate Nearest Neighbors), quick method but leading to less correct matches than the others.
\end{enumerate}

To compare their descriptor with a distance measurement, SIFT and SURF descriptors rely on the euclidean distance.
The other descriptors presented in this report use the Hamming distance.
FLANN matching algorithm work only with euclidean distances, and hence only with SIFT and SURF descriptors.


\subsection{Optional polar transform to deal with the spherical projection}
\label{ssec:polar-transform}

\begin{figure}[htbp]
  \centering
  \includegraphics[width=90mm]{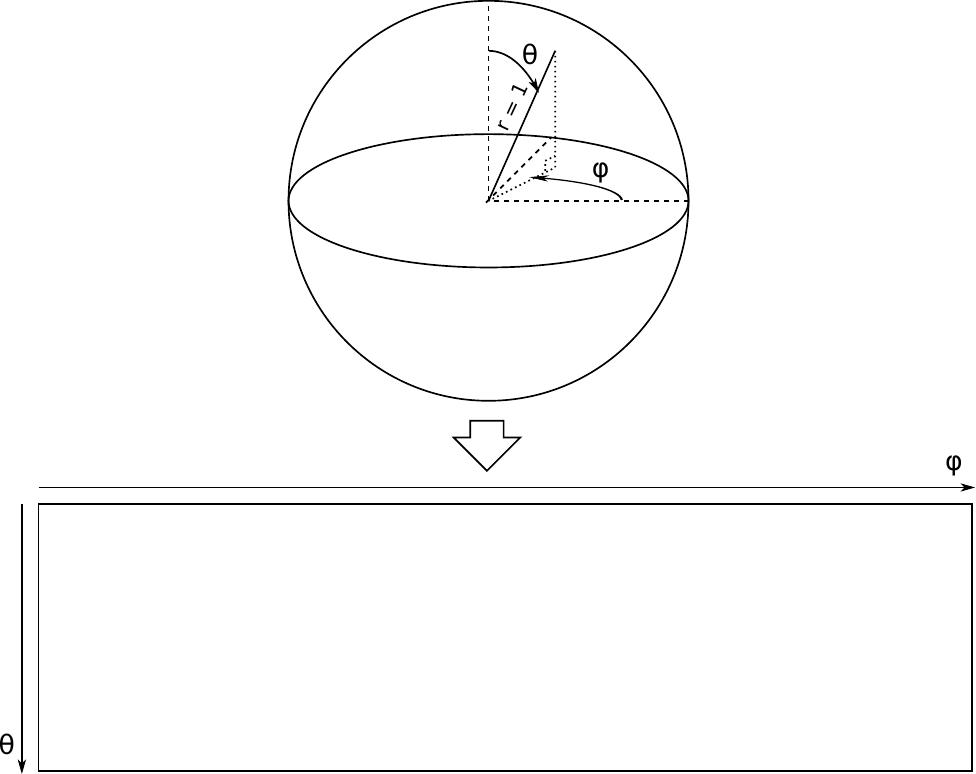} 
  \caption{
Transformation of the unit sphere image in a spherical angular coordinates matrix.
}
\label{fig:transfo-pol}
\end{figure}

\begin{figure}[htbp]
  \centering
  \subfloat[Before rectification (3296 \texttimes{} 2520 pixels).]{
  \includegraphics[height=35mm]{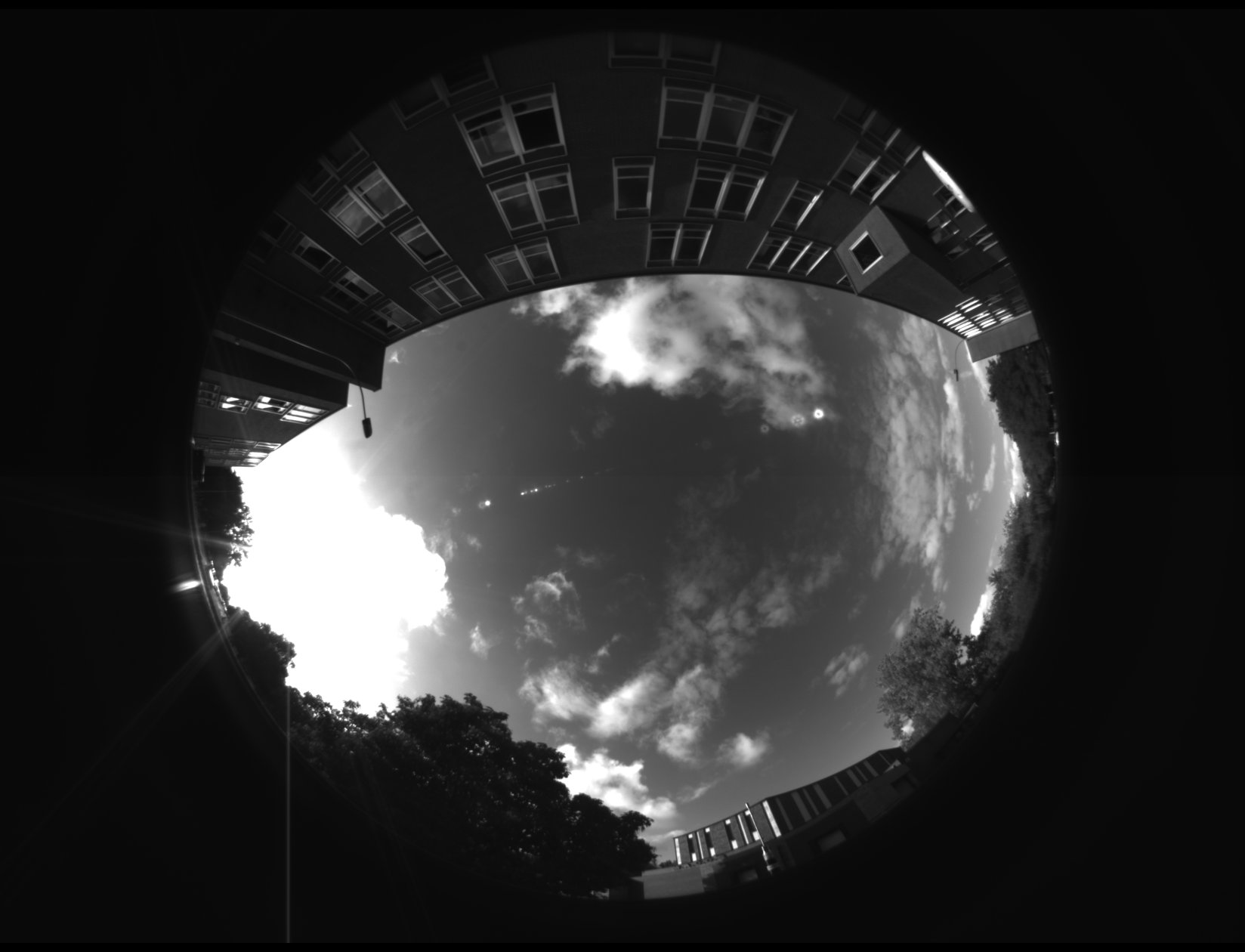} 
  }\\
  \subfloat[After polar rectification (7431 \texttimes{} 1859 pixels).]{
  \includegraphics[height=35mm]{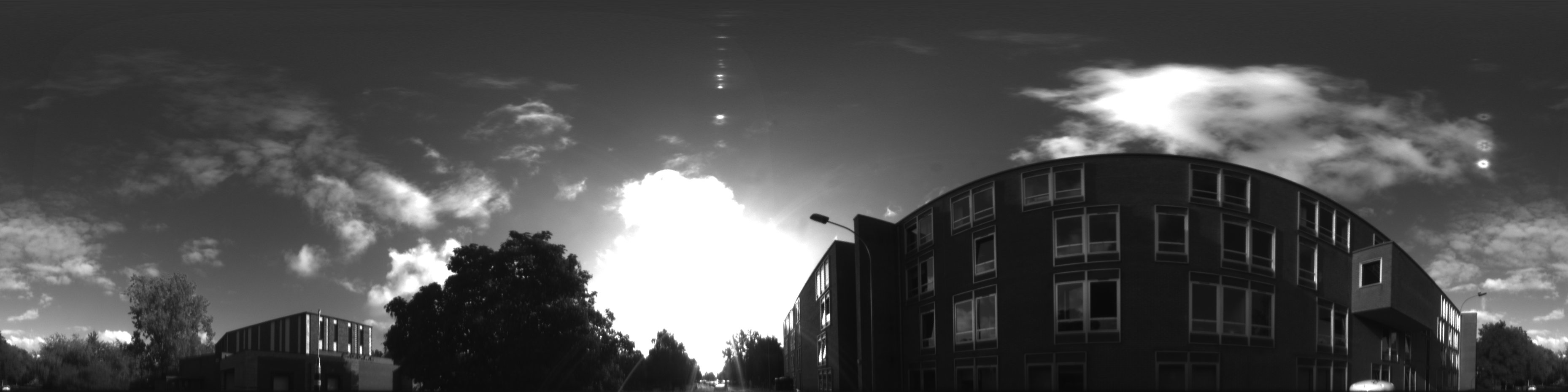}
  }
  \caption{
Example of polar rectification of a real fisheye image to a spherical coordinates frame.
}
\label{fig:transfo-pol-ex}
\end{figure}

The goal of this rectification is to mimic the features computation onto the surface of a unit sphere.
We can represent the spherical image in a rectangular matrix whose abscissa and ordinate give the spherical coordinates angles.
To proceed the rectification, projection function and parameters shall be known (or at least approximately known).
In our case, we assume an approximate field of view of 180\textdegree{} and do not apply interpolation, \textit{i.e.} we consider the intensity of the nearest neighbour for the transformed image to save computational cost and keep maximum sharpness.
Figure~\ref{fig:transfo-pol} illustrates how to build the angular coordinates matrix, and figure~\ref{fig:transfo-pol-ex} provides an example before and after the polar rectification.
Proposed implementation adapts the angular resolution to keep at least a representation of every pixel of the original image disc, leading to larger images.
For all the detection and description methods, we will proceed to the evaluation of their direct use on the original omnidirectional image and on the transformed image in spherical coordinates.

\subsection{Detectors evaluation}
\label{ssec:eval-det}


Reported detectors settings in following figures and tables provide the best result in number of matches or in repeatability, at least once considering each PFSeq distance tolerance and evaluation image, as described in section~\ref{ssec:pfseq}.
Reporting all the possible alternatives would drown the interesting results into the mass of weaker results (in total, we proceed 112 settings).
For this reason, some detectors from section~\ref{ssec:detecteurs} will not be reported here.
\\




\def\inputchemin#1{
\input{./2014-04-07-plot-v2/#1} 
}

\input{./2014-04-07-plot-v2/mix_detecteurs_0_nbcorresp_best_gphe.tex} 

\input{./2014-04-07-plot-v2/mix_detecteurs_0_repetabilite_best_gphe.tex} 

\input{./2014-04-07-plot-v2/mix_detecteurs_0_nbcorresp_best_tab.tex} 

\input{./2014-04-07-plot-v2/mix_detecteurs_0_repetabilite_best_tab.tex} 

\input{./2014-04-07-plot-v2/mix_detecteurs_40_nbcorresp_best_gphe.tex} 

\input{./2014-04-07-plot-v2/mix_detecteurs_40_repetabilite_best_gphe.tex} 

\input{./2014-04-07-plot-v2/mix_detecteurs_40_nbcorresp_best_tab.tex} 

\input{./2014-04-07-plot-v2/mix_detecteurs_40_repetabilite_best_tab.tex} 

\input{./2014-04-07-plot-v2/mix_detecteurs_24_nbcorresp_best_gphe.tex} 

\input{./2014-04-07-plot-v2/mix_detecteurs_24_repetabilite_best_gphe.tex} 

\input{./2014-04-07-plot-v2/mix_detecteurs_24_nbcorresp_best_tab.tex} 

\input{./2014-04-07-plot-v2/mix_detecteurs_24_repetabilite_best_tab.tex} 

From this evaluation, twelve different detectors go ahead.
Blob detectors such as SIFT and CenSurE stand out from corner detectors ORB and BRISK, they even not appear in the highlighted detectors setups subset in our figures and tables.
SIFT is the more prominent with 9 of these best detectors setups, CenSurE being represented only 3 times.
Also, only 4 of these remained setups do use the polar rectification, and they rely on SIFT.\\
Usually, SIFT provides lot more potential matches than CenSurE, tables~\ref{tab:nombre-de-correspondants-image-0}, \ref{tab:nombre-de-correspondants-image-40}, \ref{tab:nombre-de-correspondants-image-24}, while also reaching higher repeatability (except in few cases and only for higher distance tolerance), tables~\ref{tab:repetabilite-image-0}, \ref{tab:repetabilite-image-40}, \ref{tab:repetabilite-image-24}.
In addition, CenSurE does not provide enough matches with short distance tolerances (it provides less accurate points) and is even totally out for pair 24 ``translation and rotation motion'' (quasi no potential matches), the most challenging sample with rotation and motion blur.
This makes CenSurE hardly usable in practice with acceptable matches accuracy.
We can explain the disagreement between our accuracy results and expected accuracy accordingly to CenSurE article~\cite{Agrawal08}, by the fact our sequences are very challenging with strong fisheye distortions and motion blur.\\
Regarding the polar rectification, SIFT with or without rectification are usually equivalent in terms of number of detections and repeatability (tuning its parameters), but SIFT without polar rectification has the advantage to usually reach a better repeatability at tighter distance tolerances, hence it is more accurate (it saves also rectification computing cost).
The exception might be the hardest scene, pair 24 ``translation and rotation motion'', where SIFT with polar rectification can be the best with both criteria, tables~\ref{tab:nombre-de-correspondants-image-24} and \ref{tab:repetabilite-image-24}.
\\

Consequently, for targeted applications, the usual best compromise in terms of number of matches and repeatability is the SIFT detector, and we can distinguish 2 settings groups:
\begin{enumerate}
 \item setups SIFT \emph{nopol 0.02 20 1.6} (preferred for its slightly better repeatability) and \emph{nopol 0.02 20 1.0}, they always provide a large number of matches, that is an advantage for the more challenging scene to leverage a lower repeatability, but, getting thousands of potential matches can also be a paradox in the extreme cases as it shall imply too many detections (especially if repeatability is low), that can be a disadvantage for description and matching,
 \item setups SIFT \emph{nopol 0.04 5 1.6} (preferred as it is slightly better even if less robust to the hardest scene) and \emph{nopol 0.04 5 3.0}, more accurate as they usually have a better repeatability for shorter distance thresholds and equivalent for the other thresholds, however, while providing a good quantity of matches on most of the scenes, they are very few on the most challenging case, meaning the subsequent algorithms have more chances to fail.
\end{enumerate}
Between them, we can recommend the SIFT \emph{nopol 0.02 20 1.6} for general use in all conditions or SIFT \emph{nopol 0.04 5 1.6} when conditions are not so challenging.
SIFT \emph{nopol 0.04 5 1.6} has usually a higher repeatability even if lower number of matches, that shall provide a lower ratio of outliers for \textsc{ransac}-type usage.
An other option to explicitly deal with very challenging conditions is to use two detectors in parallel, combining one of these detectors with SIFT \emph{pol 0.02 5 3.0}, which is the only setup clearly providing the best repeatability and accuracy on challenging scenes and an acceptable number of matches, and which is also overall acceptable on the others samples among the variants with polar rectification.

\subsection{Detector / descriptor / matching triplet evaluation}
\label{ssec:eval-desc-match}


Reported combinations of detector / descriptor / matching settings in following figures and tables provide the best result in number of correct matches or in matching score, at least once considering each PFSeq distance tolerance and evaluation image, as described in section~\ref{ssec:pfseq}.
For this reason, some detectors from section~\ref{ssec:detecteurs} and some detectors and matching algorithms from section~\ref{ssec:descripteurs} will not be reported here  (in total, we proceed 2096 settings).
\\

\input{./2014-04-07-plot-v2/mix_det-desc-match_SIFT_0_bonscorresp_best_gphe.tex} 

\input{./2014-04-07-plot-v2/mix_det-desc-match_SIFT_0_mscore_best_gphe.tex} 

\input{./2014-04-07-plot-v2/mix_det-desc-match_SIFT_0_bonscorresp_best_tab.tex} 

\input{./2014-04-07-plot-v2/mix_det-desc-match_SIFT_0_mscore_best_tab.tex} 

\input{./2014-04-07-plot-v2/mix_det-desc-match_SIFT_40_bonscorresp_best_gphe.tex} 

\input{./2014-04-07-plot-v2/mix_det-desc-match_SIFT_40_mscore_best_gphe.tex} 

\input{./2014-04-07-plot-v2/mix_det-desc-match_SIFT_40_bonscorresp_best_tab.tex} 

\input{./2014-04-07-plot-v2/mix_det-desc-match_SIFT_40_mscore_best_tab.tex} 

\input{./2014-04-07-plot-v2/mix_det-desc-match_SIFT_24_bonscorresp_best_gphe.tex} 

\input{./2014-04-07-plot-v2/mix_det-desc-match_SIFT_24_mscore_best_gphe.tex} 

\input{./2014-04-07-plot-v2/mix_det-desc-match_SIFT_24_bonscorresp_best_tab.tex} 

\input{./2014-04-07-plot-v2/mix_det-desc-match_SIFT_24_mscore_best_tab.tex} 




\input{./2014-04-07-plot-v2/mix_det-desc-match_CenSurE_0_bonscorresp_best_gphe.tex} 

\input{./2014-04-07-plot-v2/mix_det-desc-match_CenSurE_0_mscore_best_gphe.tex} 

\input{./2014-04-07-plot-v2/mix_det-desc-match_CenSurE_0_bonscorresp_best_tab.tex} 

\input{./2014-04-07-plot-v2/mix_det-desc-match_CenSurE_0_mscore_best_tab.tex} 

\input{./2014-04-07-plot-v2/mix_det-desc-match_CenSurE_40_bonscorresp_best_gphe.tex} 

\input{./2014-04-07-plot-v2/mix_det-desc-match_CenSurE_40_mscore_best_gphe.tex} 

\input{./2014-04-07-plot-v2/mix_det-desc-match_CenSurE_40_bonscorresp_best_tab.tex} 

\input{./2014-04-07-plot-v2/mix_det-desc-match_CenSurE_40_mscore_best_tab.tex} 

\input{./2014-04-07-plot-v2/mix_det-desc-match_CenSurE_24_bonscorresp_best_gphe.tex} 

\input{./2014-04-07-plot-v2/mix_det-desc-match_CenSurE_24_bonscorresp_best_tab.tex} 

CenSurE fails for all the setups on pair 24 ``translation and rotation motion'', with both rotation and motion blur, there are too few correct matches to be exploitable, as shown in figure~\ref{fig:nombre-de-bons-correspondants-image-24-censure} and table~\ref{tab:nombre-de-bons-correspondants-image-24-censure}.
For this reason we skip the figure and table showing its matching score.
\\

Similarly to the detectors evaluation in section~\ref{ssec:eval-det}, only CenSurE and SIFT detectors can provide good results.
There are eleven different setups based on SIFT detector, 8 together with SIFT descriptor and the 3 others with FREAK descriptor.
There are seven different setups based on CenSurE detector, with similar proportions of all the evaluated descriptors (SIFT, ORB, BRISK, FREAK).
\\
From reported results, the clear best matching method is BruteForce.
BruteForce is the only matching method remaining with SIFT detector.
With CenSurE detector, 1 setup rely on RatioTest and 2 on FLANN, but they have less correct (attempted) matches and lower matching scores than the setups using BruteForce.
This result shows the need for optimal matching methods at lower cost, and compatible with any kind of distance function (FLANN being compatible with euclidean distance only).
Further evaluations and researches are required in this direction.
\\
Compared to SIFT detector setups, CenSurE detector setups have lot less correct (attempted) matches for all the scenes (even might be not enough), tables~\ref{tab:nombre-de-bons-correspondants-image-0-sift}, \ref{tab:nombre-de-bons-correspondants-image-40-sift}, \ref{tab:nombre-de-bons-correspondants-image-24-sift}, \ref{tab:nombre-de-bons-correspondants-image-0-censure}, \ref{tab:nombre-de-bons-correspondants-image-40-censure}, \ref{tab:nombre-de-bons-correspondants-image-24-censure}, for a slightly lower or equivalent overall matching score, tables~\ref{tab:matching-score-image-0-sift}, \ref{tab:matching-score-image-40-sift}, \ref{tab:matching-score-image-24-sift}, \ref{tab:matching-score-image-0-censure}, \ref{tab:matching-score-image-40-censure}. 
In addition, CenSurE totally fails for pair 24 ``translation and rotation motion'', the most challenging case.
Again, SIFT detector shows to be the best feature detection algorithm.
\\
The best setup to choose must be based on SIFT detector, but tuning at best the parameters is not straightforward.
We observe that setups providing the most correct (attempted) matches have usually a lower matching score, meaning that even more points have been detected but wrongly matched, which is less desirable for targeted applications (higher proportion of outliers).
For scenes 0 ``translation'' and 40 ``translation motion'', the best compromise seems to use a SIFT detector with \emph{nopol 0.04 5.0 ... ... BruteForce} setup (3 close options are shown). 
However, for the most challenging scene 24 ``translation and rotation motion'', they have too few correct (attempted) matches for a good optimization as for calibration (less than 100 correct matches).
Other setups have up to 6 times more correct (attempted) matches, and close matching score.
The best overall in this case is SIFT with \emph{pol 0.02 20.0 3.0 SIFT BruteForce}, closely followed by SIFT with \emph{nopol 0.02 20.0 1.6 SIFT BruteForce} being better for the other scenes, and with the economy of the polar transformation (even if SIFT with \emph{pol 0.02 20.0 3.0 SIFT BruteForce} is the best of the setups with polar rectification for all the scenes).
\\

For general use and good enough robustness to challenging scenes, we would finally recommend SIFT with \emph{nopol 0.02 20.0 1.6 SIFT BruteForce}, for which we can expect a matching score from approximately 35\% for easy scenes to 15\% for challenging scenes with an accuracy tolerance no greater than 4cm.
A viable alternative can be to hold two setups concurrently for maximum quality: SIFT with \emph{nopol 0.04 5.0 1.6 SIFT BruteForce} the best for less challenging scenes where we can expect a matching score of 50\%, and SIFT with \emph{pol 0.02 20.0 3.0 SIFT BruteForce} the best for most challenging scenes where we can expect a matching score of 20\% for similar tolerance.\\
Note that while SIFT dominates both the detection and description, the FREAK descriptor reveals to be also able to provide among the better results, for example with the setup SIFT \emph{nopol 0.04 5.0 1.6 FREAK 44.0 BruteForce}, close to SIFT with \emph{nopol 0.04 5.0 1.6 SIFT BruteForce}. 
An interesting idea could be to use as detector the SIFT \emph{nopol 0.04 5.0 1.6} together with 2 descriptors SIFT and FREAK \emph{44.0}, the former performing better for standard scenes and the latter slightly better for the most challenging cases where we can expect a matching score close to 30\% (reducing the attractiveness of using SIFT with \emph{pol 0.02 20.0 3.0 SIFT BruteForce} even if it provides more correct matches).
\\


As can be seen in section~\ref{ssec:eval-det}, SIFT with \emph{nopol 0.02 20 1.6} and with \emph{nopol 0.04 5.0 1.6} are the two most promising detector setups.
This fits well with the conclusion of the detection / description / matching evaluation.
However, we do not find again a better specialized detector option for the most challenging case such as SIFT with \emph{pol 0.02 5 3.0} in the detector only results.
Finally, our results show there is no advantage to use the polar rectification.
\\

To measure the drop from potential matches to correct attempted matches, we can compare the matching score to the repeatability of these detectors.
If description and matching works ideally, the matching score shall reach the repeatability value.
To this aim, we synthesize these data in table~\ref{tab:repeatability-vs-mscore} (reported from tables~\ref{tab:repetabilite-image-0}, \ref{tab:repetabilite-image-40}, \ref{tab:repetabilite-image-24} for the repeatability and tables~\ref{tab:matching-score-image-0-sift}, \ref{tab:matching-score-image-40-sift}, \ref{tab:matching-score-image-24-sift} for the matching score).\\
The detector with higher repeatability usually reaches better matching score, tendencies stay the same.
The drop from repeatability to matching score increases generally with the distance tolerance.
This is especially true with SIFT with \emph{nopol 0.02 20.0 1.6 SIFT BruteForce}, meaning that a higher tolerance adds more false (attempted) matches than correct (attempted) matches.
The behaviour of SIFT with \emph{nopol 0.04 5.0 1.6} variants is better as the matching score remains stable between distance tolerance from approximately 0.04m to 0.1m.
For the setups based on SIFT with \emph{nopol 0.04 5.0 1.6} detector, the matching scores based on SIFT descriptor are better only for less-challenging scenes, FREAK \emph{44.0} being better for the most challenging scene 24 ``translation and rotation motion''.\\
Strongest drops from repeatability to matching score are gotten for the scenes 0 ``translation`` and 24 ''translation and rotation motion``, while the scene 40 ''translation motion`` gets the overall lowest drops.
The exception is with SIFT with \emph{nopol 0.04 5.0 1.6 FREAK 44.0} for which the drops are the lowest for the hardest scene 24 ''translation and rotation motion``, and for which matching scores with the scene 40 ''translation motion`` can be better than for the easy scene 0 ''translation``, while having lesser good repeatabilities.
\\
Hence, SIFT with \emph{nopol 0.04 5.0 1.6} is the most stable detector in terms of expected results when applied to automatic description and matching.
It is also interesting to note that the FREAK descriptor is able to surpass SIFT descriptor for scenes strongly affected by blur.
FREAK might be optimal for blurry content, it would be worth to try to artificially add blur to the static scene before detection and/or description to compare the matching scores.
Using more than one descriptor, or designing a new descriptor fusing strengths of SIFT and FREAK, makes sense to be more robust to all the scenarios.

\begin{table}[htbp]
\caption{Repeatability and matching score of the best SIFT setups.
Repeatabilities are written in italics.
Matching scores are given for BruteForce matching.
Values into parenthesis are the decreases from the repeatability of the detector to the matching score.
\label{tab:repeatability-vs-mscore}}
\centering
\begin{adjustbox}{width=1\textwidth}
\begin{tblr}{l |c|c|[dashed]c*{2}{|c} |[1pt]c|c|[dashed]c*{2}{|c} |[1pt]c|c|[dashed]c*{2}{|c}} 
\SetCell[c=1]{c} \rotatebox{90}{Distance tolerance (m)} &
\rotatebox{90}{\textit{Repeatability SIFT nopol 0.02 20.0 1.6}} & \rotatebox{90}{Matching score SIFT nopol 0.02 20.0 1.6 SIFT} & \rotatebox{90}{\textit{Repeatability SIFT nopol 0.04 5.0 1.6}} & \rotatebox{90}{Matching score SIFT nopol 0.04 5.0 1.6 FREAK 44.0} & \rotatebox{90}{Matching score SIFT nopol 0.04 5.0 1.6 SIFT} &
\rotatebox{90}{\textit{Repeatability SIFT nopol 0.02 20.0 1.6}} & \rotatebox{90}{Matching score SIFT nopol 0.02 20.0 1.6 SIFT} & \rotatebox{90}{\textit{Repeatability SIFT nopol 0.04 5.0 1.6}} & \rotatebox{90}{Matching score SIFT nopol 0.04 5.0 1.6 FREAK 44.0} & \rotatebox{90}{Matching score SIFT nopol 0.04 5.0 1.6 SIFT} &
\rotatebox{90}{\textit{Repeatability SIFT nopol 0.02 20.0 1.6}} & \rotatebox{90}{Matching score SIFT nopol 0.02 20.0 1.6 SIFT} & \rotatebox{90}{\textit{Repeatability SIFT nopol 0.04 5.0 1.6}} & \rotatebox{90}{Matching score SIFT nopol 0.04 5.0 1.6 FREAK 44.0} & \rotatebox{90}{Matching score SIFT nopol 0.04 5.0 1.6 SIFT} \\
\hline
\SetCell[c=1]{r} Seq. & \SetCell[c=5]{c} 0 {``translation''} & & & & & \SetCell[c=5]{c} 40 {``translation motion''} & & & & & \SetCell[c=5]{c} 24 {``translation and rotation motion''} & & & & \\
\hline
\hline
0.005 & \textit{0.07} & 0.06 (0.01) & \textit{0.10} & 0.09 (0.01) & 0.09 (0.01) & \textit{0.03} & 0.03 (0.00) & \textit{0.07} & 0.06 (0.01) & 0.06 (0.01) & \textit{0.02} & 0.01 (0.01) & \textit{0.03} & 0.03 (0.00) & 0.01 (0.02) \\
\hline
0.01 & \textit{0.19} & 0.16 (0.03) & \textit{0.26} & 0.21 (0.05) & 0.24 (0.02) & \textit{0.13} & 0.12 (0.01) & \textit{0.23} & 0.22 (0.01) & 0.21 (0.02) & \textit{0.07} & 0.05 (0.02) & \textit{0.10} & 0.09 (0.01) & 0.08 (0.02) \\
\hline
0.02 & \textit{0.34} & 0.27 (0.07) & \textit{0.44} & 0.35 (0.09) & 0.39 (0.05) & \textit{0.29} & 0.26 (0.03) & \textit{0.43} & 0.40 (0.03) & 0.40 (0.03) & \textit{0.15} & 0.10 (0.05) & \textit{0.21} & 0.20 (0.01) & 0.15 (0.06) \\
\hline
0.03 & \textit{0.43} & 0.33 (0.10) & \textit{0.53} & 0.42 (0.11) & 0.46 (0.07) & \textit{0.38} & 0.34 (0.04) & \textit{0.51} & 0.46 (0.05) & 0.47 (0.04) & \textit{0.21} & 0.14 (0.07) & \textit{0.29} & 0.27 (0.02) & 0.20 (0.09) \\
\hline
0.04 & \textit{0.49} & 0.36 (0.13) & \textit{0.58} & 0.46 (0.12) & 0.51 (0.07) & \textit{0.43} & 0.38 (0.05) & \textit{0.54} & 0.50 (0.04) & 0.51 (0.03) & \textit{0.26} & 0.17 (0.09) & \textit{0.35} & 0.31 (0.04) & 0.25 (0.10) \\
\hline
0.05 & \textit{0.53} & 0.39 (0.14) & \textit{0.61} & 0.48 (0.13) & 0.53 (0.08) & \textit{0.46} & 0.40 (0.06) & \textit{0.56} & 0.51 (0.05) & 0.52 (0.04) & \textit{0.30} & 0.19 (0.11) & \textit{0.38} & 0.34 (0.04) & 0.27 (0.11) \\
\hline
0.06 & \textit{0.56} & 0.40 (0.16) & \textit{0.63} & 0.49 (0.14) & 0.54 (0.09) & \textit{0.48} & 0.41 (0.07) & \textit{0.57} & 0.52 (0.05) & 0.53 (0.04) & \textit{0.33} & 0.21 (0.11) & \textit{0.40} & 0.36 (0.04) & 0.29 (0.11) \\
\hline
0.07 & \textit{0.58} & 0.41 (0.17) & \textit{0.64} & 0.50 (0.14) & 0.55 (0.09) & \textit{0.50} & 0.43 (0.07) & \textit{0.59} & 0.53 (0.06) & 0.54 (0.05) & \textit{0.36} & 0.23 (0.13) & \textit{0.40} & 0.36 (0.04) & 0.29 (0.11) \\
\hline
0.08 & \textit{0.60} & 0.42 (0.18) & \textit{0.65} & 0.51 (0.14) & 0.56 (0.09) & \textit{0.51} & 0.44 (0.07) & \textit{0.60} & 0.54 (0.06) & 0.55 (0.05) & \textit{0.38} & 0.24 (0.14) & \textit{0.42} & 0.37 (0.05) & 0.31 (0.11) \\
\hline
0.09 & \textit{0.62} & 0.43 (0.19) & \textit{0.66} & 0.52 (0.14) & 0.57 (0.09) & \textit{0.52} & 0.44 (0.08) & \textit{0.61} & 0.54 (0.07) & 0.56 (0.05) & \textit{0.40} & 0.25 (0.15) & \textit{0.44} & 0.40 (0.04) & 0.33 (0.11) \\
\hline
0.1 & \textit{0.64} & 0.43 (0.21) & \textit{0.67} & 0.52 (0.15) & 0.57 (0.10) & \textit{0.53} & 0.45 (0.08) & \textit{0.61} & 0.55 (0.06) & 0.56 (0.05) & \textit{0.42} & 0.26 (0.16) & \textit{0.45} & 0.40 (0.05) & 0.34 (0.11) \\
\hline
0.15 & \textit{0.70} & 0.45 (0.25) & \textit{0.70} & 0.53 (0.17) & 0.59 (0.11) & \textit{0.59} & 0.48 (0.11) & \textit{0.65} & 0.57 (0.08) & 0.59 (0.06) & \textit{0.51} & 0.30 (0.21) & \textit{0.53} & 0.47 (0.06) & 0.41 (0.12) \\
\end{tblr}
\end{adjustbox}
\end{table}

\section{Evaluation with real images in the case of self-calibration application}
\label{sec:eval-reel}

In order to validate the evaluation results from simulated images, we propose to measure the self-calibration stability for various runs on real images.
The problem with a real camera is that we do not a have ground truth for the calibration parameters.
Although it would be interesting, we did not perform a strong calibration to get reference parameters to compare our estimations with.
However, we can check how each setup agrees with each other on their final estimations.\\
We use a \textsc{ransac}-based algorithm to estimate the calibration parameters, which is a probabilistic algorithm.
So, the goal is to conduct estimations a large number of iterations to check their stability.
To do so, we ran 1000 iterations of the calibration for each parameter sets of the selected detectors / descriptors / matching algorithms.
From the discussions of section~\ref{ssec:eval-desc-match} and the matching scores of the best reported setups, we can reasonably expect to extract approximately 35\% of correct matches among the attempted matches with most of the scenes (the matching score being given relatively to the number of detections, it is always lower than this ratio).
This way, following our work~\cite{Moreau16}, \textsc{ransac} maximum number of iterations is set to 50000 with some margin, and it can converge and stop before.
\\
We assumed that the better the conditions are, the more accurate and more stable the algorithm to be.
However, we noted that detector / descriptor / matching combinations with very few points usually provide totally wrong results, but with a very high stability.
Consequently, we conduct this stability evaluation only for the algorithms setups highlighted within PFSeq simulated images experiments of section~\ref{ssec:eval-desc-match}.
\\

\begin{figure}[htbp]
  \centering
  \subfloat[Pair 7100 ``translation''.]{
  \includegraphics[width=45mm]{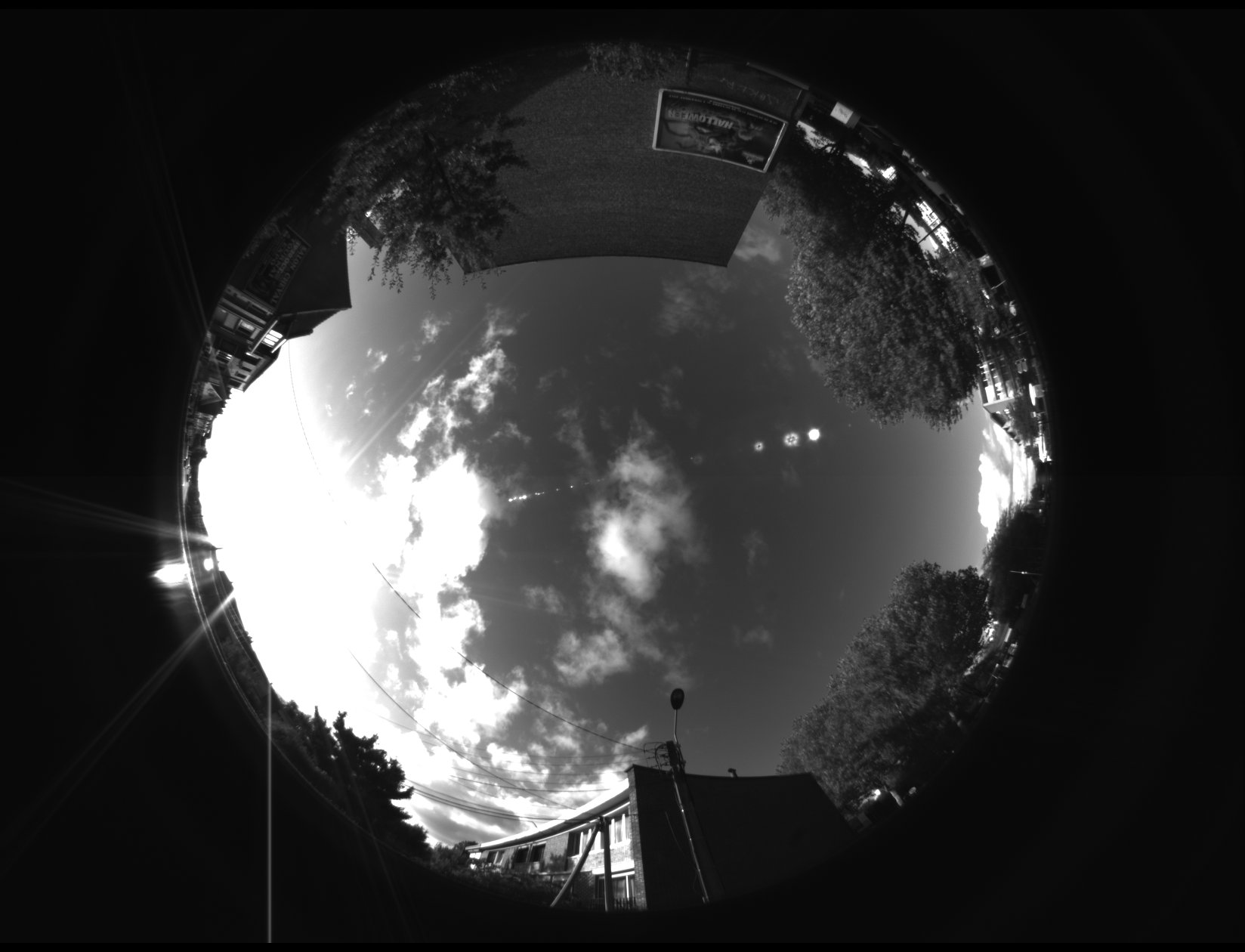}
  }
  \subfloat[Pair 10627 ``translation motion''.]{
  \includegraphics[width=45mm]{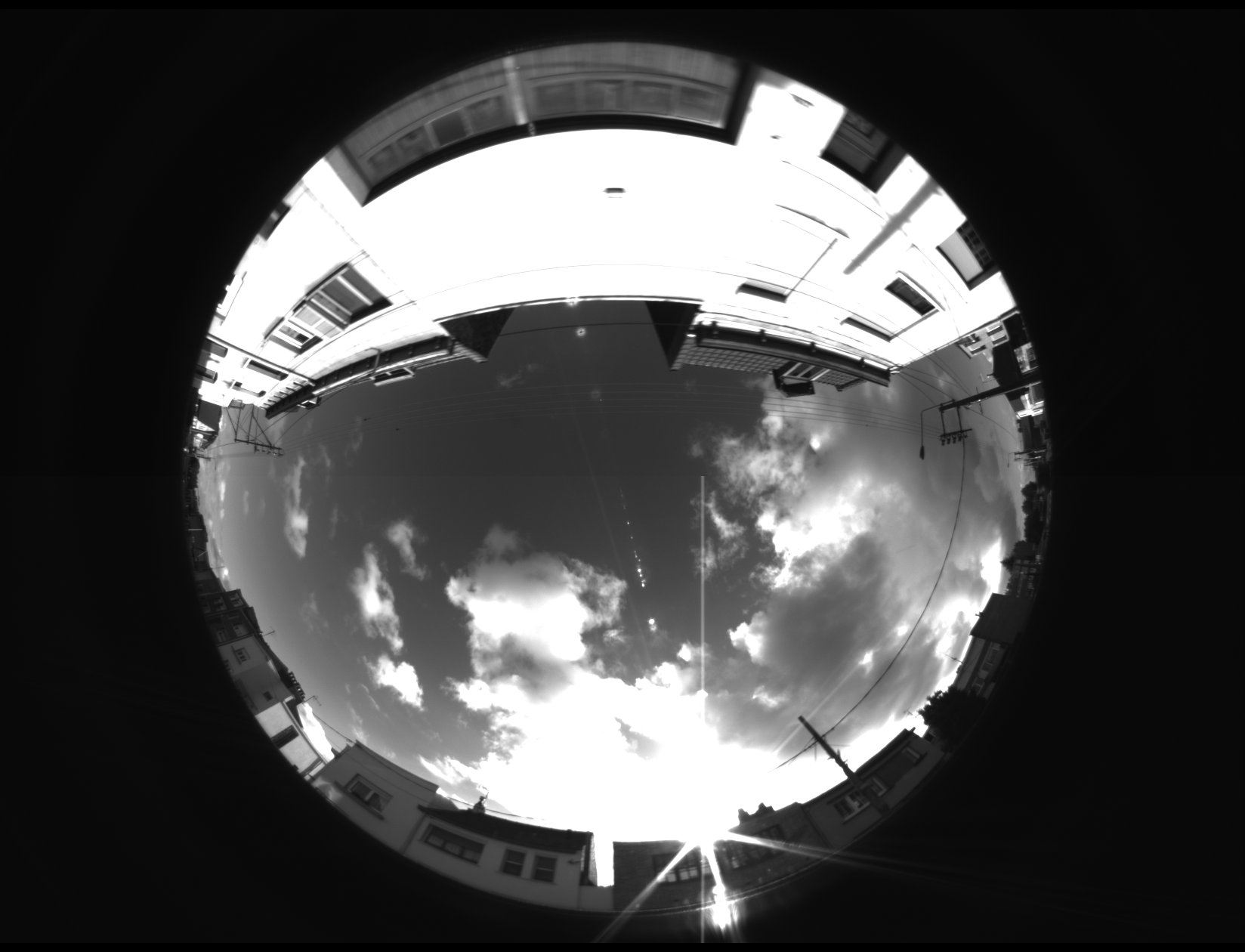}
  }
  \subfloat[Pair 11048 ``translation and rotation motion''.]{
  \includegraphics[width=45mm]{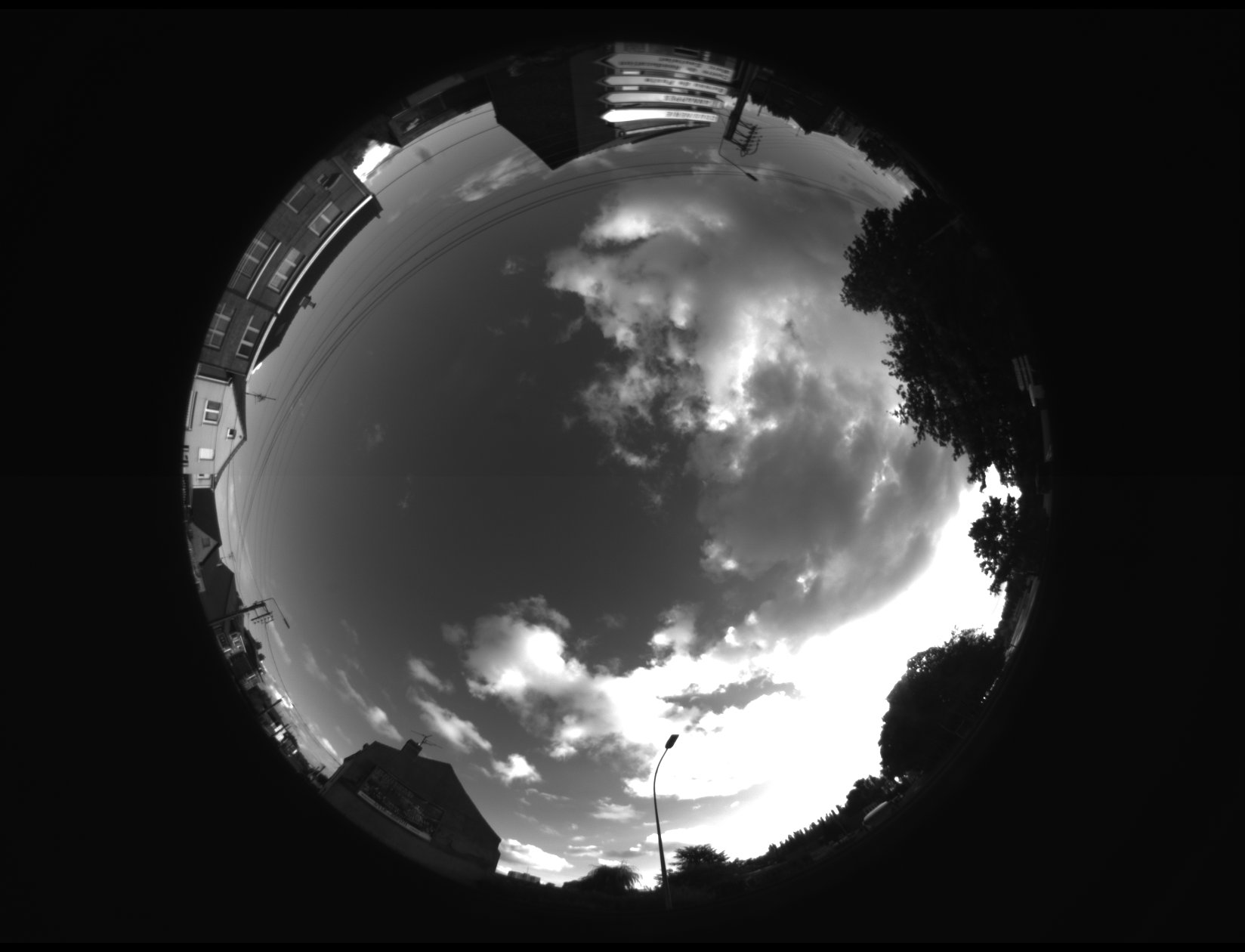}
  }
  \caption{
Real images used for the stability evaluation.
They represent the conditions met when moving the ego-vehicle.
}
\label{fig:3imagesrelles}
\end{figure}

We selected three real scenes acquired from a fixed stereo setup (meaning that the parameters are the same for all of them), figure~\ref{fig:3imagesrelles}, that mimics the highlighted simulated images of PFSeq section~\ref{ssec:pfseq}:
\begin{enumerate}
  \item static views in translation with no motion blur (pair 7100),
  \item moving translation with motion blur (pair 10627),
  \item rotation and translation, with motion blur, in a street corner, the most challenging case (pair 11048).
\end{enumerate}
Cameras are placed on top of a van similarly to PFSeq, along the longitudinal axis and with a baseline of 2m.
This is the setup (Cfg~1) of~\cite{Moreau16}, with greyscale cameras of 3296 \texttimes{} 2520px resolution and Sigma 4.5mm \textsc{ex hsm if} lens, that follows equisolid angle projection with a field of view of 180\textdegree{} accordingly to manufacturer's specifications.
Resultant image circle diameter is about 2310px.
\\

In spite of our efforts to have equivalent simulated and real scenes, these real scenes may present different difficulties than PFSeq sequences:
the real lens sharpness is lower than the simulation,
the real images never have such strong motion blur as in PFSeq (could balance the low lens sharpness),
real cameras have a lower dynamic range than the simulation,
and real recorded buildings are lower than those in PFSeq (giving a lower potential spatial distribution of the detected features).


For the real scenes, we propose to report two types of metrics: outcome application stability (through calibration parameters), and statistics inspired by the features evaluation metrics of section~\ref{ssec:evmetrics}.
\\
Stability is represented by the standard deviation of a set of representative parameters in fisheye calibration and fisheye epipolar geometry (stereovision), that are:
\begin{itemize}
 \item $a$ the fisheye projection parameter (following the one-parameter equisolid angle projection $r = 2 f \sin(\frac{\theta}{2}), a = \frac{1}{2f}$~\cite{Moreau16}),
 \item $(X_{e_l}, Y_{e_l}, Z_{e_l})$ and $(X_{e_r}, Y_{e_r}, Z_{e_r})$ the coordinates of the left and right epipoles respectively for the reference image (front image),
 \item and $\varepsilon_\text{\textsc{ransac}}$ the angular error for the best obtained model according to a \textsc{ransac} loop.
\end{itemize}
Feature metrics to provide similar information as for the simulations in sections~\ref{ssec:eval-det} and~\ref{ssec:eval-desc-match} are the following:
\begin{itemize}
  \item The number of attempted matches and their ratio over the lowest number of detections among the two images.
  \item The average number of \textsc{ransac} inliers (over all the iterations) and,
  \begin{itemize}
    \item its ratio over the lowest number of detections among the image pair (to be compared to the matching score assuming the vast majority of the inliers out from the \textsc{ransac} are true).
    \item its ratio over the number of attempted matches.
  \end{itemize}
\end{itemize}


In the real case, we shall keep in mind that we never know if matched points are good or wrong.
For this reason we can not call these measures neither repeatability nor matching score.
The information provided by the \textsc{ransac} are the inliers, they are likely to be truly matching pairs, according to the applied error model.
We express the ratio of the number of inliers over the number of detected points to be compared to the matching score (at least, tendencies between both are comparable), and also over the number of attempted matches because it represents the chances to pick only inliers for a \textsc{ransac} iteration.
\\

We do not report the graphs in this evaluation as they are too cluttered to provide insights.
\\

\def\inputcheminreel#1{
\input{./2014-05-11-pts-reel-cluster-v2/#1} 
}

\begin{table}[htbp]
\caption{Self-calibration statistics with SIFT detector based setups on images pair 7100 ``translation''.
$a$ is the fisheye projection parameter.
$(X_{e_l}, Y_{e_l}, Z_{e_l})$ and $(X_{e_r}, Y_{e_r}, Z_{e_r})$ are the coordinates of the left and right epipoles respectively (for the reference image).
$\varepsilon_\text{\textsc{ransac}}$ is the angular error for the best obtained model according to a \textsc{ransac} loop.
Avg stands for average value.
Std stands for standard deviation.
Underlined setups show high standard deviation(s) meaning they are unstable.
BF stands for BruteForce matching.
\label{tab:statistiques-stabilite-image-7100-sift}}
\centering
\begin{adjustbox}{width=1\textwidth}
\begin{tabular}{l*{16}{|c}}
 & \multicolumn{2}{|c}{$a$} & \multicolumn{2}{|c}{$X_{e_l}$} & \multicolumn{2}{|c}{$Y_{e_l}$} & \multicolumn{2}{|c}{$Z_{e_l}$} & \multicolumn{2}{|c}{$X_{e_r}$} & \multicolumn{2}{|c}{$Y_{e_r}$} & \multicolumn{2}{|c}{$Z_{e_r}$} & \multicolumn{2}{|c}{$\varepsilon_\text{\textsc{ransac}}$} \\
Algorithms setup: SIFT ... & Avg & Std & Avg & Std & Avg & Std & Avg & Std & Avg & Std & Avg & Std & Avg & Std & Avg & Std
 \\
\hline
\hline
{\small nopol 0.02 20.0 1.0 SIFT BF} & 0.720 & 0.01 & 1.000 & 0.00 & -0.003 & 0.03 & 0.006 & 0.02 & 0.999 & 0.00 & 0.001 & 0.03 & -0.020 & 0.02 & 2.4e-07 & 0.00
 \\
\hline
{\small nopol 0.02 20.0 1.6 SIFT BF} & 0.720 & 0.01 & 1.000 & 0.00 & -0.003 & 0.03 & 0.007 & 0.01 & 0.999 & 0.00 & 0.001 & 0.03 & -0.019 & 0.01 & 2.3e-07 & 0.00
 \\
\hline
\underline{\small nopol 0.04 5.0 1.6 FREAK 44.0 BF} & 0.706 & 0.04 & 0.891 & 0.21 & 0.059 & 0.30 & -0.013 & 0.26 & 0.892 & 0.21 & 0.057 & 0.29 & -0.037 & 0.27 & 2.7e-07 & 0.00
 \\
\hline
{\small nopol 0.04 5.0 1.6 SIFT BF} & 0.723 & 0.01 & 1.000 & 0.00 & 0.001 & 0.02 & 0.007 & 0.01 & 0.999 & 0.00 & 0.005 & 0.02 & -0.018 & 0.01 & 2.5e-07 & 0.00
 \\
\hline
{\small nopol 0.04 5.0 3.0 SIFT BF} & 0.721 & {0.01} & 1.000 & {0.00} & 0.001 & {0.02} & 0.010 & 0.01 & 1.000 & {0.00} & 0.005 & {0.02} & -0.015 & 0.01 & 2.5e-07 & 0.00
 \\
\hline
{\small nopol 0.08 10.0 3.0 SIFT BF} & 0.713 & 0.01 & 1.000 & 0.00 & -0.012 & 0.04 & -0.008 & 0.02 & 0.999 & 0.00 & -0.009 & 0.04 & -0.034 & 0.02 & 2.7e-07 & 0.00
 \\
\hline
\underline{\small nopol 0.08 5.0 3.0 FREAK 44.0 BF} & 0.697 & 0.05 & 0.847 & 0.25 & 0.144 & 0.31 & -0.128 & 0.29 & 0.791 & 0.32 & 0.145 & 0.29 & -0.148 & 0.28 & 2.7e-07 & 0.00
 \\
\hline
{\small nopol 0.08 5.0 3.0 SIFT BF} & 0.713 & 0.01 & 0.993 & 0.08 & -0.008 & 0.03 & -0.006 & {0.01} & 0.999 & 0.00 & -0.005 & 0.03 & -0.032 & {0.01} & 2.7e-07 & 0.00
 \\
\hline
{\small pol 0.02 20.0 1.6 SIFT BF} & 0.720 & 0.01 & 1.000 & 0.00 & -0.003 & 0.03 & 0.007 & 0.02 & 0.999 & 0.00 & 0.001 & 0.03 & -0.019 & 0.02 & 2.4e-07 & 0.00
 \\
\hline
{\small pol 0.02 20.0 3.0 SIFT BF} & 0.720 & 0.01 & 1.000 & 0.00 & -0.002 & 0.02 & 0.008 & 0.01 & 1.000 & 0.00 & 0.002 & 0.02 & -0.018 & 0.01 & 2.3e-07 & {0.00}
 \\
\hline
\underline{\small pol 0.04 5.0 3.0 FREAK 44.0 BF} & 0.736 & 0.02 & 0.687 & 0.29 & -0.091 & 0.57 & 0.014 & 0.31 & 0.758 & 0.27 & -0.098 & 0.50 & -0.008 & 0.29 & 3.2e-07 & 0.00
 \\
\end{tabular}
\end{adjustbox}
\end{table}

\begin{table}[htbp]
\caption{Self-calibration statistics with SIFT detector based setups on images pair 10627 ``translation motion''.
$a$ is the fisheye projection parameter.
$(X_{e_l}, Y_{e_l}, Z_{e_l})$ and $(X_{e_r}, Y_{e_r}, Z_{e_r})$ are the coordinates of the left and right epipoles respectively (for the reference image).
$\varepsilon_\text{\textsc{ransac}}$ is the angular error for the best obtained model according to a \textsc{ransac} loop.
Avg stands for average value.
Std stands for standard deviation.
Underlined setups show high standard deviation(s) meaning they are unstable.
BF stands for BruteForce matching.
\label{tab:statistiques-stabilite-image-10627-sift}}
\centering
\begin{adjustbox}{width=1\textwidth}
\begin{tabular}{l*{16}{|c}}
 & \multicolumn{2}{|c}{$a$} & \multicolumn{2}{|c}{$X_{e_l}$} & \multicolumn{2}{|c}{$Y_{e_l}$} & \multicolumn{2}{|c}{$Z_{e_l}$} & \multicolumn{2}{|c}{$X_{e_r}$} & \multicolumn{2}{|c}{$Y_{e_r}$} & \multicolumn{2}{|c}{$Z_{e_r}$} & \multicolumn{2}{|c}{$\varepsilon_\text{\textsc{ransac}}$} \\
Algorithms setup: SIFT ... & Avg & Std & Avg & Std & Avg & Std & Avg & Std & Avg & Std & Avg & Std & Avg & Std & Avg & Std
 \\
\hline
\hline
{\small nopol 0.02 20.0 1.0 SIFT BF} & 0.718 & 0.01 & 1.000 & 0.00 & 0.005 & 0.02 & 0.005 & 0.01 & 0.997 & 0.00 & -0.007 & 0.02 & 0.071 & 0.02 & 2.6e-07 & {0.00}
 \\
\hline
{\small nopol 0.02 20.0 1.6 SIFT BF} & 0.720 & 0.01 & 1.000 & {0.00} & 0.003 & 0.01 & 0.005 & 0.01 & 0.997 & 0.00 & -0.009 & 0.01 & 0.071 & 0.01 & 2.5e-07 & 0.00
 \\
\hline
\underline{\small nopol 0.04 5.0 1.6 FREAK 44.0 BF} & 0.720 & 0.03 & 0.768 & 0.27 & -0.229 & 0.44 & -0.031 & 0.29 & 0.757 & 0.27 & -0.242 & 0.45 & 0.012 & 0.29 & 2.9e-07 & 0.00
 \\
\hline
{\small nopol 0.04 5.0 1.6 SIFT BF} & 0.721 & 0.01 & 0.999 & 0.03 & 0.003 & 0.02 & 0.002 & 0.01 & 0.996 & 0.03 & -0.009 & 0.02 & 0.069 & 0.01 & 2.5e-07 & 0.00
 \\
\hline
{\small nopol 0.04 5.0 3.0 SIFT BF} & 0.722 & {0.00} & 0.999 & 0.03 & -0.004 & {0.01} & 0.008 & {0.01} & 0.996 & 0.03 & -0.016 & {0.01} & 0.076 & {0.01} & 2.4e-07 & 0.00
 \\
\hline
{\small nopol 0.08 10.0 3.0 SIFT BF} & 0.719 & 0.01 & 0.998 & 0.04 & 0.003 & 0.02 & -0.003 & 0.01 & 0.997 & 0.03 & -0.009 & 0.02 & 0.063 & 0.01 & 2.8e-07 & 0.00
 \\
\hline
\underline{\small nopol 0.08 5.0 3.0 FREAK 44.0 BF} & 0.719 & 0.03 & 0.808 & 0.30 & -0.075 & 0.47 & 0.044 & 0.16 & 0.796 & 0.30 & -0.075 & 0.48 & 0.097 & 0.17 & 3.1e-07 & 0.00
 \\
\hline
{\small nopol 0.08 5.0 3.0 SIFT BF} & 0.721 & 0.01 & 1.000 & 0.00 & 0.012 & 0.02 & 0.002 & 0.01 & 0.997 & {0.00} & -0.000 & 0.02 & 0.068 & 0.01 & 2.9e-07 & 0.00
 \\
\hline
{\small pol 0.02 20.0 1.6 SIFT BF} & 0.721 & 0.01 & 1.000 & 0.00 & 0.003 & 0.02 & 0.005 & 0.02 & 0.997 & 0.00 & -0.009 & 0.02 & 0.072 & 0.02 & 2.5e-07 & 0.00
 \\
\hline
{\small pol 0.02 20.0 3.0 SIFT BF} & 0.722 & 0.01 & 1.000 & 0.00 & 0.002 & 0.01 & 0.008 & 0.01 & 0.996 & 0.03 & -0.011 & 0.01 & 0.075 & 0.01 & 2.4e-07 & 0.00
 \\
\hline
\underline{\small pol 0.04 5.0 3.0 FREAK 44.0 BF} & 0.738 & 0.01 & 0.684 & 0.30 & 0.026 & 0.53 & -0.009 & 0.39 & 0.776 & 0.27 & -0.105 & 0.50 & 0.028 & 0.20 & 3.0e-07 & 0.00
 \\
\end{tabular}
\end{adjustbox}
\end{table}

\begin{table}[htbp]
\caption{Self-calibration statistics with SIFT detector based setups on images pair 11048 ``translation and rotation motion''.
$a$ is the fisheye projection parameter.
$(X_{e_l}, Y_{e_l}, Z_{e_l})$ and $(X_{e_r}, Y_{e_r}, Z_{e_r})$ are the coordinates of the left and right epipoles respectively (for the reference image).
$\varepsilon_\text{\textsc{ransac}}$ is the angular error for the best obtained model according to a \textsc{ransac} loop.
Avg stands for average value.
Std stands for standard deviation.
Underlined setups show high standard deviation(s) meaning they are unstable.
BF stands for BruteForce matching.
\label{tab:statistiques-stabilite-image-11048-sift}}
\centering
\begin{adjustbox}{width=1\textwidth}
\begin{tabular}{l*{16}{|c}}
 & \multicolumn{2}{|c}{$a$} & \multicolumn{2}{|c}{$X_{e_l}$} & \multicolumn{2}{|c}{$Y_{e_l}$} & \multicolumn{2}{|c}{$Z_{e_l}$} & \multicolumn{2}{|c}{$X_{e_r}$} & \multicolumn{2}{|c}{$Y_{e_r}$} & \multicolumn{2}{|c}{$Z_{e_r}$} & \multicolumn{2}{|c}{$\varepsilon_\text{\textsc{ransac}}$} \\
Algorithms setup: SIFT ... & Avg & Std & Avg & Std & Avg & Std & Avg & Std & Avg & Std & Avg & Std & Avg & Std & Avg & Std
 \\
\hline
\hline
{\small nopol 0.02 20.0 1.0 SIFT BF} & 0.729 & 0.01 & 1.000 & 0.00 & 0.004 & 0.02 & -0.016 & 0.01 & 0.998 & 0.00 & -0.015 & 0.03 & 0.050 & 0.01 & 2.4e-07 & 0.00
 \\
\hline
{\small nopol 0.02 20.0 1.6 SIFT BF} & 0.726 & {0.01} & 1.000 & 0.00 & 0.001 & 0.02 & -0.011 & {0.01} & 0.998 & 0.00 & -0.018 & 0.02 & 0.054 & {0.01} & 2.3e-07 & 0.00
 \\
\hline
\underline{\small nopol 0.04 5.0 1.6 FREAK 44.0 BF} & 0.681 & 0.05 & 0.792 & 0.26 & -0.286 & 0.42 & -0.009 & 0.21 & 0.767 & 0.27 & -0.313 & 0.43 & 0.037 & 0.23 & 2.6e-07 & 0.00
 \\
\hline
{\small nopol 0.04 5.0 1.6 SIFT BF} & 0.727 & 0.01 & 1.000 & {0.00} & -0.000 & 0.01 & -0.008 & 0.01 & 0.997 & 0.03 & -0.019 & 0.01 & 0.057 & 0.01 & 2.1e-07 & {0.00}
 \\
\hline
{\small nopol 0.04 5.0 3.0 SIFT BF} & 0.723 & 0.01 & 0.999 & 0.03 & -0.000 & 0.01 & -0.005 & 0.01 & 0.998 & 0.00 & -0.019 & 0.01 & 0.061 & 0.01 & 2.2e-07 & 0.00
 \\
\hline
{\small nopol 0.08 10.0 3.0 SIFT BF} & 0.720 & 0.01 & 1.000 & 0.00 & -0.005 & 0.01 & -0.010 & 0.02 & 0.998 & 0.00 & -0.025 & 0.01 & 0.053 & 0.02 & 2.4e-07 & 0.00
 \\
\hline
\underline{\small nopol 0.08 5.0 3.0 FREAK 44.0 BF} & 0.682 & 0.05 & 0.781 & 0.26 & -0.218 & 0.48 & 0.004 & 0.22 & 0.762 & 0.26 & -0.234 & 0.49 & 0.051 & 0.23 & 2.7e-07 & 0.00
 \\
\hline
{\small nopol 0.08 5.0 3.0 SIFT BF} & 0.723 & 0.01 & 1.000 & 0.00 & -0.003 & {0.01} & -0.015 & 0.02 & 0.998 & 0.00 & -0.022 & {0.01} & 0.048 & 0.02 & 2.2e-07 & 0.00
 \\
\hline
{\small pol 0.02 20.0 1.6 SIFT BF} & 0.727 & 0.01 & 0.999 & 0.03 & 0.001 & 0.02 & -0.010 & 0.01 & 0.997 & 0.03 & -0.018 & 0.02 & 0.055 & 0.01 & 2.4e-07 & 0.00
 \\
\hline
{\small pol 0.02 20.0 3.0 SIFT BF} & 0.728 & 0.01 & 1.000 & 0.00 & 0.006 & 0.02 & -0.012 & 0.01 & 0.998 & {0.00} & -0.012 & 0.02 & 0.054 & 0.01 & 2.3e-07 & 0.00
 \\
\hline
\underline{\small pol 0.04 5.0 3.0 FREAK 44.0 BF} & 0.732 & 0.02 & 0.687 & 0.29 & -0.066 & 0.57 & 0.216 & 0.23 & 0.724 & 0.24 & -0.038 & 0.49 & 0.333 & 0.24 & 3.0e-07 & 0.00
 \\
\end{tabular}
\end{adjustbox}
\end{table}


%
%
%
%
%
%
%
%

\begin{table}[htbp]
\caption{Self-calibration statistics with CenSurE detector based setups on images pair 7100 ``translation''.
$a$ is the fisheye projection parameter.
$(X_{e_l}, Y_{e_l}, Z_{e_l})$ and $(X_{e_r}, Y_{e_r}, Z_{e_r})$ are the coordinates of the left and right epipoles respectively (for the reference image).
$\varepsilon_\text{\textsc{ransac}}$ is the angular error for the best obtained model according to a \textsc{ransac} loop.
Avg stands for average value.
Std stands for standard deviation.
All the setups provide low and acceptable standard deviations and stability.
BF stands for BruteForce matching.
RT stands for RatioTest matching.
\label{tab:statistiques-stabilite-image-7100-censure}}
\centering
\begin{adjustbox}{width=1\textwidth}
\begin{tabular}{l*{16}{|c}}
 & \multicolumn{2}{|c}{$a$} & \multicolumn{2}{|c}{$X_{e_l}$} & \multicolumn{2}{|c}{$Y_{e_l}$} & \multicolumn{2}{|c}{$Z_{e_l}$} & \multicolumn{2}{|c}{$X_{e_r}$} & \multicolumn{2}{|c}{$Y_{e_r}$} & \multicolumn{2}{|c}{$Z_{e_r}$} & \multicolumn{2}{|c}{$\varepsilon_\text{\textsc{ransac}}$} \\
Algorithms setup: CenSurE ... & Avg & Std & Avg & Std & Avg & Std & Avg & Std & Avg & Std & Avg & Std & Avg & Std & Avg & Std
 \\
\hline
\hline
{\small nopol 16 10 7 ORB 2 BF} & 0.720 & {0.01} & 1.000 & 0.00 & -0.009 & {0.02} & 0.002 & {0.01} & 0.999 & {0.00} & -0.006 & {0.02} & -0.023 & {0.01} & 2.9e-07 & 0.00
 \\
\hline
{\small nopol 16 20 3 BRISK 1.0 RT} & 0.702 & 0.05 & 0.996 & 0.04 & -0.025 & 0.03 & -0.023 & 0.04 & 0.995 & 0.04 & -0.023 & 0.03 & -0.049 & 0.04 & 2.5e-07 & 0.00
 \\
\hline
{\small nopol 32 20 5 SIFT FLANN} & 0.698 & 0.03 & 0.997 & 0.03 & -0.036 & 0.03 & -0.022 & 0.03 & 0.997 & 0.03 & -0.033 & 0.03 & -0.048 & 0.03 & 2.6e-07 & 0.00
 \\
\hline
{\small nopol 32 20 7 FREAK 22.0 BF} & 0.717 & 0.02 & 0.999 & 0.00 & -0.027 & 0.03 & -0.012 & 0.02 & 0.998 & 0.00 & -0.024 & 0.03 & -0.037 & 0.02 & 2.8e-07 & {0.00}
 \\
\hline
{\small nopol 8 10 5 ORB 2 BF} & 0.728 & 0.01 & 0.999 & {0.00} & -0.004 & 0.03 & -0.007 & 0.02 & 0.999 & 0.00 & -0.000 & 0.03 & -0.032 & 0.02 & 2.8e-07 & 0.00
 \\
\hline
{\small pol 32 10 3 SIFT FLANN} & 0.702 & 0.05 & 0.995 & 0.06 & -0.023 & 0.05 & -0.008 & 0.02 & 0.994 & 0.06 & -0.019 & 0.05 & -0.035 & 0.02 & 2.4e-07 & 0.00
 \\
\hline
{\small pol 32 20 7 ORB 2 BF} & 0.718 & 0.01 & 0.999 & 0.00 & -0.008 & 0.02 & -0.010 & 0.02 & 0.999 & 0.00 & -0.005 & 0.02 & -0.036 & 0.02 & 2.8e-07 & 0.00
 \\
\end{tabular}
\end{adjustbox}
\end{table}


%
%
%
%
%
%
%
%

\begin{table}[htbp]
\caption{Self-calibration statistics with CenSurE detector based setups on images pair 10627 ``translation motion''.
$a$ is the fisheye projection parameter.
$(X_{e_l}, Y_{e_l}, Z_{e_l})$ and $(X_{e_r}, Y_{e_r}, Z_{e_r})$ are the coordinates of the left and right epipoles respectively (for the reference image).
$\varepsilon_\text{\textsc{ransac}}$ is the angular error for the best obtained model according to a \textsc{ransac} loop.
Avg stands for average value.
Std stands for standard deviation.
Underlined setups show high standard deviation(s) meaning they are unstable.
BF stands for BruteForce matching.
RT stands for RatioTest matching.
\label{tab:statistiques-stabilite-image-10627-censure}}
\centering
\begin{adjustbox}{width=1\textwidth}
\begin{tabular}{l*{16}{|c}}
 & \multicolumn{2}{|c}{$a$} & \multicolumn{2}{|c}{$X_{e_l}$} & \multicolumn{2}{|c}{$Y_{e_l}$} & \multicolumn{2}{|c}{$Z_{e_l}$} & \multicolumn{2}{|c}{$X_{e_r}$} & \multicolumn{2}{|c}{$Y_{e_r}$} & \multicolumn{2}{|c}{$Z_{e_r}$} & \multicolumn{2}{|c}{$\varepsilon_\text{\textsc{ransac}}$} \\
Algorithms setup: CenSurE ... & Avg & Std & Avg & Std & Avg & Std & Avg & Std & Avg & Std & Avg & Std & Avg & Std & Avg & Std
 \\
\hline
\hline
{\small nopol 16 10 7 ORB 2 BF} & 0.726 & 0.01 & 1.000 & 0.00 & 0.017 & 0.02 & -0.005 & 0.02 & 0.998 & 0.00 & 0.005 & 0.02 & 0.062 & 0.02 & 2.9e-07 & 0.00
 \\
\hline
{\small nopol 16 20 3 BRISK 1.0 RT} & 0.721 & 0.05 & 0.994 & 0.07 & 0.013 & 0.03 & -0.000 & 0.02 & 0.992 & 0.07 & 0.003 & 0.03 & 0.066 & 0.02 & 3.3e-07 & 0.00
 \\
\hline
\underline{\small nopol 32 20 5 SIFT FLANN} & 0.724 & 0.06 & 0.863 & 0.34 & 0.015 & 0.02 & -0.010 & 0.01 & 0.992 & 0.08 & 0.006 & 0.02 & 0.056 & 0.02 & 2.6e-07 & 0.00
 \\
\hline
{\small nopol 32 20 7 FREAK 22.0 BF} & 0.719 & 0.01 & 1.000 & 0.00 & 0.016 & 0.01 & -0.007 & 0.01 & 0.998 & 0.00 & 0.004 & 0.01 & 0.058 & 0.02 & 3.0e-07 & {0.00}
 \\
\hline
{\small nopol 8 10 5 ORB 2 BF} & 0.720 & 0.01 & 0.998 & 0.01 & 0.029 & 0.05 & -0.002 & 0.03 & 0.996 & 0.02 & 0.015 & 0.05 & 0.062 & 0.04 & 3.0e-07 & 0.00
 \\
\hline
\underline{\small pol 32 10 3 SIFT FLANN} & 0.718 & 0.08 & 0.967 & 0.18 & 0.007 & 0.02 & -0.006 & {0.01} & 0.987 & 0.10 & -0.004 & 0.02 & 0.060 & 0.01 & 2.7e-07 & 0.00
 \\
\hline
{\small pol 32 20 7 ORB 2 BF} & 0.723 & {0.01} & 1.000 & {0.00} & -0.001 & {0.01} & 0.006 & 0.01 & 0.997 & {0.00} & -0.012 & {0.01} & 0.074 & {0.01} & 2.9e-07 & 0.00
 \\
\end{tabular}
\end{adjustbox}
\end{table}


%
%
%
%
%
%
%
%

\begin{table}[htbp]
\caption{Self-calibration statistics with CenSurE detector based setups on images pair 11048 ``translation and rotation motion''.
$a$ is the fisheye projection parameter.
$(X_{e_l}, Y_{e_l}, Z_{e_l})$ and $(X_{e_r}, Y_{e_r}, Z_{e_r})$ are the coordinates of the left and right epipoles respectively (for the reference image).
$\varepsilon_\text{\textsc{ransac}}$ is the angular error for the best obtained model according to a \textsc{ransac} loop.
Avg stands for average value.
Std stands for standard deviation.
Underlined setups show high standard deviation(s) meaning they are unstable.
BF stands for BruteForce matching.
RT stands for RatioTest matching.
\label{tab:statistiques-stabilite-image-11048-censure}}
\centering
\begin{adjustbox}{width=1\textwidth}
\begin{tabular}{l*{16}{|c}}
 & \multicolumn{2}{|c}{$a$} & \multicolumn{2}{|c}{$X_{e_l}$} & \multicolumn{2}{|c}{$Y_{e_l}$} & \multicolumn{2}{|c}{$Z_{e_l}$} & \multicolumn{2}{|c}{$X_{e_r}$} & \multicolumn{2}{|c}{$Y_{e_r}$} & \multicolumn{2}{|c}{$Z_{e_r}$} & \multicolumn{2}{|c}{$\varepsilon_\text{\textsc{ransac}}$} \\
Algorithms setup: CenSurE ... & Avg & Std & Avg & Std & Avg & Std & Avg & Std & Avg & Std & Avg & Std & Avg & Std & Avg & Std
 \\
\hline
\hline
{\small nopol 16 10 7 ORB 2 BF} & 0.706 & 0.01 & 1.000 & {0.00} & 0.003 & 0.02 & -0.013 & {0.02} & 0.998 & {0.00} & -0.016 & 0.02 & 0.045 & {0.02} & 2.7e-07 & 0.00
 \\
\hline
{\small nopol 16 20 3 BRISK 1.0 RT} & 0.715 & 0.03 & 0.998 & 0.03 & -0.002 & 0.02 & -0.005 & 0.03 & 0.997 & 0.03 & -0.023 & 0.02 & 0.055 & 0.02 & 2.5e-07 & 0.00
 \\
\hline
\underline{\small nopol 32 20 5 SIFT FLANN} & 0.621 & 0.02 & 0.822 & 0.15 & -0.473 & 0.16 & -0.155 & 0.07 & 0.830 & 0.09 & -0.514 & 0.15 & -0.109 & 0.07 & 2.3e-07 & 0.00
 \\
\hline
{\small nopol 32 20 7 FREAK 22.0 BF} & 0.722 & 0.01 & 0.999 & 0.00 & 0.003 & 0.03 & -0.016 & 0.02 & 0.998 & 0.00 & -0.016 & 0.03 & 0.047 & 0.02 & 2.8e-07 & 0.00
 \\
\hline
{\small nopol 8 10 5 ORB 2 BF} & 0.716 & 0.03 & 0.996 & 0.03 & 0.047 & 0.03 & -0.028 & 0.03 & 0.997 & 0.03 & 0.030 & 0.04 & 0.030 & 0.03 & 2.4e-07 & 0.00
 \\
\hline
{\small pol 32 10 3 SIFT FLANN} & 0.717 & 0.01 & 0.999 & 0.00 & 0.009 & 0.02 & -0.008 & 0.02 & 0.998 & 0.00 & -0.010 & 0.02 & 0.054 & 0.02 & 2.6e-07 & {0.00}
 \\
\hline
{\small pol 32 20 7 ORB 2 BF} & 0.718 & {0.01} & 0.999 & 0.00 & 0.022 & {0.02} & -0.002 & 0.02 & 0.998 & 0.00 & 0.004 & {0.02} & 0.060 & 0.02 & 2.5e-07 & 0.00
 \\
\end{tabular}
\end{adjustbox}
\end{table}


We can check the stability of highlighted setups by looking at the tables~\ref{tab:statistiques-stabilite-image-7100-sift}, \ref{tab:statistiques-stabilite-image-10627-sift}, \ref{tab:statistiques-stabilite-image-11048-sift}, \ref{tab:statistiques-stabilite-image-7100-censure}, \ref{tab:statistiques-stabilite-image-10627-censure} and \ref{tab:statistiques-stabilite-image-11048-censure}.
Standard deviations can be close and low for all the parameters, meaning that many setups provide stable final estimations.
Instead of looking for the most stable of them, we shall withdraw the setups that are prone to have unpredictable behaviour.
\\
With SIFT detector, the only represented matching procedure is BruteForce.
We can note a strong loss of stability when using FREAK descriptor.
Hence, we shall avoid using this descriptor with SIFT detector and stick with SIFT descriptor.
Also, in some cases, average estimations from the highlighted unstable setups can diverge from the overall tendencies.
It is interesting to note that the stability of the $a$ parameter, the only intrinsic camera parameter in our use case, is not affected even if different from the global tendency.
\\
FREAK seems to behave better and to be competitive when used with CenSurE detector.
There are fewer highlighted setups with CenSurE detector, but with more varied alternatives.
SIFT descriptor and FLANN matching occur together only.
BRISK descriptor and RatioTest matching occur together only also.
And BruteForce matching is highlighted with either ORB or FREAK descriptor.
Stability with SIFT descriptor and FLANN matching is lower than all the other alternatives, with some disagreements with the average tendencies also.
Regarding the other setups, BRISK descriptor with RatioTest matching seems slightly less stable than the other options, relying all on BruteForce matching.
Finally, getting stable results with CenSurE detector imply using BruteForce matching, associated with ORB or FREAK descriptor.
\\
As a general rule, stability is better when using BruteForce matching, and SIFT detector shall fit only with SIFT descriptor while CenSurE detector may fit either with ORB or FREAK descriptor.
This stability study either cancels the potential interest in using FREAK descriptor with SIFT detector suggested as a viable alternative in the case of strong motion blur in section~\ref{ssec:eval-desc-match}, and/or illustrates that proposed real examples do not have the same challenges as PFSeq (less motion blur, but also less textures).
\\


\input{./2014-05-11-pts-reel-cluster-v2/stats_SIFT_7100_tab.tex} 

\input{./2014-05-11-pts-reel-cluster-v2/stats_SIFT_10627_tab.tex} 

\input{./2014-05-11-pts-reel-cluster-v2/stats_SIFT_11048_tab.tex} 

\input{./2014-05-11-pts-reel-cluster-v2/stats_CenSurE_7100_tab.tex} 

\input{./2014-05-11-pts-reel-cluster-v2/stats_CenSurE_10627_tab.tex} 

\input{./2014-05-11-pts-reel-cluster-v2/stats_CenSurE_11048_tab.tex} 

The other metrics might help to find the best setups, with tables~\ref{tab:statistiques-image-reelle-numero-7100-sift}, \ref{tab:statistiques-image-reelle-numero-10627-sift}, \ref{tab:statistiques-image-reelle-numero-11048-sift}, \ref{tab:statistiques-image-reelle-numero-7100-censure}, \ref{tab:statistiques-image-reelle-numero-10627-censure} and \ref{tab:statistiques-image-reelle-numero-11048-censure}.
In all the cases, setups previously identified as unstable provide also low ratio of attempted matches over the number of detections, and low ratio of inliers over the number of detections and/or over the number of attempted matches, including CenSurE \emph{nopol 16 20 3 BRISK 1.0 RatioTest}.
These setups have usually not so many matches, and surprisingly, those based on CenSurE detector show a low average number of \textsc{ransac} iterations (with a maximum of approximately 8000 iterations).
Near all the other setups reach the limit of 50000 \textsc{ransac} iterations.
This early \textsc{ransac} stop, coupled with the lower stability of these setups, is a signal they can not converge to accurate parameters estimations.
So again, for the following, we shall consider SIFT descriptor with BruteForce matching for SIFT detector, and ORB and FREAK descriptor with BruteForce matching for CenSurE detector.
\\
From these remaining setups, SIFT detector based setups usually reach much more inliers than CenSurE detector based setups.
Having more \textsc{ransac} inliers is important to guarantee accurate results leveraging the data noise with numerous inputs in the refinement step (or, we can also use a tighter \textsc{ransac} error threshold for its evaluation step).
Anyway, it seems some setups based on CenSurE detector might work well enough if faster algorithms than SIFT are required.
We could consider CenSurE with \emph{nopol 16 10 7 ORB 2 BruteForce}, with \emph{nopol 32 20 7 FREAK 22.0 BruteForce} and with \emph{CenSurE pol 32 20 7 ORB 2 BruteForce}, as they provide an acceptable quantity of inliers.
\\
SIFT detector setups rarely reach the expected 35\% of inliers over the attempted matches and always need the 50000 or nearby \textsc{ransac} iterations, meaning it is hard to reach this ratio of inliers in real usage.
On the contrary, some CenSurE setups do surpass the 35\% ratio of inliers over the attempted matches, thus requiring less \textsc{ransac} iterations.
It occurs for all the CenSurE detector coupled with ORB descriptor setups for the most challenging scene 11048 ``translation and rotation motion'', and CenSurE \emph{nopol 16 10 7 ORB 2 BruteForce} for the scene 7100 ``translation''.
With no reference calibration parameters, we can not make definitive conclusions on these results, but we still see differences between the scenes.
One hypothesis could be, as CenSurE detector setups provide usually less detections and attempted matches than SIFT detector setups, there might be less erroneous matches, meaning a higher proportion of correct attempted matches and inliers.
However, this is refuted by the results on PFSeq in section~\ref{ssec:eval-desc-match} that show CenSurE is far from being able to give as good matching scores as SIFT in the same conditions (together with less detections and correct attempted matches).
The other hypothesis is that, in the former case, for the challenging scene, we could have a similar behaviour than the unstable setups, hardly reaching accurate estimations.
In the latter case, CenSurE \emph{nopol 16 10 7 ORB 2 BruteForce} reaches almost 50000 iterations, meaning it shall behave as expected and shall be able to estimate the calibration parameters with stability and accuracy.
Through these considerations, the safest setup based on CenSurE detector is with \emph{nopol 32 20 7 FREAK 22.0 BruteForce}, showing the advantage of FREAK descriptor over ORB descriptor for challenging scenes, while being equivalent for the other scenes.
\\
Now, all the results based on SIFT detector and SIFT descriptor are likely to behave properly with all the scenes.
Between them, the setups using the polar rectification always provide a very high quantity of detections, but are on the lower end for all the other metrics, meaning most of the detected points are useless and may even disturb the subsequent tasks.
For example, the comparison between SIFT with \emph{nopol 0.02 20.0 1.6 SIFT BruteForce} and with \emph{pol 0.02 20.0 1.6 SIFT BruteForce} is straightforward: the variant with no polar rectification has usually more attempted matches and inliers after \textsc{ransac} (and always better ratios), starting with less detections!
Hence, the polar rectification is still not advantageous on real scenes.
Among the other setups, the setups with the best ratio of inliers over attempted matches are among SIFT with \emph{nopol 0.04 5.0 ...} and with \emph{nopol 0.08 ... 3.0 ...} for all the scenes with similar scores.
Between them, we would choose a SIFT with \emph{nopol 0.04 5.0 ...} alternative, as they usually allow for twice more inliers, and more specifically SIFT with \emph{nopol 0.04 5.0 3.0 SIFT BruteForce} as the best compromise over all.
However, a strange behaviour to note is we can reach and slightly surpass 35\% ratio of inliers over attempted matches for challenging scenes, but not with the less challenging scene 7100 ``translation''.
Here there are more detections, showing their descriptions are likely to be more ambiguous: descriptors may benefit from certain amount of blur.
\\
Based on the results on tested real scenes, we would recommend overall using SIFT with \emph{nopol 0.04 5.0 3.0 SIFT BruteForce}, or CenSurE with \emph{nopol 32 20 7 FREAK 22.0 BruteForce} as an acceptable alternative when fast computations are required.
Their qualities are confirmed by their good convergence ability within a \textsc{ransac} loop and their good stability.
Finally, these results highlight slightly different setups than the evaluations on PFSeq in section~\ref{ssec:eval-desc-match}, where we reported SIFT with \emph{nopol 0.04 5.0 1.6 SIFT BruteForce} as a recommended and predictable setting, and where setups based on CenSurE detector failed in case of very strong blur.


 
\section{Conclusion}
\label{sec:conclu}

Results on PFSeq in section~\ref{sec:eval-simu} find the best setup to be SIFT \emph{nopol 0.02 20.0 1.6 SIFT BruteForce}, and results on real scenes in section~\ref{sec:eval-reel} find the best setup to be SIFT \emph{nopol 0.04 5.0 3.0 SIFT BruteForce}.
This means we can always rely on SIFT detector and descriptor with BruteForce matching, but the domain change may require to tune the parameters.
Also, SIFT \emph{nopol 0.04 5.0 3.0 SIFT BruteForce} is very close to SIFT \emph{nopol 0.04 5.0 1.6 SIFT BruteForce} for all the metrics with the real scenes.
And, these two setups are among the best on PFSeq images, except with the very challenging scene 24, with rotation and motion blur.
The real results confirm well the simulated experiments despite the real scenes do not reach a so strong motion blur.
In addition, all the experiments discard any interest for the polar rectification described in section~\ref{ssec:polar-transform} (and it adds computational costs).
Finally, without features algorithms explicitly designed for fisheye images, the best overall choice is the SIFT detector with BruteForce matching, and the best settings to adopt are:
\begin{itemize}
 \item detector: SIFT with
 \begin{itemize}
  \item {\ttfamily contrastThreshold} set to 0.04
  \item {\ttfamily edgeThreshold} set to 5
  \item {\ttfamily sigma} set to 1.6 (slightly more inliers) or to 3.0 (slightly better inliers ratio)
 \end{itemize}
 \item descriptor: SIFT with the same parameters
\end{itemize}
There is a lack of very good alternative to SIFT detector and descriptor.
The only acceptable different choice might be CenSurE detector with FREAK descriptor, \textbf{with the advantage of being fast} thanks to their low computational cost, but that may still requires improvements for reliable use in all the cases.
Finally, these conclusions are valid for high resolution images similar to ours, but features parameters are not guaranteed to be optimal with lower resolution pictures despite their invariance to scale.
\\


Within this work, we propose first an extensive review of existing benchmarks to evaluate features algorithms.
Secondly, we propose a new simulated photorealistic benchmark PFSeq, to evaluate features on fisheye images, as well as other key applications such as fisheye self-calibration, fisheye visual odometry, fisheye stereovision and fisheye sky segmentation. 
Thirdly, we conduct a comprehensive evaluation of a large variety of features algorithms setups on PFSeq simulated scenes, as well as on real scenes to validate the thoroughness of our benchmark.
Finally, we are able to extract recommended settings for classic features algorithms applied to fisheye vision.
\\

Future works include new evaluations of features algorithms explicitly designed for omnidirectional images, and the definition and usage of a spreading score metric (some tracks are given in section~\ref{par:spreadingscore}).
We may also explore and validate the mix of different detectors and descriptors, 
and proceed on PFSeq the same self-calibration evaluation as in section~\ref{sec:eval-reel} with the addition of a comparison between estimated parameters and ground truth.
This study reveals the need for fast, prolific and accurate features designed for omnidirectional vision, not only inpired by SIFT, as well as faster and optimal matching algorithms implementations.
Also, an update to today's state-of-the-art will be required.
We can expect exceptional performances from learning-based approaches.

\paragraph{Acknowledgments\\}
This work has been carried out within the ANR MEDDTL PREDIT CAPLOC project.


\newpage

\bibliographystyle{alpha}
\bibliography{./morebib,./jabref/bibliographie} 

\end{document}

%% file: 2014-04-07-plot-v2/mix_detecteurs_0_nbcorresp_best_gphe.tex
\begin{figure}[htbp]
\centering
\includegraphics{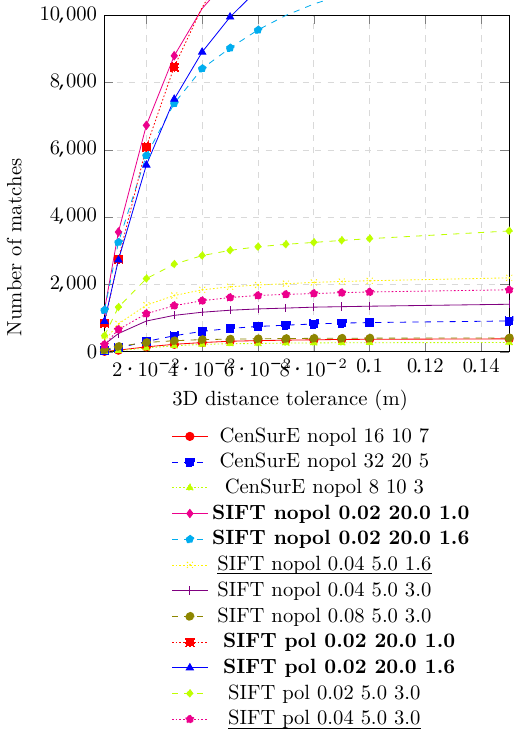}
\caption{Number of matches over distance tolerance for pair 0 "translation". Can be split easily into 2 clusters, the well performing setups are boldened in the legend.
Underlined setups in the legend have the best overall number of matches among the good performers in terms of repeatability highlighted in figure~\ref{fig:repetabilite-image-0}.}
\label{fig:nombre-de-correspondants-image-0}
\end{figure}

%% file: 2014-04-07-plot-v2/mix_detecteurs_0_repetabilite_best_gphe.tex
\begin{figure}[htbp]
\centering
\includegraphics{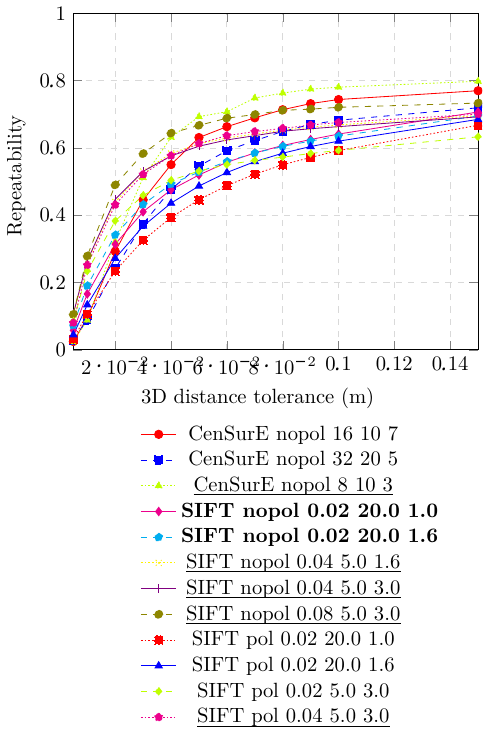}
\caption{Repeatability over distance tolerance for pair 0 "translation".
The best performing setups are underlined in the legend.
Boldened setups in the legend have the best overall repeatabilities among the good performers in terms of number of matches highlighted in figure~\ref{fig:nombre-de-correspondants-image-0}.}
\label{fig:repetabilite-image-0}
\end{figure}

%% file: 2014-04-07-plot-v2/mix_detecteurs_0_nbcorresp_best_tab.tex
\begin{table}[htbp]
\caption{Number of matches for pair 0 "translation", numerical values. Bold values are the best for a given tolerance, underlined are second best.
\label{tab:nombre-de-correspondants-image-0}}
\centering
\begin{tabular}{l*{12}{|c}}
\multicolumn{1}{c|}{\rotatebox{90}{Distance tolerance (m)}} & \rotatebox{90}{CenSurE nopol 16 10 7} & \rotatebox{90}{CenSurE nopol 32 20 5} & \rotatebox{90}{CenSurE nopol 8 10 3} & \rotatebox{90}{SIFT nopol 0.02 20.0 1.0} & \rotatebox{90}{SIFT nopol 0.02 20.0 1.6} & \rotatebox{90}{SIFT nopol 0.04 5.0 1.6} & \rotatebox{90}{SIFT nopol 0.04 5.0 3.0} & \rotatebox{90}{SIFT nopol 0.08 5.0 3.0} & \rotatebox{90}{SIFT pol 0.02 20.0 1.0} & \rotatebox{90}{SIFT pol 0.02 20.0 1.6} & \rotatebox{90}{SIFT pol 0.02 5.0 3.0} & \rotatebox{90}{SIFT pol 0.04 5.0 3.0} \\
\hline
\hline
0.005 & 12 & 37 & 8 & \underline{1237} & \textbf{1239} & 329 & 219 & 59 & 847 & 903 & 478 & 210 \\
\hline
0.01 & 47 & 113 & 30 & \textbf{3559} & \underline{3255} & 827 & 540 & 156 & 2752 & 2729 & 1324 & 661 \\
\hline
0.02 & 145 & 306 & 110 & \textbf{6732} & 5838 & 1386 & 911 & 274 & \underline{6090} & 5552 & 2173 & 1132 \\
\hline
0.03 & 221 & 475 & 175 & \textbf{8802} & 7384 & 1672 & 1081 & 326 & \underline{8475} & 7510 & 2603 & 1370 \\
\hline
0.04 & 273 & 605 & 216 & \underline{10205} & 8419 & 1841 & 1174 & 360 & \textbf{10238} & 8912 & 2857 & 1516 \\
\hline
0.05 & 313 & 693 & 237 & \underline{11150} & 9033 & 1932 & 1233 & 373 & \textbf{11616} & 9955 & 3014 & 1611 \\
\hline
0.06 & 329 & 751 & 242 & \underline{11960} & 9570 & 1986 & 1271 & 385 & \textbf{12707} & 10777 & 3122 & 1672 \\
\hline
0.07 & 342 & 789 & 256 & \underline{12559} & 10003 & 2024 & 1297 & 391 & \textbf{13596} & 11437 & 3194 & 1704 \\
\hline
0.08 & 354 & 824 & 261 & \underline{13036} & 10339 & 2060 & 1324 & 398 & \textbf{14325} & 11949 & 3249 & 1729 \\
\hline
0.09 & 363 & 850 & 265 & \underline{13414} & 10608 & 2088 & 1339 & 400 & \textbf{14892} & 12358 & 3310 & 1755 \\
\hline
0.1 & 369 & 866 & 267 & \underline{13762} & 10862 & 2109 & 1352 & 403 & \textbf{15422} & 12685 & 3362 & 1775 \\
\hline
0.15 & 382 & 913 & 273 & \underline{15173} & 11885 & 2196 & 1411 & 410 & \textbf{17384} & 14007 & 3591 & 1840 \\
\end{tabular}
\end{table}

%% file: 2014-04-07-plot-v2/mix_detecteurs_0_repetabilite_best_tab.tex
\begin{table}[htbp]
\caption{Repeatability for pair 0 "translation", numerical values. Bold values are the best for a given tolerance, underlined are second best.
\label{tab:repetabilite-image-0}}
\centering
\begin{tabular}{l*{12}{|c}}
\multicolumn{1}{c|}{\rotatebox{90}{Distance tolerance (m)}} & \rotatebox{90}{CenSurE nopol 16 10 7} & \rotatebox{90}{CenSurE nopol 32 20 5} & \rotatebox{90}{CenSurE nopol 8 10 3} & \rotatebox{90}{SIFT nopol 0.02 20.0 1.0} & \rotatebox{90}{SIFT nopol 0.02 20.0 1.6} & \rotatebox{90}{SIFT nopol 0.04 5.0 1.6} & \rotatebox{90}{SIFT nopol 0.04 5.0 3.0} & \rotatebox{90}{SIFT nopol 0.08 5.0 3.0} & \rotatebox{90}{SIFT pol 0.02 20.0 1.0} & \rotatebox{90}{SIFT pol 0.02 20.0 1.6} & \rotatebox{90}{SIFT pol 0.02 5.0 3.0} & \rotatebox{90}{SIFT pol 0.04 5.0 3.0} \\
\hline
\hline
0.005 & 0.02 & 0.03 & 0.02 & 0.06 & 0.07 & 0.10 & \textbf{0.11} & \underline{0.11} & 0.03 & 0.04 & 0.08 & 0.08 \\
\hline
0.01 & 0.09 & 0.09 & 0.09 & 0.17 & 0.19 & 0.26 & \underline{0.26} & \textbf{0.28} & 0.11 & 0.13 & 0.23 & 0.25 \\
\hline
0.02 & 0.29 & 0.24 & 0.32 & 0.31 & 0.34 & 0.44 & \underline{0.45} & \textbf{0.49} & 0.23 & 0.27 & 0.38 & 0.43 \\
\hline
0.03 & 0.45 & 0.37 & 0.51 & 0.41 & 0.43 & 0.53 & \underline{0.53} & \textbf{0.58} & 0.33 & 0.37 & 0.46 & 0.52 \\
\hline
0.04 & 0.55 & 0.48 & \underline{0.63} & 0.48 & 0.49 & 0.58 & 0.58 & \textbf{0.64} & 0.39 & 0.44 & 0.50 & 0.58 \\
\hline
0.05 & 0.63 & 0.55 & \textbf{0.69} & 0.52 & 0.53 & 0.61 & 0.61 & \underline{0.67} & 0.45 & 0.49 & 0.53 & 0.61 \\
\hline
0.06 & 0.66 & 0.59 & \textbf{0.71} & 0.56 & 0.56 & 0.63 & 0.62 & \underline{0.69} & 0.49 & 0.53 & 0.55 & 0.64 \\
\hline
0.07 & 0.69 & 0.62 & \textbf{0.75} & 0.59 & 0.58 & 0.64 & 0.64 & \underline{0.70} & 0.52 & 0.56 & 0.56 & 0.65 \\
\hline
0.08 & \underline{0.71} & 0.65 & \textbf{0.76} & 0.61 & 0.60 & 0.65 & 0.65 & 0.71 & 0.55 & 0.58 & 0.57 & 0.66 \\
\hline
0.09 & \underline{0.73} & 0.67 & \textbf{0.77} & 0.62 & 0.62 & 0.66 & 0.66 & 0.72 & 0.57 & 0.60 & 0.58 & 0.67 \\
\hline
0.1 & \underline{0.74} & 0.68 & \textbf{0.78} & 0.64 & 0.64 & 0.67 & 0.66 & 0.72 & 0.59 & 0.62 & 0.59 & 0.68 \\
\hline
0.15 & \underline{0.77} & 0.72 & \textbf{0.80} & 0.71 & 0.70 & 0.70 & 0.69 & 0.73 & 0.67 & 0.68 & 0.63 & 0.70 \\
\end{tabular}
\end{table}

%% file: 2014-04-07-plot-v2/mix_detecteurs_40_nbcorresp_best_gphe.tex
\begin{figure}[htbp]
\centering
\includegraphics{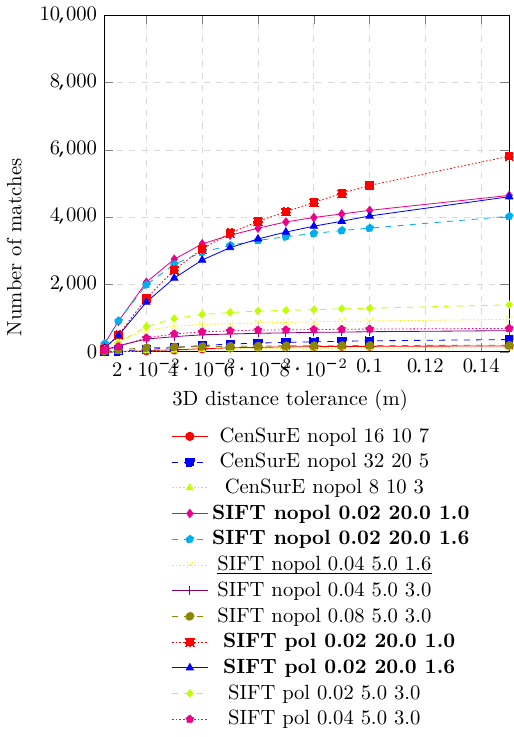}
\caption{Number of matches over distance tolerance for pair 40 "translation motion". Can be split easily into 2 clusters, the well performing setups are boldened in the legend.
Underlined setup in the legend has the best overall number of matches among the good performers in terms of repeatability highlighted in figure~\ref{fig:repetabilite-image-40}.}
\label{fig:nombre-de-correspondants-image-40}
\end{figure}

%% file: 2014-04-07-plot-v2/mix_detecteurs_40_repetabilite_best_gphe.tex
\begin{figure}[htbp]
\centering
\includegraphics{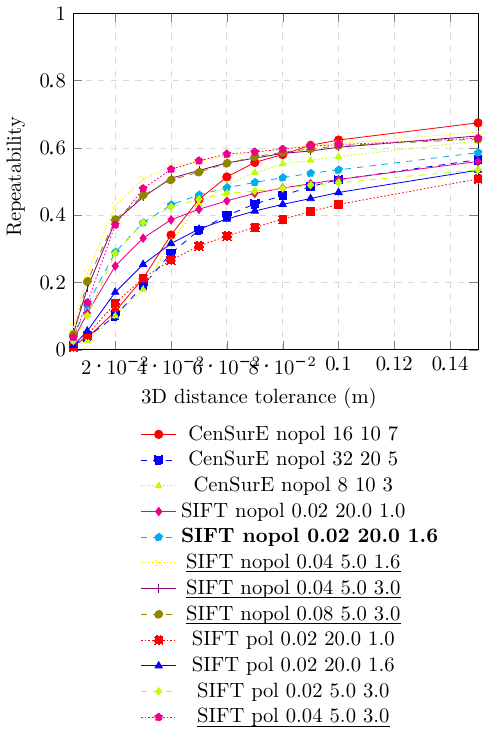}
\caption{Repeatability over distance tolerance for pair 40 "translation motion".
The best performing setups are underlined in the legend.
Boldened setup in the legend has the best overall repeatability among the good performers in terms of number of matches highlighted in figure~\ref{fig:nombre-de-correspondants-image-40}.}
\label{fig:repetabilite-image-40}
\end{figure}

%% file: 2014-04-07-plot-v2/mix_detecteurs_40_nbcorresp_best_tab.tex
\begin{table}[htbp]
\caption{Number of matches for pair 40 "translation motion", numerical values. Bold values are the best for a given tolerance, underlined are second best.
\label{tab:nombre-de-correspondants-image-40}}
\centering
\begin{tabular}{l*{12}{|c}}
\multicolumn{1}{c|}{\rotatebox{90}{Distance tolerance (m)}} & \rotatebox{90}{CenSurE nopol 16 10 7} & \rotatebox{90}{CenSurE nopol 32 20 5} & \rotatebox{90}{CenSurE nopol 8 10 3} & \rotatebox{90}{SIFT nopol 0.02 20.0 1.0} & \rotatebox{90}{SIFT nopol 0.02 20.0 1.6} & \rotatebox{90}{SIFT nopol 0.04 5.0 1.6} & \rotatebox{90}{SIFT nopol 0.04 5.0 3.0} & \rotatebox{90}{SIFT nopol 0.08 5.0 3.0} & \rotatebox{90}{SIFT pol 0.02 20.0 1.0} & \rotatebox{90}{SIFT pol 0.02 20.0 1.6} & \rotatebox{90}{SIFT pol 0.02 5.0 3.0} & \rotatebox{90}{SIFT pol 0.04 5.0 3.0} \\
\hline
\hline
0.005 & 4 & 8 & 2 & \textbf{238} & \textbf{238} & 99 & 47 & 14 & 89 & \underline{107} & 65 & 41 \\
\hline
0.01 & 9 & 24 & 5 & \underline{905} & \textbf{915} & 336 & 180 & 62 & 503 & 479 & 262 & 154 \\
\hline
0.02 & 29 & 64 & 19 & \textbf{2065} & \underline{1992} & 638 & 376 & 118 & 1586 & 1475 & 746 & 409 \\
\hline
0.03 & 54 & 122 & 34 & \textbf{2751} & \underline{2596} & 753 & 447 & 140 & 2435 & 2190 & 982 & 528 \\
\hline
0.04 & 87 & 183 & 61 & \textbf{3208} & 2960 & 810 & 502 & 154 & \underline{3057} & 2726 & 1105 & 591 \\
\hline
0.05 & 114 & 227 & 82 & \underline{3465} & 3162 & 832 & 525 & 161 & \textbf{3529} & 3096 & 1164 & 619 \\
\hline
0.06 & 131 & 255 & 93 & \underline{3675} & 3310 & 854 & 547 & 169 & \textbf{3873} & 3355 & 1209 & 641 \\
\hline
0.07 & 142 & 277 & 100 & \underline{3858} & 3419 & 874 & 561 & 175 & \textbf{4164} & 3561 & 1228 & 648 \\
\hline
0.08 & 148 & 293 & 105 & \underline{3990} & 3516 & 890 & 575 & 179 & \textbf{4437} & 3732 & 1249 & 657 \\
\hline
0.09 & 155 & 310 & 107 & \underline{4096} & 3605 & 903 & 582 & 182 & \textbf{4708} & 3880 & 1273 & 667 \\
\hline
0.1 & 159 & 323 & 109 & \underline{4203} & 3675 & 912 & 593 & 185 & \textbf{4944} & 4037 & 1293 & 672 \\
\hline
0.15 & 172 & 361 & 118 & \underline{4647} & 4027 & 962 & 626 & 191 & \textbf{5814} & 4609 & 1393 & 693 \\
\end{tabular}
\end{table}

%% file: 2014-04-07-plot-v2/mix_detecteurs_40_repetabilite_best_tab.tex
\begin{table}[htbp]
\caption{Repeatability for pair 40 "translation motion", numerical values. Bold values are the best for a given tolerance, underlined are second best.
\label{tab:repetabilite-image-40}}
\centering
\begin{tabular}{l*{12}{|c}}
\multicolumn{1}{c|}{\rotatebox{90}{Distance tolerance (m)}} & \rotatebox{90}{CenSurE nopol 16 10 7} & \rotatebox{90}{CenSurE nopol 32 20 5} & \rotatebox{90}{CenSurE nopol 8 10 3} & \rotatebox{90}{SIFT nopol 0.02 20.0 1.0} & \rotatebox{90}{SIFT nopol 0.02 20.0 1.6} & \rotatebox{90}{SIFT nopol 0.04 5.0 1.6} & \rotatebox{90}{SIFT nopol 0.04 5.0 3.0} & \rotatebox{90}{SIFT nopol 0.08 5.0 3.0} & \rotatebox{90}{SIFT pol 0.02 20.0 1.0} & \rotatebox{90}{SIFT pol 0.02 20.0 1.6} & \rotatebox{90}{SIFT pol 0.02 5.0 3.0} & \rotatebox{90}{SIFT pol 0.04 5.0 3.0} \\
\hline
\hline
0.005 & 0.02 & 0.01 & 0.01 & 0.03 & 0.03 & \textbf{0.07} & 0.05 & \underline{0.05} & 0.01 & 0.01 & 0.02 & 0.04 \\
\hline
0.01 & 0.04 & 0.04 & 0.03 & 0.11 & 0.13 & \textbf{0.23} & 0.18 & \underline{0.20} & 0.04 & 0.06 & 0.10 & 0.14 \\
\hline
0.02 & 0.11 & 0.10 & 0.10 & 0.25 & 0.29 & \textbf{0.43} & 0.38 & \underline{0.39} & 0.14 & 0.17 & 0.29 & 0.37 \\
\hline
0.03 & 0.21 & 0.19 & 0.18 & 0.33 & 0.38 & \textbf{0.51} & 0.45 & 0.46 & 0.21 & 0.25 & 0.38 & \underline{0.48} \\
\hline
0.04 & 0.34 & 0.29 & 0.32 & 0.39 & 0.43 & \textbf{0.54} & 0.51 & 0.50 & 0.27 & 0.32 & 0.42 & \underline{0.54} \\
\hline
0.05 & 0.45 & 0.36 & 0.43 & 0.42 & 0.46 & \underline{0.56} & 0.53 & 0.53 & 0.31 & 0.36 & 0.45 & \textbf{0.56} \\
\hline
0.06 & 0.51 & 0.40 & 0.49 & 0.44 & 0.48 & \underline{0.57} & 0.56 & 0.55 & 0.34 & 0.39 & 0.46 & \textbf{0.58} \\
\hline
0.07 & 0.56 & 0.43 & 0.53 & 0.46 & 0.50 & \underline{0.59} & 0.57 & 0.57 & 0.36 & 0.41 & 0.47 & \textbf{0.59} \\
\hline
0.08 & 0.58 & 0.46 & 0.55 & 0.48 & 0.51 & \textbf{0.60} & 0.58 & 0.59 & 0.39 & 0.43 & 0.48 & \underline{0.60} \\
\hline
0.09 & \textbf{0.61} & 0.49 & 0.56 & 0.49 & 0.52 & 0.61 & 0.59 & 0.60 & 0.41 & 0.45 & 0.49 & \underline{0.61} \\
\hline
0.1 & \textbf{0.62} & 0.51 & 0.57 & 0.51 & 0.53 & \underline{0.61} & 0.60 & 0.61 & 0.43 & 0.47 & 0.50 & 0.61 \\
\hline
0.15 & \textbf{0.67} & 0.56 & 0.62 & 0.56 & 0.59 & \underline{0.65} & 0.64 & 0.63 & 0.51 & 0.53 & 0.53 & 0.63 \\
\end{tabular}
\end{table}

%% file: 2014-04-07-plot-v2/mix_detecteurs_24_nbcorresp_best_gphe.tex
\begin{figure}[htbp]
\centering
\includegraphics{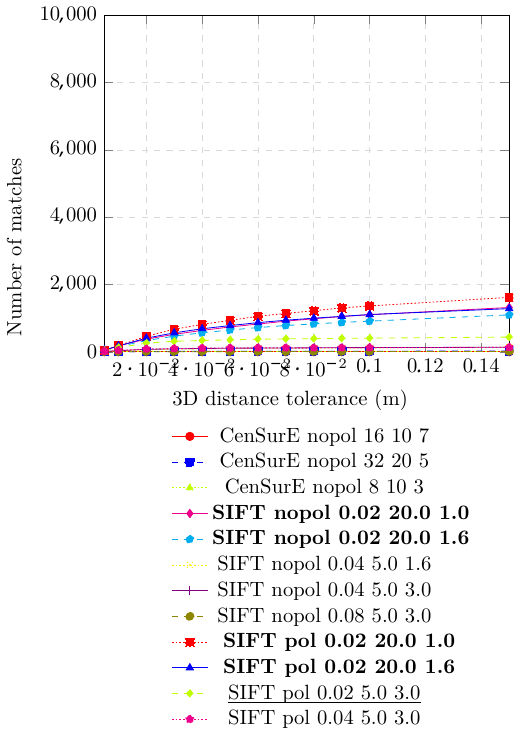}
\caption{Number of matches over distance tolerance for pair 24 "translation and rotation motion". Can be split into 2 clusters, the well performing setups are boldened in the legend.
Underlined setup in the legend has the best overall number of matches among the good performers in terms of repeatability highlighted in figure~\ref{fig:repetabilite-image-24}.}
\label{fig:nombre-de-correspondants-image-24}
\end{figure}

%% file: 2014-04-07-plot-v2/mix_detecteurs_24_repetabilite_best_gphe.tex
\begin{figure}[htbp]
\centering
\includegraphics{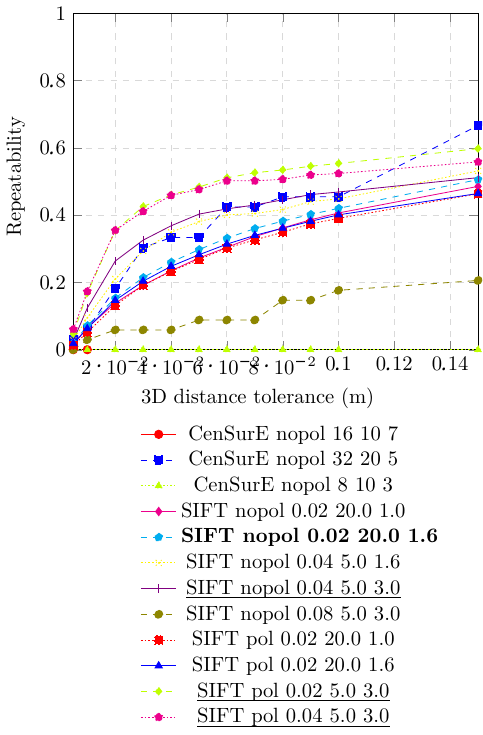}
\caption{Repeatability over distance tolerance for pair 24 "translation and rotation motion".
The best performing setups are underlined in the legend.
Interestingly, the two low performing setups here where among the betters with pair 0 "translation" figure~\ref{fig:repetabilite-image-0} and pair 40 "translation motion" figure~\ref{fig:repetabilite-image-40}.
Boldened setup in the legend has the best overall repeatability among the good performers in terms of number of matches highlighted in figure~\ref{fig:nombre-de-correspondants-image-24}.}
\label{fig:repetabilite-image-24}
\end{figure}

%% file: 2014-04-07-plot-v2/mix_detecteurs_24_nbcorresp_best_tab.tex
\begin{table}[htbp]
\caption{Number of matches for pair 24 "translation and rotation motion", numerical values. Bold values are the best for a given tolerance, underlined are second best.
\label{tab:nombre-de-correspondants-image-24}}
\centering
\begin{tabular}{l*{12}{|c}}
\multicolumn{1}{c|}{\rotatebox{90}{Distance tolerance (m)}} & \rotatebox{90}{CenSurE nopol 16 10 7} & \rotatebox{90}{CenSurE nopol 32 20 5} & \rotatebox{90}{CenSurE nopol 8 10 3} & \rotatebox{90}{SIFT nopol 0.02 20.0 1.0} & \rotatebox{90}{SIFT nopol 0.02 20.0 1.6} & \rotatebox{90}{SIFT nopol 0.04 5.0 1.6} & \rotatebox{90}{SIFT nopol 0.04 5.0 3.0} & \rotatebox{90}{SIFT nopol 0.08 5.0 3.0} & \rotatebox{90}{SIFT pol 0.02 20.0 1.0} & \rotatebox{90}{SIFT pol 0.02 20.0 1.6} & \rotatebox{90}{SIFT pol 0.02 5.0 3.0} & \rotatebox{90}{SIFT pol 0.04 5.0 3.0} \\
\hline
\hline
0.005 & 0 & 1 & 0 & \underline{46} & 43 & 7 & 7 & 0 & 47 & \textbf{50} & 34 & 14 \\
\hline
0.01 & 0 & 2 & 0 & \underline{177} & 155 & 24 & 32 & 1 & \textbf{180} & 173 & 122 & 40 \\
\hline
0.02 & 1 & 6 & 0 & 375 & 332 & 53 & 68 & 2 & \textbf{459} & \underline{405} & 257 & 82 \\
\hline
0.03 & 1 & 10 & 0 & 518 & 464 & 74 & 84 & 2 & \textbf{667} & \underline{562} & 309 & 95 \\
\hline
0.04 & 1 & 11 & 0 & 635 & 563 & 88 & 95 & 2 & \textbf{811} & \underline{684} & 334 & 106 \\
\hline
0.05 & 1 & 11 & 0 & 738 & 644 & 96 & 104 & 3 & \textbf{935} & \underline{779} & 351 & 110 \\
\hline
0.06 & 1 & 14 & 0 & 824 & 717 & 101 & 108 & 3 & \textbf{1053} & \underline{866} & 371 & 116 \\
\hline
0.07 & 1 & 14 & 0 & 910 & 779 & 102 & 111 & 3 & \textbf{1138} & \underline{938} & 382 & 116 \\
\hline
0.08 & 1 & 15 & 0 & 980 & 827 & 105 & 115 & 5 & \textbf{1217} & \underline{997} & 388 & 117 \\
\hline
0.09 & 1 & 15 & 0 & 1045 & 870 & 111 & 119 & 5 & \textbf{1304} & \underline{1053} & 396 & 120 \\
\hline
0.1 & 1 & 15 & 0 & 1100 & 909 & 113 & 121 & 6 & \textbf{1360} & \underline{1106} & 402 & 121 \\
\hline
0.15 & 1 & 22 & 0 & \underline{1314} & 1094 & 134 & 132 & 7 & \textbf{1616} & 1281 & 434 & 129 \\
\end{tabular}
\end{table}

%% file: 2014-04-07-plot-v2/mix_detecteurs_24_repetabilite_best_tab.tex
\begin{table}[htbp]
\caption{Repeatability for pair 24 "translation and rotation motion", numerical values. Bold values are the best for a given tolerance, underlined are second best.
-1 value means repeatability is 1 but meaningless as there is only 1 match for 1 detection only in an image of the pair.
\label{tab:repetabilite-image-24}}
\centering
\begin{tabular}{l*{12}{|c}}
\multicolumn{1}{c|}{\rotatebox{90}{Distance tolerance (m)}} & \rotatebox{90}{CenSurE nopol 16 10 7} & \rotatebox{90}{CenSurE nopol 32 20 5} & \rotatebox{90}{CenSurE nopol 8 10 3} & \rotatebox{90}{SIFT nopol 0.02 20.0 1.0} & \rotatebox{90}{SIFT nopol 0.02 20.0 1.6} & \rotatebox{90}{SIFT nopol 0.04 5.0 1.6} & \rotatebox{90}{SIFT nopol 0.04 5.0 3.0} & \rotatebox{90}{SIFT nopol 0.08 5.0 3.0} & \rotatebox{90}{SIFT pol 0.02 20.0 1.0} & \rotatebox{90}{SIFT pol 0.02 20.0 1.6} & \rotatebox{90}{SIFT pol 0.02 5.0 3.0} & \rotatebox{90}{SIFT pol 0.04 5.0 3.0} \\
\hline
\hline
0.005 & 0.00 & 0.03 & 0.00 & 0.02 & 0.02 & 0.03 & 0.03 & 0.00 & 0.01 & 0.02 & \underline{0.05} & \textbf{0.06} \\
\hline
0.01 & 0.00 & 0.06 & 0.00 & 0.07 & 0.07 & 0.10 & 0.12 & 0.03 & 0.05 & 0.06 & \underline{0.17} & \textbf{0.17} \\
\hline
0.02 & -1 & 0.18 & 0.00 & 0.14 & 0.15 & 0.21 & 0.26 & 0.06 & 0.13 & 0.15 & \underline{0.35} & \textbf{0.35} \\
\hline
0.03 & -1 & 0.30 & 0.00 & 0.19 & 0.21 & 0.29 & 0.33 & 0.06 & 0.19 & 0.20 & \textbf{0.43} & \underline{0.41} \\
\hline
0.04 & -1 & 0.33 & 0.00 & 0.23 & 0.26 & 0.35 & 0.37 & 0.06 & 0.23 & 0.25 & \textbf{0.46} & \underline{0.46} \\
\hline
0.05 & -1 & 0.33 & 0.00 & 0.27 & 0.30 & 0.38 & 0.40 & 0.09 & 0.27 & 0.28 & \textbf{0.48} & \underline{0.48} \\
\hline
0.06 & -1 & 0.42 & 0.00 & 0.30 & 0.33 & 0.40 & 0.42 & 0.09 & 0.30 & 0.31 & \textbf{0.51} & \underline{0.50} \\
\hline
0.07 & -1 & 0.42 & 0.00 & 0.34 & 0.36 & 0.40 & 0.43 & 0.09 & 0.33 & 0.34 & \textbf{0.53} & \underline{0.50} \\
\hline
0.08 & -1 & 0.45 & 0.00 & 0.36 & 0.38 & 0.42 & 0.45 & 0.15 & 0.35 & 0.36 & \textbf{0.53} & \underline{0.51} \\
\hline
0.09 & -1 & 0.45 & 0.00 & 0.39 & 0.40 & 0.44 & 0.46 & 0.15 & 0.37 & 0.38 & \textbf{0.55} & \underline{0.52} \\
\hline
0.1 & -1 & 0.45 & 0.00 & 0.41 & 0.42 & 0.45 & 0.47 & 0.18 & 0.39 & 0.40 & \textbf{0.55} & \underline{0.52} \\
\hline
0.15 & -1 & \textbf{0.67} & 0.00 & 0.49 & 0.51 & 0.53 & 0.51 & 0.21 & 0.46 & 0.46 & \underline{0.60} & 0.56 \\
\end{tabular}
\end{table}

%% file: 2014-04-07-plot-v2/mix_det-desc-match_SIFT_0_bonscorresp_best_gphe.tex
\begin{figure}[htbp]
\centering
\includegraphics{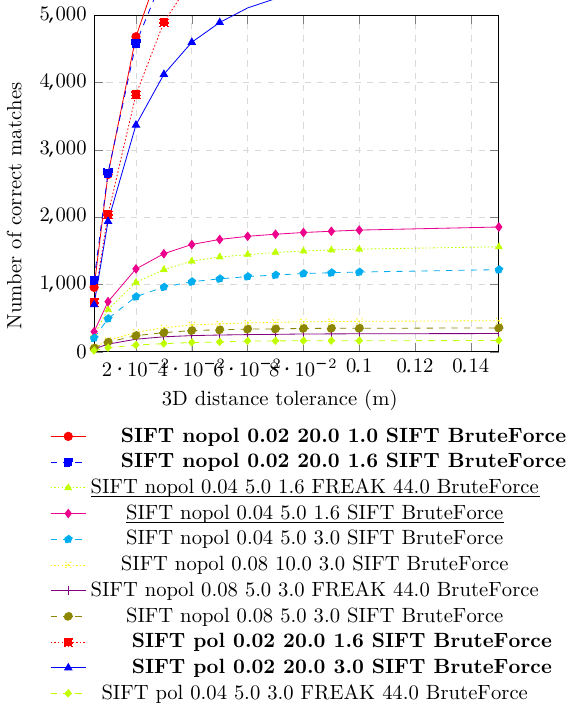}
\caption{Number of correct matches with SIFT setups over distance tolerance for pair 0 "translation". Can be split easily into 2 clusters, the well performing setups are boldened in the legend.
Underlined setups in the legend have the best overall number of correct matches among the good performers in terms of matching score highlighted in figure~\ref{fig:matching-score-image-0-sift}.}
\label{fig:nombre-de-bons-correspondants-image-0-sift}
\end{figure}

%% file: 2014-04-07-plot-v2/mix_det-desc-match_SIFT_0_mscore_best_gphe.tex
\begin{figure}[htbp]
\centering
\includegraphics{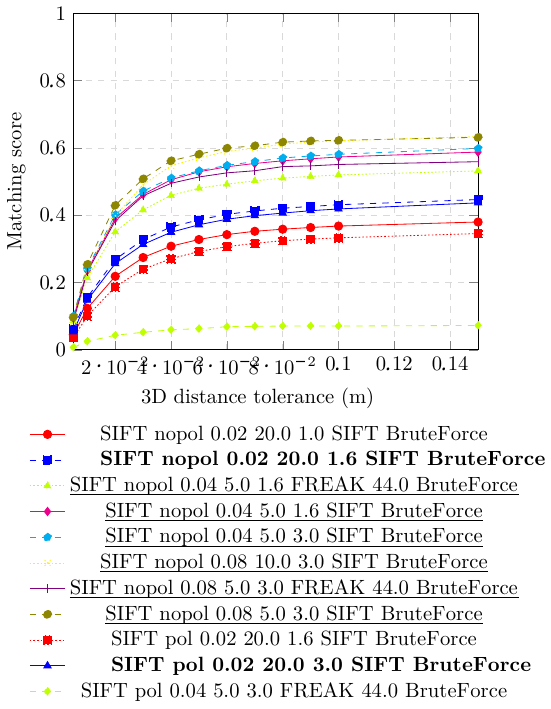}
\caption{Matching score with SIFT setups over distance tolerance for pair 0 "translation".
The best performing setups are underlined in the legend.
Boldened setups in the legend have the best overall matching score among the good performers in terms of number of correct matches highlighted in figure~\ref{fig:nombre-de-bons-correspondants-image-0-sift}.}
\label{fig:matching-score-image-0-sift}
\end{figure}

%% file: 2014-04-07-plot-v2/mix_det-desc-match_SIFT_0_bonscorresp_best_tab.tex
\begin{table}[htbp]
\caption{Number of correct matches with SIFT setups for pair 0 "translation", numerical values. Bold values are the best for a given tolerance, underlined are second best.
\label{tab:nombre-de-bons-correspondants-image-0-sift}}
\centering
\begin{tabular}{l*{11}{|c}}
\multicolumn{1}{c|}{\rotatebox{90}{Distance tolerance (m)}} & \rotatebox{90}{SIFT nopol 0.02 20.0 1.0 SIFT BruteForce} & \rotatebox{90}{SIFT nopol 0.02 20.0 1.6 SIFT BruteForce} & \rotatebox{90}{SIFT nopol 0.04 5.0 1.6 FREAK 44.0 BruteForce} & \rotatebox{90}{SIFT nopol 0.04 5.0 1.6 SIFT BruteForce} & \rotatebox{90}{SIFT nopol 0.04 5.0 3.0 SIFT BruteForce} & \rotatebox{90}{SIFT nopol 0.08 10.0 3.0 SIFT BruteForce} & \rotatebox{90}{SIFT nopol 0.08 5.0 3.0 FREAK 44.0 BruteForce} & \rotatebox{90}{SIFT nopol 0.08 5.0 3.0 SIFT BruteForce} & \rotatebox{90}{SIFT pol 0.02 20.0 1.6 SIFT BruteForce} & \rotatebox{90}{SIFT pol 0.02 20.0 3.0 SIFT BruteForce} & \rotatebox{90}{SIFT pol 0.04 5.0 3.0 FREAK 44.0 BruteForce} \\
\hline
\hline
0.005 & \underline{957} & \textbf{1064} & 250 & 297 & 204 & 59 & 41 & 54 & 737 & 699 & 16 \\
\hline
0.01 & \underline{2636} & \textbf{2657} & 627 & 744 & 492 & 167 & 112 & 142 & 2042 & 1939 & 57 \\
\hline
0.02 & \textbf{4683} & \underline{4585} & 1029 & 1231 & 818 & 289 & 185 & 240 & 3824 & 3371 & 99 \\
\hline
0.03 & \textbf{5883} & \underline{5598} & 1220 & 1457 & 961 & 356 & 221 & 284 & 4899 & 4125 & 119 \\
\hline
0.04 & \textbf{6605} & \underline{6239} & 1346 & 1594 & 1040 & 397 & 239 & 314 & 5529 & 4601 & 136 \\
\hline
0.05 & \textbf{7034} & \underline{6603} & 1409 & 1669 & 1084 & 414 & 248 & 325 & 5972 & 4896 & 144 \\
\hline
0.06 & \textbf{7345} & \underline{6883} & 1444 & 1716 & 1117 & 428 & 254 & 335 & 6268 & 5112 & 157 \\
\hline
0.07 & \textbf{7558} & \underline{7054} & 1474 & 1747 & 1140 & 435 & 257 & 339 & 6469 & 5251 & 160 \\
\hline
0.08 & \textbf{7691} & \underline{7192} & 1497 & 1772 & 1162 & 444 & 263 & 345 & 6627 & 5354 & 162 \\
\hline
0.09 & \textbf{7799} & \underline{7282} & 1512 & 1791 & 1174 & 448 & 264 & 347 & 6723 & 5447 & 162 \\
\hline
0.1 & \textbf{7891} & \underline{7370} & 1525 & 1808 & 1184 & 451 & 266 & 348 & 6804 & 5515 & 162 \\
\hline
0.15 & \textbf{8149} & \underline{7634} & 1560 & 1854 & 1220 & 460 & 270 & 353 & 7061 & 5754 & 166 \\
\end{tabular}
\end{table}

%% file: 2014-04-07-plot-v2/mix_det-desc-match_SIFT_0_mscore_best_tab.tex
\begin{table}[htbp]
\caption{Matching score with SIFT setups for pair 0 "translation", numerical values. Bold values are the best for a given tolerance, underlined are second best.
\label{tab:matching-score-image-0-sift}}
\centering
\begin{tabular}{l*{11}{|c}}
\multicolumn{1}{c|}{\rotatebox{90}{Distance tolerance (m)}} & \rotatebox{90}{SIFT nopol 0.02 20.0 1.0 SIFT BruteForce} & \rotatebox{90}{SIFT nopol 0.02 20.0 1.6 SIFT BruteForce} & \rotatebox{90}{SIFT nopol 0.04 5.0 1.6 FREAK 44.0 BruteForce} & \rotatebox{90}{SIFT nopol 0.04 5.0 1.6 SIFT BruteForce} & \rotatebox{90}{SIFT nopol 0.04 5.0 3.0 SIFT BruteForce} & \rotatebox{90}{SIFT nopol 0.08 10.0 3.0 SIFT BruteForce} & \rotatebox{90}{SIFT nopol 0.08 5.0 3.0 FREAK 44.0 BruteForce} & \rotatebox{90}{SIFT nopol 0.08 5.0 3.0 SIFT BruteForce} & \rotatebox{90}{SIFT pol 0.02 20.0 1.6 SIFT BruteForce} & \rotatebox{90}{SIFT pol 0.02 20.0 3.0 SIFT BruteForce} & \rotatebox{90}{SIFT pol 0.04 5.0 3.0 FREAK 44.0 BruteForce} \\
\hline
\hline
0.005 & 0.04 & 0.06 & 0.09 & 0.09 & \textbf{0.10} & 0.08 & 0.08 & \underline{0.10} & 0.04 & 0.05 & 0.01 \\
\hline
0.01 & 0.12 & 0.16 & 0.21 & 0.24 & \underline{0.24} & 0.23 & 0.23 & \textbf{0.25} & 0.10 & 0.15 & 0.02 \\
\hline
0.02 & 0.22 & 0.27 & 0.35 & 0.39 & 0.40 & \underline{0.40} & 0.38 & \textbf{0.43} & 0.19 & 0.26 & 0.04 \\
\hline
0.03 & 0.27 & 0.33 & 0.42 & 0.46 & 0.47 & \underline{0.49} & 0.46 & \textbf{0.51} & 0.24 & 0.31 & 0.05 \\
\hline
0.04 & 0.31 & 0.36 & 0.46 & 0.51 & 0.51 & \underline{0.55} & 0.49 & \textbf{0.56} & 0.27 & 0.35 & 0.06 \\
\hline
0.05 & 0.33 & 0.39 & 0.48 & 0.53 & 0.53 & \underline{0.57} & 0.51 & \textbf{0.58} & 0.29 & 0.37 & 0.06 \\
\hline
0.06 & 0.34 & 0.40 & 0.49 & 0.54 & 0.55 & \underline{0.59} & 0.53 & \textbf{0.60} & 0.31 & 0.39 & 0.07 \\
\hline
0.07 & 0.35 & 0.41 & 0.50 & 0.55 & 0.56 & \underline{0.60} & 0.53 & \textbf{0.61} & 0.32 & 0.40 & 0.07 \\
\hline
0.08 & 0.36 & 0.42 & 0.51 & 0.56 & 0.57 & \underline{0.61} & 0.54 & \textbf{0.62} & 0.32 & 0.41 & 0.07 \\
\hline
0.09 & 0.36 & 0.43 & 0.52 & 0.57 & 0.58 & \underline{0.62} & 0.55 & \textbf{0.62} & 0.33 & 0.41 & 0.07 \\
\hline
0.1 & 0.37 & 0.43 & 0.52 & 0.57 & 0.58 & \underline{0.62} & 0.55 & \textbf{0.62} & 0.33 & 0.42 & 0.07 \\
\hline
0.15 & 0.38 & 0.45 & 0.53 & 0.59 & 0.60 & \textbf{0.63} & 0.56 & \underline{0.63} & 0.35 & 0.44 & 0.07 \\
\end{tabular}
\end{table}

%% file: 2014-04-07-plot-v2/mix_det-desc-match_SIFT_40_bonscorresp_best_gphe.tex
\begin{figure}[htbp]
\centering
\includegraphics{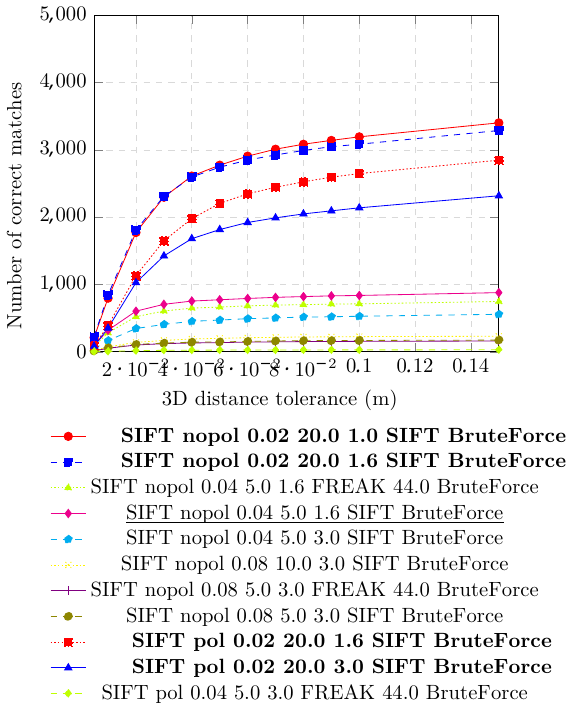}
\caption{Number of correct matches with SIFT setups over distance tolerance for pair 40 "translation motion". Can be split easily into 2 clusters, the well performing setups are boldened in the legend.
Underlined setup in the legend has the best overall number of correct matches among the good performers in terms of matching score highlighted in figure~\ref{fig:matching-score-image-40-sift}.}
\label{fig:nombre-de-bons-correspondants-image-40-sift}
\end{figure}

%% file: 2014-04-07-plot-v2/mix_det-desc-match_SIFT_40_mscore_best_gphe.tex
\begin{figure}[htbp]
\centering
\includegraphics{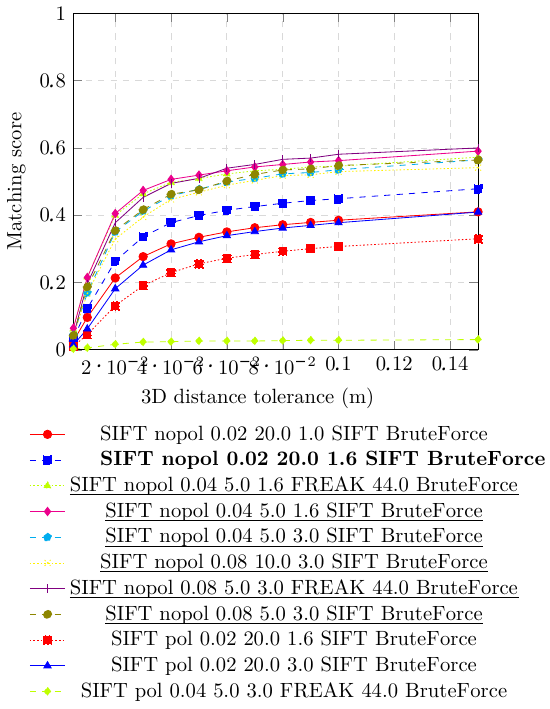}
\caption{Matching score with SIFT setups over distance tolerance for pair 40 "translation motion".
The best performing setups are underlined in the legend.
Boldened setup in the legend has the best overall matching score among the good performers in terms of number of correct matches highlighted in figure~\ref{fig:nombre-de-bons-correspondants-image-40-sift}.}
\label{fig:matching-score-image-40-sift}
\end{figure}

%% file: 2014-04-07-plot-v2/mix_det-desc-match_SIFT_40_bonscorresp_best_tab.tex
\begin{table}[htbp]
\caption{Number of correct matches with SIFT setups for pair 40 "translation motion", numerical values. Bold values are the best for a given tolerance, underlined are second best.
\label{tab:nombre-de-bons-correspondants-image-40-sift}}
\centering
\begin{tabular}{l*{11}{|c}}
\multicolumn{1}{c|}{\rotatebox{90}{Distance tolerance (m)}} & \rotatebox{90}{SIFT nopol 0.02 20.0 1.0 SIFT BruteForce} & \rotatebox{90}{SIFT nopol 0.02 20.0 1.6 SIFT BruteForce} & \rotatebox{90}{SIFT nopol 0.04 5.0 1.6 FREAK 44.0 BruteForce} & \rotatebox{90}{SIFT nopol 0.04 5.0 1.6 SIFT BruteForce} & \rotatebox{90}{SIFT nopol 0.04 5.0 3.0 SIFT BruteForce} & \rotatebox{90}{SIFT nopol 0.08 10.0 3.0 SIFT BruteForce} & \rotatebox{90}{SIFT nopol 0.08 5.0 3.0 FREAK 44.0 BruteForce} & \rotatebox{90}{SIFT nopol 0.08 5.0 3.0 SIFT BruteForce} & \rotatebox{90}{SIFT pol 0.02 20.0 1.6 SIFT BruteForce} & \rotatebox{90}{SIFT pol 0.02 20.0 3.0 SIFT BruteForce} & \rotatebox{90}{SIFT pol 0.04 5.0 3.0 FREAK 44.0 BruteForce} \\
\hline
\hline
0.005 & \underline{211} & \textbf{222} & 79 & 95 & 43 & 17 & 11 & 13 & 89 & 82 & 2 \\
\hline
0.01 & \underline{796} & \textbf{843} & 281 & 319 & 166 & 71 & 52 & 57 & 388 & 350 & 5 \\
\hline
0.02 & \underline{1772} & \textbf{1809} & 521 & 602 & 345 & 139 & 100 & 108 & 1129 & 1027 & 15 \\
\hline
0.03 & \underline{2297} & \textbf{2309} & 604 & 704 & 406 & 167 & 120 & 127 & 1649 & 1424 & 21 \\
\hline
0.04 & \textbf{2612} & \underline{2597} & 648 & 753 & 451 & 189 & 131 & 141 & 1978 & 1681 & 22 \\
\hline
0.05 & \textbf{2774} & \underline{2743} & 663 & 772 & 470 & 198 & 135 & 145 & 2205 & 1817 & 24 \\
\hline
0.06 & \textbf{2908} & \underline{2850} & 681 & 791 & 491 & 207 & 143 & 153 & 2347 & 1921 & 24 \\
\hline
0.07 & \textbf{3011} & \underline{2925} & 693 & 808 & 502 & 213 & 146 & 159 & 2445 & 1990 & 24 \\
\hline
0.08 & \textbf{3085} & \underline{2992} & 699 & 819 & 515 & 219 & 150 & 163 & 2526 & 2049 & 25 \\
\hline
0.09 & \textbf{3142} & \underline{3049} & 708 & 830 & 519 & 220 & 151 & 164 & 2595 & 2094 & 26 \\
\hline
0.1 & \textbf{3195} & \underline{3085} & 711 & 836 & 527 & 224 & 154 & 167 & 2649 & 2139 & 26 \\
\hline
0.15 & \textbf{3403} & \underline{3289} & 746 & 878 & 556 & 229 & 159 & 172 & 2848 & 2318 & 28 \\
\end{tabular}
\end{table}

%% file: 2014-04-07-plot-v2/mix_det-desc-match_SIFT_40_mscore_best_tab.tex
\begin{table}[htbp]
\caption{Matching score with SIFT setups for pair 40 "translation motion", numerical values. Bold values are the best for a given tolerance, underlined are second best.
\label{tab:matching-score-image-40-sift}}
\centering
\begin{tabular}{l*{11}{|c}}
\multicolumn{1}{c|}{\rotatebox{90}{Distance tolerance (m)}} & \rotatebox{90}{SIFT nopol 0.02 20.0 1.0 SIFT BruteForce} & \rotatebox{90}{SIFT nopol 0.02 20.0 1.6 SIFT BruteForce} & \rotatebox{90}{SIFT nopol 0.04 5.0 1.6 FREAK 44.0 BruteForce} & \rotatebox{90}{SIFT nopol 0.04 5.0 1.6 SIFT BruteForce} & \rotatebox{90}{SIFT nopol 0.04 5.0 3.0 SIFT BruteForce} & \rotatebox{90}{SIFT nopol 0.08 10.0 3.0 SIFT BruteForce} & \rotatebox{90}{SIFT nopol 0.08 5.0 3.0 FREAK 44.0 BruteForce} & \rotatebox{90}{SIFT nopol 0.08 5.0 3.0 SIFT BruteForce} & \rotatebox{90}{SIFT pol 0.02 20.0 1.6 SIFT BruteForce} & \rotatebox{90}{SIFT pol 0.02 20.0 3.0 SIFT BruteForce} & \rotatebox{90}{SIFT pol 0.04 5.0 3.0 FREAK 44.0 BruteForce} \\
\hline
\hline
0.005 & 0.03 & 0.03 & \underline{0.06} & \textbf{0.06} & 0.04 & 0.04 & 0.04 & 0.04 & 0.01 & 0.01 & 0.00 \\
\hline
0.01 & 0.10 & 0.12 & \textbf{0.22} & \underline{0.21} & 0.17 & 0.17 & 0.20 & 0.19 & 0.04 & 0.06 & 0.01 \\
\hline
0.02 & 0.21 & 0.26 & \underline{0.40} & \textbf{0.40} & 0.35 & 0.33 & 0.38 & 0.35 & 0.13 & 0.18 & 0.02 \\
\hline
0.03 & 0.28 & 0.34 & \underline{0.46} & \textbf{0.47} & 0.41 & 0.39 & 0.45 & 0.42 & 0.19 & 0.25 & 0.02 \\
\hline
0.04 & 0.31 & 0.38 & \underline{0.50} & \textbf{0.51} & 0.46 & 0.45 & 0.49 & 0.46 & 0.23 & 0.30 & 0.02 \\
\hline
0.05 & 0.33 & 0.40 & 0.51 & \textbf{0.52} & 0.48 & 0.47 & \underline{0.51} & 0.48 & 0.26 & 0.32 & 0.03 \\
\hline
0.06 & 0.35 & 0.41 & 0.52 & \underline{0.53} & 0.50 & 0.49 & \textbf{0.54} & 0.50 & 0.27 & 0.34 & 0.03 \\
\hline
0.07 & 0.36 & 0.43 & 0.53 & \underline{0.54} & 0.51 & 0.50 & \textbf{0.55} & 0.52 & 0.28 & 0.35 & 0.03 \\
\hline
0.08 & 0.37 & 0.44 & 0.54 & \underline{0.55} & 0.52 & 0.52 & \textbf{0.57} & 0.53 & 0.29 & 0.36 & 0.03 \\
\hline
0.09 & 0.38 & 0.44 & 0.54 & \underline{0.56} & 0.53 & 0.52 & \textbf{0.57} & 0.54 & 0.30 & 0.37 & 0.03 \\
\hline
0.1 & 0.38 & 0.45 & 0.55 & \underline{0.56} & 0.54 & 0.53 & \textbf{0.58} & 0.55 & 0.31 & 0.38 & 0.03 \\
\hline
0.15 & 0.41 & 0.48 & 0.57 & \underline{0.59} & 0.56 & 0.54 & \textbf{0.60} & 0.56 & 0.33 & 0.41 & 0.03 \\
\end{tabular}
\end{table}

%% file: 2014-04-07-plot-v2/mix_det-desc-match_SIFT_24_bonscorresp_best_gphe.tex
\begin{figure}[htbp]
\centering
\includegraphics{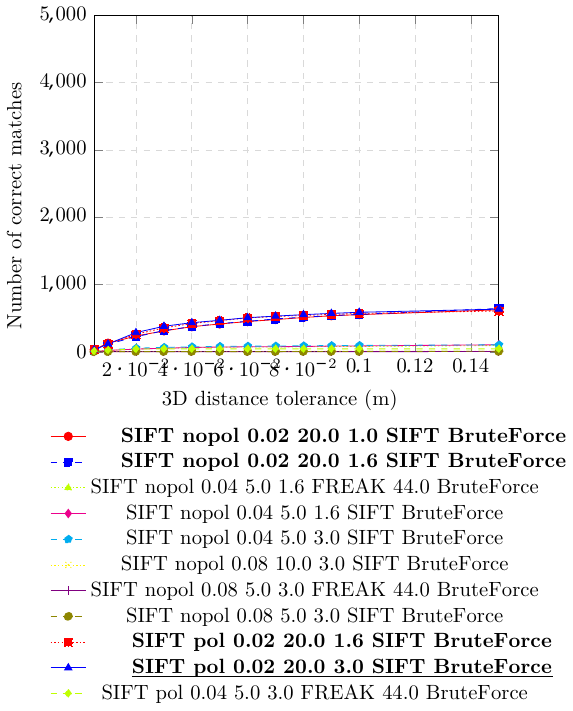}
\caption{Number of correct matches with SIFT setups over distance tolerance for pair 24 "translation and rotation motion". Can be split easily into 2 clusters, the well performing setups are boldened in the legend.
Underlined setup in the legend has the best overall number of correct matches among the good performers in terms of matching score highlighted in figure~\ref{fig:matching-score-image-24-sift}.}
\label{fig:nombre-de-bons-correspondants-image-24-sift}
\end{figure}

%% file: 2014-04-07-plot-v2/mix_det-desc-match_SIFT_24_mscore_best_gphe.tex
\begin{figure}[htbp]
\centering
\includegraphics{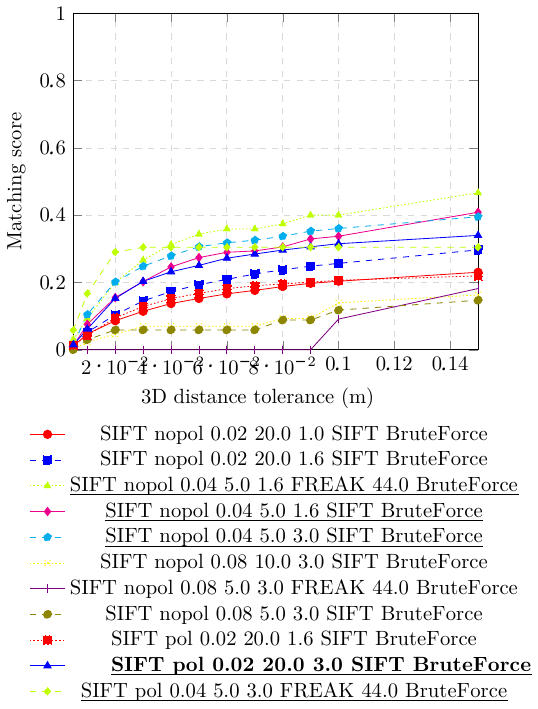}
\caption{Matching score with SIFT setups over distance tolerance for pair 24 "translation and rotation motion".
The best performing setups are underlined in the legend.
Boldened setup in the legend has the best overall matching score among the good performers in terms of number of correct matches highlighted in figure~\ref{fig:nombre-de-bons-correspondants-image-24-sift}.}
\label{fig:matching-score-image-24-sift}
\end{figure}

%% file: 2014-04-07-plot-v2/mix_det-desc-match_SIFT_24_bonscorresp_best_tab.tex
\begin{table}[htbp]
\caption{Number of correct matches with SIFT setups for pair 24 "translation and rotation motion", numerical values. Bold values are the best for a given tolerance, underlined are second best.
\label{tab:nombre-de-bons-correspondants-image-24-sift}}
\centering
\begin{tabular}{l*{11}{|c}}
\multicolumn{1}{c|}{\rotatebox{90}{Distance tolerance (m)}} & \rotatebox{90}{SIFT nopol 0.02 20.0 1.0 SIFT BruteForce} & \rotatebox{90}{SIFT nopol 0.02 20.0 1.6 SIFT BruteForce} & \rotatebox{90}{SIFT nopol 0.04 5.0 1.6 FREAK 44.0 BruteForce} & \rotatebox{90}{SIFT nopol 0.04 5.0 1.6 SIFT BruteForce} & \rotatebox{90}{SIFT nopol 0.04 5.0 3.0 SIFT BruteForce} & \rotatebox{90}{SIFT nopol 0.08 10.0 3.0 SIFT BruteForce} & \rotatebox{90}{SIFT nopol 0.08 5.0 3.0 FREAK 44.0 BruteForce} & \rotatebox{90}{SIFT nopol 0.08 5.0 3.0 SIFT BruteForce} & \rotatebox{90}{SIFT pol 0.02 20.0 1.6 SIFT BruteForce} & \rotatebox{90}{SIFT pol 0.02 20.0 3.0 SIFT BruteForce} & \rotatebox{90}{SIFT pol 0.04 5.0 3.0 FREAK 44.0 BruteForce} \\
\hline
\hline
0.005 & 28 & \underline{29} & 6 & 3 & 4 & 0 & 0 & 0 & \textbf{37} & 27 & 8 \\
\hline
0.01 & \textbf{127} & 112 & 18 & 19 & 27 & 1 & 0 & 1 & \underline{118} & 114 & 23 \\
\hline
0.02 & 233 & 224 & 39 & 39 & 52 & 2 & 0 & 2 & \underline{261} & \textbf{284} & 40 \\
\hline
0.03 & 309 & 313 & 52 & 51 & 64 & 3 & 0 & 2 & \underline{354} & \textbf{378} & 42 \\
\hline
0.04 & 371 & 374 & 61 & 62 & 72 & 3 & 0 & 2 & \underline{419} & \textbf{431} & 42 \\
\hline
0.05 & 412 & 418 & 67 & 69 & 79 & 3 & 0 & 2 & \underline{461} & \textbf{466} & 42 \\
\hline
0.06 & 449 & 452 & 70 & 73 & 82 & 3 & 0 & 2 & \underline{501} & \textbf{506} & 42 \\
\hline
0.07 & 477 & 487 & 70 & 74 & 84 & 3 & 0 & 2 & \underline{524} & \textbf{529} & 42 \\
\hline
0.08 & 508 & 514 & 73 & 77 & 87 & 4 & 0 & 3 & \underline{539} & \textbf{551} & 42 \\
\hline
0.09 & 534 & 537 & 78 & 83 & 91 & 4 & 0 & 3 & \underline{554} & \textbf{569} & 42 \\
\hline
0.1 & 551 & 555 & 78 & 85 & 93 & 6 & 1 & 4 & \underline{569} & \textbf{586} & 42 \\
\hline
0.15 & 623 & \textbf{642} & 91 & 103 & 102 & 7 & 2 & 5 & 605 & \underline{632} & 42 \\
\end{tabular}
\end{table}

%% file: 2014-04-07-plot-v2/mix_det-desc-match_SIFT_24_mscore_best_tab.tex
\begin{table}[htbp]
\caption{Matching score with SIFT setups for pair 24 "translation and rotation motion", numerical values. Bold values are the best for a given tolerance, underlined are second best.
\label{tab:matching-score-image-24-sift}}
\centering
\begin{tabular}{l*{11}{|c}}
\multicolumn{1}{c|}{\rotatebox{90}{Distance tolerance (m)}} & \rotatebox{90}{SIFT nopol 0.02 20.0 1.0 SIFT BruteForce} & \rotatebox{90}{SIFT nopol 0.02 20.0 1.6 SIFT BruteForce} & \rotatebox{90}{SIFT nopol 0.04 5.0 1.6 FREAK 44.0 BruteForce} & \rotatebox{90}{SIFT nopol 0.04 5.0 1.6 SIFT BruteForce} & \rotatebox{90}{SIFT nopol 0.04 5.0 3.0 SIFT BruteForce} & \rotatebox{90}{SIFT nopol 0.08 10.0 3.0 SIFT BruteForce} & \rotatebox{90}{SIFT nopol 0.08 5.0 3.0 FREAK 44.0 BruteForce} & \rotatebox{90}{SIFT nopol 0.08 5.0 3.0 SIFT BruteForce} & \rotatebox{90}{SIFT pol 0.02 20.0 1.6 SIFT BruteForce} & \rotatebox{90}{SIFT pol 0.02 20.0 3.0 SIFT BruteForce} & \rotatebox{90}{SIFT pol 0.04 5.0 3.0 FREAK 44.0 BruteForce} \\
\hline
\hline
0.005 & 0.01 & 0.01 & \underline{0.03} & 0.01 & 0.02 & 0.00 & 0.00 & 0.00 & 0.01 & 0.01 & \textbf{0.06} \\
\hline
0.01 & 0.05 & 0.05 & \underline{0.09} & 0.08 & 0.10 & 0.02 & 0.00 & 0.03 & 0.04 & 0.06 & \textbf{0.17} \\
\hline
0.02 & 0.09 & 0.10 & \underline{0.20} & 0.15 & 0.20 & 0.05 & 0.00 & 0.06 & 0.09 & 0.15 & \textbf{0.29} \\
\hline
0.03 & 0.11 & 0.14 & \underline{0.27} & 0.20 & 0.25 & 0.07 & 0.00 & 0.06 & 0.13 & 0.20 & \textbf{0.30} \\
\hline
0.04 & 0.14 & 0.17 & \textbf{0.31} & 0.25 & 0.28 & 0.07 & 0.00 & 0.06 & 0.15 & 0.23 & \underline{0.30} \\
\hline
0.05 & 0.15 & 0.19 & \textbf{0.34} & 0.27 & \underline{0.31} & 0.07 & 0.00 & 0.06 & 0.17 & 0.25 & 0.30 \\
\hline
0.06 & 0.17 & 0.21 & \textbf{0.36} & 0.29 & \underline{0.32} & 0.07 & 0.00 & 0.06 & 0.18 & 0.27 & 0.30 \\
\hline
0.07 & 0.18 & 0.23 & \textbf{0.36} & 0.29 & \underline{0.33} & 0.07 & 0.00 & 0.06 & 0.19 & 0.28 & 0.30 \\
\hline
0.08 & 0.19 & 0.24 & \textbf{0.37} & 0.31 & \underline{0.34} & 0.09 & 0.00 & 0.09 & 0.20 & 0.30 & 0.30 \\
\hline
0.09 & 0.20 & 0.25 & \textbf{0.40} & 0.33 & \underline{0.35} & 0.09 & 0.00 & 0.09 & 0.20 & 0.31 & 0.30 \\
\hline
0.1 & 0.20 & 0.26 & \textbf{0.40} & 0.34 & \underline{0.36} & 0.14 & 0.09 & 0.12 & 0.21 & 0.32 & 0.30 \\
\hline
0.15 & 0.23 & 0.30 & \textbf{0.47} & \underline{0.41} & 0.40 & 0.16 & 0.18 & 0.15 & 0.22 & 0.34 & 0.30 \\
\end{tabular}
\end{table}

%% file: 2014-04-07-plot-v2/mix_det-desc-match_CenSurE_0_bonscorresp_best_gphe.tex
\begin{figure}[htbp]
\centering
\includegraphics{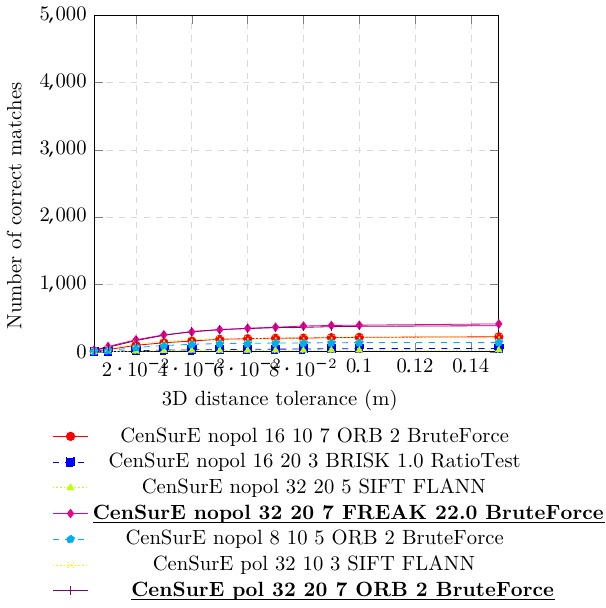}
\caption{Number of correct matches with CenSurE setups over distance tolerance for pair 0 "translation". Can be split into 2 clusters, the well performing setups are boldened in the legend.
Underlined setups in the legend have the best overall number of correct matches among the good performers in terms of matching score highlighted in figure~\ref{fig:matching-score-image-0-censure}.}
\label{fig:nombre-de-bons-correspondants-image-0-censure}
\end{figure}

%% file: 2014-04-07-plot-v2/mix_det-desc-match_CenSurE_0_mscore_best_gphe.tex
\begin{figure}[htbp]
\centering
\includegraphics{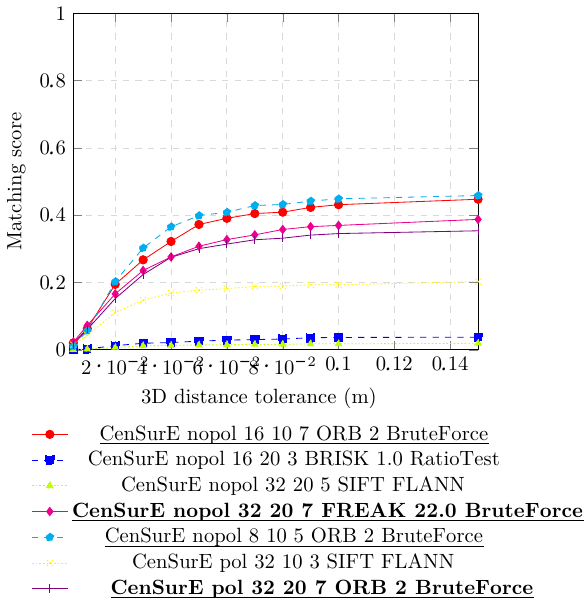}
\caption{Matching score with CenSurE setups over distance tolerance for pair 0 "translation".
The best performing setups are underlined in the legend.
Boldened setup in the legend has the best overall matching score among the good performers in terms of number of correct matches highlighted in figure~\ref{fig:nombre-de-bons-correspondants-image-0-censure}.}
\label{fig:matching-score-image-0-censure}
\end{figure}

%% file: 2014-04-07-plot-v2/mix_det-desc-match_CenSurE_0_bonscorresp_best_tab.tex
\begin{table}[htbp]
\caption{Number of correct matches with CenSurE setups for pair 0 "translation", numerical values. Bold values are the best for a given tolerance, underlined are second best.
\label{tab:nombre-de-bons-correspondants-image-0-censure}}
\centering
\begin{tabular}{l*{7}{|c}}
\multicolumn{1}{c|}{\rotatebox{90}{Distance tolerance (m)}} & \rotatebox{90}{CenSurE nopol 16 10 7 ORB 2 BruteForce} & \rotatebox{90}{CenSurE nopol 16 20 3 BRISK 1.0 RatioTest} & \rotatebox{90}{CenSurE nopol 32 20 5 SIFT FLANN} & \rotatebox{90}{CenSurE nopol 32 20 7 FREAK 22.0 BruteForce} & \rotatebox{90}{CenSurE nopol 8 10 5 ORB 2 BruteForce} & \rotatebox{90}{CenSurE pol 32 10 3 SIFT FLANN} & \rotatebox{90}{CenSurE pol 32 20 7 ORB 2 BruteForce} \\
\hline
\hline
0.005 & 10 & 1 & 0 & \textbf{24} & 3 & 16 & \underline{17} \\
\hline
0.01 & 31 & 3 & 2 & \textbf{77} & 17 & 46 & \underline{66} \\
\hline
0.02 & 96 & 15 & 8 & \textbf{177} & 61 & 118 & \underline{166} \\
\hline
0.03 & 132 & 24 & 15 & \textbf{251} & 91 & 157 & \underline{244} \\
\hline
0.04 & 159 & 27 & 16 & \underline{294} & 110 & 180 & \textbf{300} \\
\hline
0.05 & 184 & 32 & 18 & \textbf{328} & 120 & \underline{189} & \textbf{328} \\
\hline
0.06 & 193 & 36 & 19 & \textbf{349} & 123 & 194 & \underline{343} \\
\hline
0.07 & 200 & 38 & 20 & \textbf{364} & 129 & 201 & \underline{357} \\
\hline
0.08 & 202 & 40 & 20 & \textbf{381} & 130 & 202 & \underline{362} \\
\hline
0.09 & 209 & 45 & 22 & \textbf{390} & 133 & 206 & \underline{372} \\
\hline
0.1 & 213 & 46 & 22 & \textbf{394} & 135 & 206 & \underline{377} \\
\hline
0.15 & 221 & 47 & 24 & \textbf{413} & 138 & 216 & \underline{386} \\
\end{tabular}
\end{table}

%% file: 2014-04-07-plot-v2/mix_det-desc-match_CenSurE_0_mscore_best_tab.tex
\begin{table}[htbp]
\caption{Matching score with CenSurE setups for pair 0 "translation", numerical values. Bold values are the best for a given tolerance, underlined are second best.
\label{tab:matching-score-image-0-censure}}
\centering
\begin{tabular}{l*{7}{|c}}
\multicolumn{1}{c|}{\rotatebox{90}{Distance tolerance (m)}} & \rotatebox{90}{CenSurE nopol 16 10 7 ORB 2 BruteForce} & \rotatebox{90}{CenSurE nopol 16 20 3 BRISK 1.0 RatioTest} & \rotatebox{90}{CenSurE nopol 32 20 5 SIFT FLANN} & \rotatebox{90}{CenSurE nopol 32 20 7 FREAK 22.0 BruteForce} & \rotatebox{90}{CenSurE nopol 8 10 5 ORB 2 BruteForce} & \rotatebox{90}{CenSurE pol 32 10 3 SIFT FLANN} & \rotatebox{90}{CenSurE pol 32 20 7 ORB 2 BruteForce} \\
\hline
\hline
0.005 & \underline{0.02} & 0.00 & 0.00 & \textbf{0.02} & 0.01 & 0.01 & 0.02 \\
\hline
0.01 & \underline{0.06} & 0.00 & 0.00 & \textbf{0.07} & 0.06 & 0.04 & 0.06 \\
\hline
0.02 & \underline{0.19} & 0.01 & 0.01 & 0.17 & \textbf{0.20} & 0.11 & 0.15 \\
\hline
0.03 & \underline{0.27} & 0.02 & 0.01 & 0.24 & \textbf{0.30} & 0.15 & 0.22 \\
\hline
0.04 & \underline{0.32} & 0.02 & 0.01 & 0.28 & \textbf{0.37} & 0.17 & 0.27 \\
\hline
0.05 & \underline{0.37} & 0.03 & 0.01 & 0.31 & \textbf{0.40} & 0.18 & 0.30 \\
\hline
0.06 & \underline{0.39} & 0.03 & 0.01 & 0.33 & \textbf{0.41} & 0.18 & 0.31 \\
\hline
0.07 & \underline{0.40} & 0.03 & 0.02 & 0.34 & \textbf{0.43} & 0.19 & 0.33 \\
\hline
0.08 & \underline{0.41} & 0.03 & 0.02 & 0.36 & \textbf{0.43} & 0.19 & 0.33 \\
\hline
0.09 & \underline{0.42} & 0.04 & 0.02 & 0.37 & \textbf{0.44} & 0.19 & 0.34 \\
\hline
0.1 & \underline{0.43} & 0.04 & 0.02 & 0.37 & \textbf{0.45} & 0.19 & 0.35 \\
\hline
0.15 & \underline{0.45} & 0.04 & 0.02 & 0.39 & \textbf{0.46} & 0.20 & 0.35 \\
\end{tabular}
\end{table}

%% file: 2014-04-07-plot-v2/mix_det-desc-match_CenSurE_40_bonscorresp_best_gphe.tex
\begin{figure}[htbp]
\centering
\includegraphics{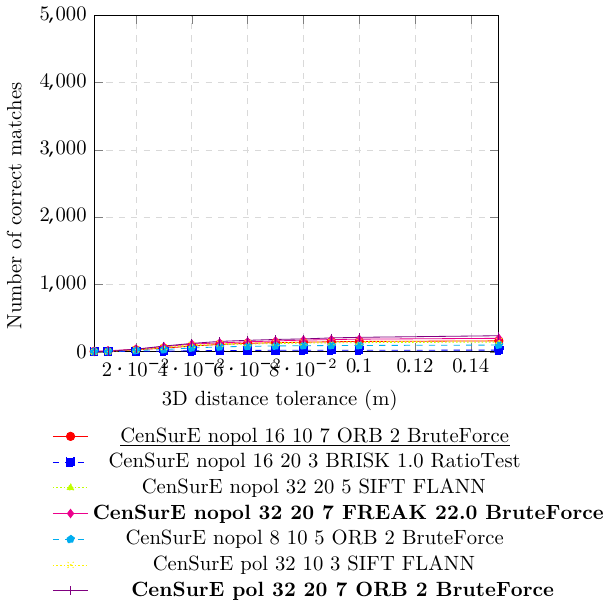}
\caption{Number of correct matches with CenSurE setups over distance tolerance for pair 40 "translation motion". Can be split into 2 clusters, the well performing setups are boldened in the legend.
Underlined setup in the legend has the best overall number of correct matches among the good performers in terms of matching score highlighted in figure~\ref{fig:matching-score-image-40-censure}.
Here, whatever the settings, the number of correct matches is too low to guarantee acceptabe performance in real use.}
\label{fig:nombre-de-bons-correspondants-image-40-censure}
\end{figure}

%% file: 2014-04-07-plot-v2/mix_det-desc-match_CenSurE_40_mscore_best_gphe.tex
\begin{figure}[htbp]
\centering
\includegraphics{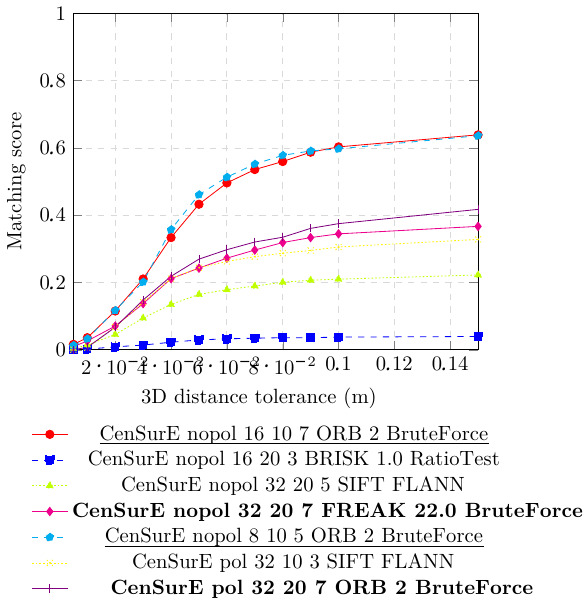}
\caption{Matching score with CenSurE setups over distance tolerance for pair 40 "translation motion".
The best performing setups are underlined in the legend.
Boldened setup in the legend has the best overall matching score among the good performers in terms of number of correct matches highlighted in figure~\ref{fig:nombre-de-bons-correspondants-image-40-censure}.}
\label{fig:matching-score-image-40-censure}
\end{figure}

%% file: 2014-04-07-plot-v2/mix_det-desc-match_CenSurE_40_bonscorresp_best_tab.tex
\begin{table}[htbp]
\caption{Number of correct matches with CenSurE setups for pair 40 "translation motion", numerical values. Bold values are the best for a given tolerance, underlined are second best.
\label{tab:nombre-de-bons-correspondants-image-40-censure}}
\centering
\begin{tabular}{l*{7}{|c}}
\multicolumn{1}{c|}{\rotatebox{90}{Distance tolerance (m)}} & \rotatebox{90}{CenSurE nopol 16 10 7 ORB 2 BruteForce} & \rotatebox{90}{CenSurE nopol 16 20 3 BRISK 1.0 RatioTest} & \rotatebox{90}{CenSurE nopol 32 20 5 SIFT FLANN} & \rotatebox{90}{CenSurE nopol 32 20 7 FREAK 22.0 BruteForce} & \rotatebox{90}{CenSurE nopol 8 10 5 ORB 2 BruteForce} & \rotatebox{90}{CenSurE pol 32 10 3 SIFT FLANN} & \rotatebox{90}{CenSurE pol 32 20 7 ORB 2 BruteForce} \\
\hline
\hline
0.005 & \underline{4} & 0 & 2 & \textbf{5} & 2 & 0 & 0 \\
\hline
0.01 & \underline{9} & 1 & \underline{9} & \textbf{14} & 5 & 3 & 4 \\
\hline
0.02 & 29 & 5 & 29 & \textbf{38} & 18 & \underline{33} & \textbf{38} \\
\hline
0.03 & 53 & 8 & 60 & \underline{74} & 31 & 69 & \textbf{84} \\
\hline
0.04 & 84 & 13 & 86 & \underline{114} & 55 & 103 & \textbf{124} \\
\hline
0.05 & 109 & 17 & 105 & \underline{131} & 71 & 119 & \textbf{153} \\
\hline
0.06 & 125 & 19 & 114 & 147 & 79 & \underline{128} & \textbf{169} \\
\hline
0.07 & 135 & 20 & 121 & \underline{160} & 85 & 135 & \textbf{182} \\
\hline
0.08 & 141 & 21 & 128 & \underline{172} & 89 & 140 & \textbf{190} \\
\hline
0.09 & 148 & 21 & 132 & \underline{180} & 91 & 144 & \textbf{205} \\
\hline
0.1 & 152 & 22 & 134 & \underline{186} & 92 & 149 & \textbf{213} \\
\hline
0.15 & 161 & 23 & 142 & \underline{198} & 98 & 160 & \textbf{237} \\
\end{tabular}
\end{table}

%% file: 2014-04-07-plot-v2/mix_det-desc-match_CenSurE_40_mscore_best_tab.tex
\begin{table}[htbp]
\caption{Matching score with CenSurE setups for pair 40 "translation motion", numerical values. Bold values are the best for a given tolerance, underlined are second best.
\label{tab:matching-score-image-40-censure}}
\centering
\begin{tabular}{l*{7}{|c}}
\multicolumn{1}{c|}{\rotatebox{90}{Distance tolerance (m)}} & \rotatebox{90}{CenSurE nopol 16 10 7 ORB 2 BruteForce} & \rotatebox{90}{CenSurE nopol 16 20 3 BRISK 1.0 RatioTest} & \rotatebox{90}{CenSurE nopol 32 20 5 SIFT FLANN} & \rotatebox{90}{CenSurE nopol 32 20 7 FREAK 22.0 BruteForce} & \rotatebox{90}{CenSurE nopol 8 10 5 ORB 2 BruteForce} & \rotatebox{90}{CenSurE pol 32 10 3 SIFT FLANN} & \rotatebox{90}{CenSurE pol 32 20 7 ORB 2 BruteForce} \\
\hline
\hline
0.005 & \textbf{0.02} & 0.00 & 0.00 & 0.01 & \underline{0.01} & 0.00 & 0.00 \\
\hline
0.01 & \textbf{0.04} & 0.00 & 0.01 & 0.03 & \underline{0.03} & 0.01 & 0.01 \\
\hline
0.02 & \underline{0.12} & 0.01 & 0.05 & 0.07 & \textbf{0.12} & 0.07 & 0.07 \\
\hline
0.03 & \textbf{0.21} & 0.01 & 0.09 & 0.14 & \underline{0.20} & 0.14 & 0.15 \\
\hline
0.04 & \underline{0.33} & 0.02 & 0.13 & 0.21 & \textbf{0.36} & 0.21 & 0.22 \\
\hline
0.05 & \underline{0.43} & 0.03 & 0.16 & 0.24 & \textbf{0.46} & 0.24 & 0.27 \\
\hline
0.06 & \underline{0.50} & 0.03 & 0.18 & 0.27 & \textbf{0.51} & 0.26 & 0.30 \\
\hline
0.07 & \underline{0.54} & 0.03 & 0.19 & 0.30 & \textbf{0.55} & 0.28 & 0.32 \\
\hline
0.08 & \underline{0.56} & 0.04 & 0.20 & 0.32 & \textbf{0.58} & 0.29 & 0.33 \\
\hline
0.09 & \underline{0.59} & 0.04 & 0.21 & 0.33 & \textbf{0.59} & 0.30 & 0.36 \\
\hline
0.1 & \textbf{0.60} & 0.04 & 0.21 & 0.34 & \underline{0.60} & 0.31 & 0.38 \\
\hline
0.15 & \textbf{0.64} & 0.04 & 0.22 & 0.37 & \underline{0.64} & 0.33 & 0.42 \\
\end{tabular}
\end{table}

%% file: 2014-04-07-plot-v2/mix_det-desc-match_CenSurE_24_bonscorresp_best_gphe.tex
\begin{figure}[htbp]
\centering
\includegraphics{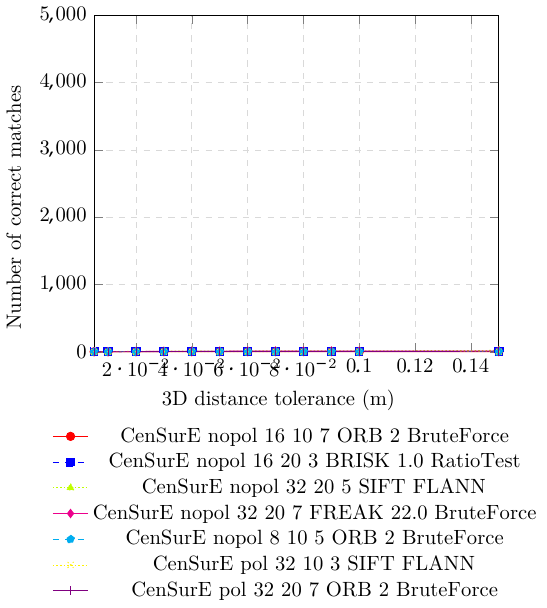}
\caption{Number of correct matches with CenSurE setups over distance tolerance for pair 24 "translation and rotation motion". All the alternative setups fail the evaluation on this scene.}
\label{fig:nombre-de-bons-correspondants-image-24-censure}
\end{figure}

%% file: 2014-04-07-plot-v2/mix_det-desc-match_CenSurE_24_bonscorresp_best_tab.tex
\begin{table}[htbp]
\caption{Number of correct matches with CenSurE setups for pair 24 "translation and rotation motion", numerical values. Bold values are the best for a given tolerance, underlined are second best. All the alternative setups fail the evaluation on this scene.
\label{tab:nombre-de-bons-correspondants-image-24-censure}}
\centering
\begin{tabular}{l*{7}{|c}}
\multicolumn{1}{c|}{\rotatebox{90}{Distance tolerance (m)}} & \rotatebox{90}{CenSurE nopol 16 10 7 ORB 2 BruteForce} & \rotatebox{90}{CenSurE nopol 16 20 3 BRISK 1.0 RatioTest} & \rotatebox{90}{CenSurE nopol 32 20 5 SIFT FLANN} & \rotatebox{90}{CenSurE nopol 32 20 7 FREAK 22.0 BruteForce} & \rotatebox{90}{CenSurE nopol 8 10 5 ORB 2 BruteForce} & \rotatebox{90}{CenSurE pol 32 10 3 SIFT FLANN} & \rotatebox{90}{CenSurE pol 32 20 7 ORB 2 BruteForce} \\
\hline
\hline
0.005 & 0 & 0 & \textbf{1} & \textbf{1} & 0 & 0 & 0 \\
\hline
0.01 & 0 & \underline{1} & \textbf{2} & \underline{1} & 0 & 0 & 0 \\
\hline
0.02 & 1 & 2 & \textbf{5} & \underline{3} & 0 & 0 & 1 \\
\hline
0.03 & 1 & 2 & \textbf{8} & \underline{6} & 0 & 0 & 2 \\
\hline
0.04 & 1 & 2 & \textbf{9} & \underline{7} & 0 & 1 & 4 \\
\hline
0.05 & 1 & 2 & \textbf{9} & \underline{7} & 0 & 1 & 4 \\
\hline
0.06 & 1 & 2 & \textbf{11} & \underline{9} & 0 & 1 & 4 \\
\hline
0.07 & 1 & 2 & \textbf{11} & \underline{9} & 0 & 1 & 4 \\
\hline
0.08 & 1 & 2 & \textbf{12} & \underline{9} & 0 & 2 & 4 \\
\hline
0.09 & 1 & 2 & \textbf{12} & \underline{9} & 0 & 2 & 4 \\
\hline
0.1 & 1 & 2 & \textbf{12} & \underline{9} & 0 & 2 & 4 \\
\hline
0.15 & 1 & 2 & \textbf{19} & \underline{9} & 0 & 2 & 5 \\
\end{tabular}
\end{table}

%% file: 2014-05-11-pts-reel-cluster-v2/stats_SIFT_7100_tab.tex
\begin{table}[htbp]
\caption{Statistics of the evaluation metrics with SIFT detector based setups for real scene 7100 "translation".
Boldened values are the most desirable for each criteria.
Higher values are better, except the \textsc{ransac} iterations that shall be lower for a quick convergence.
Underlined setups are those with low stability highlighted in table~\ref{tab:statistiques-stabilite-image-7100-sift}.
\label{tab:statistiques-image-reelle-numero-7100-sift}}
\centering
\begin{adjustbox}{width=\textwidth}
\begin{tabular}{l*{7}{|c}}
Detection/description/matching algorithms setup & \rotatebox{90}{$\sharp$detections} & \rotatebox{90}{$\sharp$attempted matches} & \rotatebox{90}{$\sharp$attempted matches / $\sharp$detections (\%)} & \rotatebox{90}{Average of $\sharp$inliers} & \rotatebox{90}{Average of $\sharp$inliers / $\sharp$detections (\%)} & \rotatebox{90}{Average of $\sharp$inliers / $\sharp$attempted matches (\%)} & \rotatebox{90}{Average of $\sharp$\textsc{ransac} iterations}
 \\
\hline
\hline
SIFT nopol 0.02 20.0 1.0 SIFT BruteForce & 9873 & \textbf{7255} & 0.73 & 1350.65 & 0.14 & 0.19 & 50000 \\
\hline
SIFT nopol 0.02 20.0 1.6 SIFT BruteForce & 9305 & 6851 & 0.74 & \textbf{1429.74} & 0.15 & 0.21 & 50000 \\
\hline
\underline{SIFT nopol 0.04 5.0 1.6 FREAK 44.0 BruteForce} & 2617 & 1569 & 0.60 & 225.50 & 0.09 & 0.14 & 50000 \\
\hline
SIFT nopol 0.04 5.0 1.6 SIFT BruteForce & 2617 & 1977 & 0.76 & 449.51 & 0.17 & 0.23 & 50000 \\
\hline
SIFT nopol 0.04 5.0 3.0 SIFT BruteForce & 1743 & 1323 & \textbf{0.76} & 363.01 & 0.21 & 0.27 & 50000 \\
\hline
SIFT nopol 0.08 10.0 3.0 SIFT BruteForce & 880 & 654 & 0.74 & 177.92 & 0.20 & 0.27 & 50000 \\
\hline
\underline{SIFT nopol 0.08 5.0 3.0 FREAK 44.0 BruteForce} & 722 & 461 & 0.64 & 112.64 & 0.16 & 0.24 & 50000 \\
\hline
SIFT nopol 0.08 5.0 3.0 SIFT BruteForce & 722 & 538 & 0.75 & 156.27 & \textbf{0.22} & \textbf{0.29} & 50000 \\
\hline
SIFT pol 0.02 20.0 1.6 SIFT BruteForce & \textbf{10522} & 6936 & 0.66 & 1369.95 & 0.13 & 0.20 & 50000 \\
\hline
SIFT pol 0.02 20.0 3.0 SIFT BruteForce & 8776 & 5677 & 0.65 & 1425.56 & 0.16 & 0.25 & 50000 \\
\hline
\underline{SIFT pol 0.04 5.0 3.0 FREAK 44.0 BruteForce} & 2439 & 1239 & 0.51 & 25.56 & 0.01 & 0.02 & 50000 \\
\end{tabular}
\end{adjustbox}
\end{table}

%% file: 2014-05-11-pts-reel-cluster-v2/stats_SIFT_10627_tab.tex
\begin{table}[htbp]
\caption{Statistics of the evaluation metrics with SIFT detector based setups for real scene 10627 "translation motion".
Boldened values are the most desirable for each criteria.
Higher values are better, except the \textsc{ransac} iterations that shall be lower for a quick convergence.
Underlined setups are those with low stability highlighted in table~\ref{tab:statistiques-stabilite-image-10627-sift}.
\label{tab:statistiques-image-reelle-numero-10627-sift}}
\centering
\begin{adjustbox}{width=\textwidth}
\begin{tabular}{l*{7}{|c}}
Detection/description/matching algorithms setup & \rotatebox{90}{$\sharp$detections} & \rotatebox{90}{$\sharp$attempted matches} & \rotatebox{90}{$\sharp$attempted matches / $\sharp$detections (\%)} & \rotatebox{90}{Average of $\sharp$inliers} & \rotatebox{90}{Average of $\sharp$inliers / $\sharp$detections (\%)} & \rotatebox{90}{Average of $\sharp$inliers / $\sharp$attempted matches (\%)} & \rotatebox{90}{Average of $\sharp$\textsc{ransac} iterations}
 \\
\hline
\hline
SIFT nopol 0.02 20.0 1.0 SIFT BruteForce & 4527 & 3032 & 0.67 & 606.82 & 0.13 & 0.20 & 50000 \\
\hline
SIFT nopol 0.02 20.0 1.6 SIFT BruteForce & 3686 & 2500 & 0.68 & \textbf{638.62} & 0.17 & 0.26 & 50000 \\
\hline
\underline{SIFT nopol 0.04 5.0 1.6 FREAK 44.0 BruteForce} & 1067 & 622 & 0.58 & 62.71 & 0.06 & 0.10 & 50000 \\
\hline
SIFT nopol 0.04 5.0 1.6 SIFT BruteForce & 1067 & 748 & 0.70 & 243.58 & 0.23 & 0.33 & 50000 \\
\hline
SIFT nopol 0.04 5.0 3.0 SIFT BruteForce & 1001 & 707 & 0.71 & 268.52 & \textbf{0.27} & \textbf{0.38} & \textbf{49956.44} \\
\hline
SIFT nopol 0.08 10.0 3.0 SIFT BruteForce & 697 & 478 & 0.69 & 160.95 & 0.23 & 0.34 & 50000 \\
\hline
\underline{SIFT nopol 0.08 5.0 3.0 FREAK 44.0 BruteForce} & 496 & 299 & 0.60 & 38.57 & 0.08 & 0.13 & 50000 \\
\hline
SIFT nopol 0.08 5.0 3.0 SIFT BruteForce & 496 & 352 & \textbf{0.71} & 119.71 & 0.24 & 0.34 & 50000 \\
\hline
SIFT pol 0.02 20.0 1.6 SIFT BruteForce & \textbf{6154} & \textbf{3199} & 0.52 & 594.51 & 0.10 & 0.19 & 50000 \\
\hline
SIFT pol 0.02 20.0 3.0 SIFT BruteForce & 4831 & 2446 & 0.51 & 605.00 & 0.13 & 0.25 & 50000 \\
\hline
\underline{SIFT pol 0.04 5.0 3.0 FREAK 44.0 BruteForce} & 1173 & 539 & 0.46 & 19.87 & 0.02 & 0.04 & 50000 \\
\end{tabular}
\end{adjustbox}
\end{table}

%% file: 2014-05-11-pts-reel-cluster-v2/stats_SIFT_11048_tab.tex
\begin{table}[htbp]
\caption{Statistics of the evaluation metrics with SIFT detector based setups for real scene 11048 "translation and rotation motion".
Boldened values are the most desirable for each criteria.
Higher values are better, except the \textsc{ransac} iterations that shall be lower for a quick convergence.
Underlined setups are those with low stability highlighted in table~\ref{tab:statistiques-stabilite-image-11048-sift}.
\label{tab:statistiques-image-reelle-numero-11048-sift}}
\centering
\begin{adjustbox}{width=\textwidth}
\begin{tabular}{l*{7}{|c}}
Detection/description/matching algorithms setup & \rotatebox{90}{$\sharp$detections} & \rotatebox{90}{$\sharp$attempted matches} & \rotatebox{90}{$\sharp$attempted matches / $\sharp$detections (\%)} & \rotatebox{90}{Average of $\sharp$inliers} & \rotatebox{90}{Average of $\sharp$inliers / $\sharp$detections (\%)} & \rotatebox{90}{Average of $\sharp$inliers / $\sharp$attempted matches (\%)} & \rotatebox{90}{Average of $\sharp$\textsc{ransac} iterations}
 \\
\hline
\hline
SIFT nopol 0.02 20.0 1.0 SIFT BruteForce & 4433 & \textbf{3028} & 0.68 & 786.41 & 0.18 & 0.26 & 50000 \\
\hline
SIFT nopol 0.02 20.0 1.6 SIFT BruteForce & 4162 & 2856 & 0.69 & \textbf{867.00} & 0.21 & 0.30 & 50000 \\
\hline
\underline{SIFT nopol 0.04 5.0 1.6 FREAK 44.0 BruteForce} & 1709 & 987 & 0.58 & 193.71 & 0.11 & 0.20 & 50000 \\
\hline
SIFT nopol 0.04 5.0 1.6 SIFT BruteForce & 1709 & 1215 & \textbf{0.71} & 431.49 & 0.25 & 0.36 & 50000 \\
\hline
SIFT nopol 0.04 5.0 3.0 SIFT BruteForce & 1335 & 945 & 0.71 & 365.01 & 0.27 & 0.39 & 49932.60 \\
\hline
SIFT nopol 0.08 10.0 3.0 SIFT BruteForce & 825 & 579 & 0.70 & 227.33 & 0.28 & 0.39 & 49377.39 \\
\hline
\underline{SIFT nopol 0.08 5.0 3.0 FREAK 44.0 BruteForce} & 677 & 399 & 0.59 & 97.97 & 0.14 & 0.25 & 50000 \\
\hline
SIFT nopol 0.08 5.0 3.0 SIFT BruteForce & 677 & 478 & 0.71 & 195.40 & \textbf{0.29} & \textbf{0.41} & \textbf{47069.40} \\
\hline
SIFT pol 0.02 20.0 1.6 SIFT BruteForce & \textbf{4531} & 2794 & 0.62 & 795.39 & 0.18 & 0.28 & 50000 \\
\hline
SIFT pol 0.02 20.0 3.0 SIFT BruteForce & 4255 & 2580 & 0.61 & 841.39 & 0.20 & 0.33 & 50000 \\
\hline
\underline{SIFT pol 0.04 5.0 3.0 FREAK 44.0 BruteForce} & 1688 & 817 & 0.48 & 23.37 & 0.01 & 0.03 & 50000 \\
\end{tabular}
\end{adjustbox}
\end{table}

%% file: 2014-05-11-pts-reel-cluster-v2/stats_CenSurE_7100_tab.tex
\begin{table}[htbp]
\caption{Statistics of the evaluation metrics with CenSurE detector based setups for real scene 7100 "translation".
Boldened values are the most desirable for each criteria.
Higher values are better, except the \textsc{ransac} iterations that shall be lower for a quick convergence.
\label{tab:statistiques-image-reelle-numero-7100-censure}}
\centering
\begin{adjustbox}{width=\textwidth}
\begin{tabular}{l*{7}{|c}}
Detection/description/matching algorithms setup & \rotatebox{90}{$\sharp$detections} & \rotatebox{90}{$\sharp$attempted matches} & \rotatebox{90}{$\sharp$attempted matches / $\sharp$detections (\%)} & \rotatebox{90}{Average of $\sharp$inliers} & \rotatebox{90}{Average of $\sharp$inliers / $\sharp$detections (\%)} & \rotatebox{90}{Average of $\sharp$inliers / $\sharp$attempted matches (\%)} & \rotatebox{90}{Average of $\sharp$\textsc{ransac} iterations}
 \\
\hline
\hline
CenSurE nopol 16 10 7 ORB 2 BruteForce & 645 & 394 & 0.61 & 146.06 & \textbf{0.23} & 0.37 & 49987.35 \\
\hline
CenSurE nopol 16 20 3 BRISK 1.0 RatioTest & 1330 & 175 & 0.13 & 132.13 & 0.10 & 0.76 & 332.89 \\
\hline
CenSurE nopol 32 20 5 SIFT FLANN & 1178 & 207 & 0.18 & 148.30 & 0.13 & 0.72 & 727.90 \\
\hline
CenSurE nopol 32 20 7 FREAK 22.0 BruteForce & 973 & 559 & 0.57 & 142.52 & 0.15 & 0.25 & 50000 \\
\hline
CenSurE nopol 8 10 5 ORB 2 BruteForce & 422 & 261 & \textbf{0.62} & 84.17 & 0.20 & 0.32 & 50000 \\
\hline
CenSurE pol 32 10 3 SIFT FLANN & \textbf{1359} & 151 & 0.11 & 124.54 & 0.09 & \textbf{0.82} & \textbf{126.83} \\
\hline
CenSurE pol 32 20 7 ORB 2 BruteForce & 1056 & \textbf{583} & 0.55 & \textbf{171.32} & 0.16 & 0.29 & 50000 \\
\end{tabular}
\end{adjustbox}
\end{table}

%% file: 2014-05-11-pts-reel-cluster-v2/stats_CenSurE_10627_tab.tex
\begin{table}[htbp]
\caption{Statistics of the evaluation metrics with CenSurE detector based setups for real scene 10627 "translation motion".
Boldened values are the most desirable for each criteria.
Higher values are better, except the \textsc{ransac} iterations that shall be lower for a quick convergence.
Underlined setups are those with low stability highlighted in table~\ref{tab:statistiques-stabilite-image-10627-censure}.
\label{tab:statistiques-image-reelle-numero-10627-censure}}
\centering
\begin{adjustbox}{width=\textwidth}
\begin{tabular}{l*{7}{|c}}
Detection/description/matching algorithms setup & \rotatebox{90}{$\sharp$detections} & \rotatebox{90}{$\sharp$attempted matches} & \rotatebox{90}{$\sharp$attempted matches / $\sharp$detections (\%)} & \rotatebox{90}{Average of $\sharp$inliers} & \rotatebox{90}{Average of $\sharp$inliers / $\sharp$detections (\%)} & \rotatebox{90}{Average of $\sharp$inliers / $\sharp$attempted matches (\%)} & \rotatebox{90}{Average of $\sharp$\textsc{ransac} iterations}
 \\
\hline
\hline
CenSurE nopol 16 10 7 ORB 2 BruteForce & 322 & 209 & 0.65 & 67.50 & 0.21 & 0.32 & 50000 \\
\hline
CenSurE nopol 16 20 3 BRISK 1.0 RatioTest & 1092 & 93 & 0.09 & 52.12 & 0.05 & 0.56 & 8158.02 \\
\hline
\underline{CenSurE nopol 32 20 5 SIFT FLANN} & 1106 & 121 & 0.11 & 82.83 & 0.07 & 0.68 & 1579.07 \\
\hline
CenSurE nopol 32 20 7 FREAK 22.0 BruteForce & 853 & \textbf{509} & 0.60 & \textbf{116.51} & 0.14 & 0.23 & 50000 \\
\hline
CenSurE nopol 8 10 5 ORB 2 BruteForce & 137 & 88 & \textbf{0.64} & 30.91 & \textbf{0.23} & 0.35 & 50000 \\
\hline
\underline{CenSurE pol 32 10 3 SIFT FLANN} & 920 & 69 & 0.07 & 53.98 & 0.06 & \textbf{0.78} & \textbf{271.87} \\
\hline
CenSurE pol 32 20 7 ORB 2 BruteForce & \textbf{1142} & 474 & 0.42 & 111.21 & 0.10 & 0.23 & 50000 \\
\end{tabular}
\end{adjustbox}
\end{table}

%% file: 2014-05-11-pts-reel-cluster-v2/stats_CenSurE_11048_tab.tex
\begin{table}[htbp]
\caption{Statistics of the evaluation metrics with CenSurE detector based setups for real scene 11048 "translation and rotation motion".
Boldened values are the most desirable for each criteria.
Higher values are better, except the \textsc{ransac} iterations that shall be lower for a quick convergence.
Underlined setup is the one with low stability highlighted in table~\ref{tab:statistiques-stabilite-image-11048-censure}.
\label{tab:statistiques-image-reelle-numero-11048-censure}}
\centering
\begin{adjustbox}{width=\textwidth}
\begin{tabular}{l*{7}{|c}}
Detection/description/matching algorithms setup & \rotatebox{90}{$\sharp$detections} & \rotatebox{90}{$\sharp$attempted matches} & \rotatebox{90}{$\sharp$attempted matches / $\sharp$detections (\%)} & \rotatebox{90}{Average of $\sharp$inliers} & \rotatebox{90}{Average of $\sharp$inliers / $\sharp$detections (\%)} & \rotatebox{90}{Average of $\sharp$inliers / $\sharp$attempted matches (\%)} & \rotatebox{90}{Average of $\sharp$\textsc{ransac} iterations}
 \\
\hline
\hline
CenSurE nopol 16 10 7 ORB 2 BruteForce & 518 & 347 & \textbf{0.67} & 164.10 & 0.32 & 0.47 & 25020.49 \\
\hline
CenSurE nopol 16 20 3 BRISK 1.0 RatioTest & \textbf{1219} & 193 & 0.16 & 142.19 & 0.12 & 0.74 & \textbf{628.90} \\
\hline
\underline{CenSurE nopol 32 20 5 SIFT FLANN} & 1117 & 131 & 0.12 & 98.18 & 0.09 & \textbf{0.75} & 1471.43 \\
\hline
CenSurE nopol 32 20 7 FREAK 22.0 BruteForce & 910 & 544 & 0.60 & 157.46 & 0.17 & 0.29 & 50000 \\
\hline
CenSurE nopol 8 10 5 ORB 2 BruteForce & 294 & 187 & 0.64 & 96.42 & \textbf{0.33} & 0.52 & 14591.15 \\
\hline
CenSurE pol 32 10 3 SIFT FLANN & 1102 & \textbf{601} & 0.55 & \textbf{333.70} & 0.30 & 0.56 & 7624.01 \\
\hline
CenSurE pol 32 20 7 ORB 2 BruteForce & 884 & 516 & 0.58 & 220.35 & 0.25 & 0.43 & 40261.36 \\
\end{tabular}
\end{adjustbox}
\end{table}